\documentclass{article}

\usepackage[table,xcdraw,dvipsnames]{xcolor}
\usepackage[final]{corl_2026} 
\usepackage{booktabs}
\usepackage{multirow}
\usepackage{graphicx}
\usepackage{amsmath}
\usepackage{amssymb}
\usepackage{caption}
\usepackage{lipsum}
\usepackage{subcaption}
\usepackage{wrapfig}
\usepackage{cleveref}
\usepackage{float}

\title{HumanoidNDM: Closing the Training-Deployment Gap of Motion Tracking via Humanoid Neural Dynamics Model\vspace{-0.6em}}

\author{
    Chao Chen$^{1,3,*}$ \quad
    Zhenrong Wang$^{2,3,4,*}$ \quad
    Jilong Wang$^3$ \quad
    Yusen Feng$^{1,3}$ \\
    \bf Jiangran Lyu$^{1,3}$ \quad
    Li Yi$^{3,5,\dagger}$ \quad
    He Wang$^{1,3,\dagger}$ \\
    $^1$CFCS, School of Computer Science, Peking University \\
    $^2$Institute of Automation, Chinese Academy of Sciences \\
    $^3$Galbot \quad $^4$Beijing Academy of Artificial Intelligence \quad $^5$Tsinghua University\\
    \small $^{*}$Equal contribution. $^{\dagger}$Corresponding authors.\\[0.5em]
    \small \textbf{Project Page:} \url{https://pku-epic.github.io/HumanoidNDM}
}

\begin{document}
\maketitle

\vspace{-2.5em}
\begin{figure}[htbp]
    \centering
    \includegraphics[width=\textwidth, height=17\baselineskip, keepaspectratio]{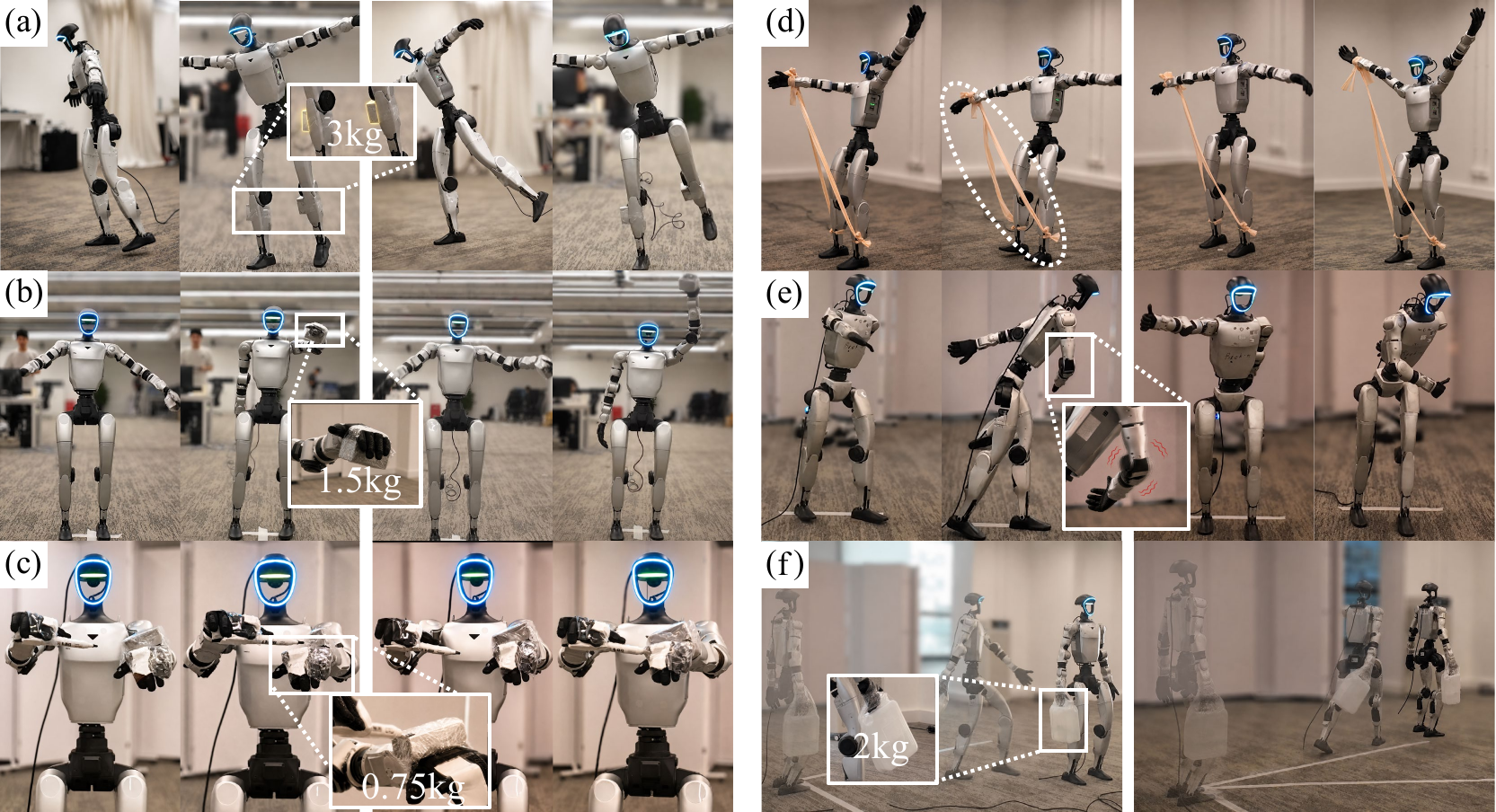}
    \caption{Comparison of direct transfer (left two frames) and fine-tuned performance (right two frames): (a) \textbf{Swallow Balance}: 1.5 kg weights on each calf; (b) \textbf{Arm Swing}: 1.5 kg weight on the left hand; (c) \textbf{Pen Capping}: 0.75 kg load on the left hand; (d) \textbf{Lateral Raise}: two elastic bands attached to wrists; (e) \textbf{Bowing}: damaged arm motor; and (f) \textbf{Walking}: 2 kg water bucket in hand.}
    \label{fig:teaser}
\end{figure}
\vspace{-1.5em}

\begin{abstract}
    Humanoid robots have recently achieved remarkable success in motion tracking.
    However, during real-world deployment, robots inevitably encounter dynamics gaps—arising from either extra physical disturbances or inaccurate simulation modeling—resulting in a significant training-deployment gap. Therefore, effectively leveraging real-world data to fine-tune policies represents the crucial ``last mile'' for robust deployment.
    To bridge this gap, we propose HumanoidNDM, an online fine-tuning framework for humanoid motion tracking driven by a real-world-trained neural dynamics model.
    First, in the data collection stage, we introduce a low-frequency reference perturbation method with an adaptive intensity mechanism, balancing deployment feasibility and exploration efficiency.
    Second, for the model training stage, we train a robust history-based, proprioceptive neural dynamics model in Euclidean space by incorporating differentiable forward kinematics, eliminating the reliance on costly motion capture systems.
    Finally, in the policy fine-tuning stage, we directly fine-tune the policy via backpropagated model gradients, achieving high data efficiency.
    Through physical validation on the Unitree G1 robot, we demonstrate that HumanoidNDM enables rapid real-world adaptation to various perturbations, representing the first online, MoCap-free fine-tuning framework for humanoid tracking policies.
\end{abstract}



\section{Introduction}
Advancements in reinforcement learning~\cite{PPO, SAC} and motors~\cite{motor1, motor2} have enabled remarkable success in robot control~\cite{Exbody2, Omnih2o}, especially in humanoid motion tracking~\cite{Any2Track, GMT, BeyondMimic}. However, most methods still rely on zero-shot real-world deployment of simulation-trained policies~\cite{Any2Track, GMT, BeyondMimic}. We argue that this paradigm has critical limitations.
First, during real-world deployment, robots inevitably encounter training-deployment gaps, such as extra physical disturbances or inaccurate simulation modeling, which severely degrade policy performance.
Second, when real-world performance is poor, adjusting simulation training parameters is highly inefficient as it relies on black-box tuning. Therefore, effectively leveraging real-world data to fine-tune simulation-trained policies represents the crucial "last mile" for achieving robust practical deployment.

In this paper, using tracking as a representative task, we focus on the real-world adaptation of humanoid tracking policies. Inspired by how humans continuously refine their physical skills, we propose \textbf{HumanoidNDM}, which abstracts the online learning paradigm into three stages: \textbf{Data Collection}, \textbf{Model Training}, and \textbf{Policy Fine-tuning}.

However, direct hardware deployment introduces stage-specific challenges across these three stages. First, during data collection, existing approaches often either employ exogenous noise injection (e.g. add Gaussian noise to action), which leads to violent joint jitter and whole-body shaking on physical robots~\cite{gSDE, Energy-Consumption}, or depend solely on environmental stochasticity for passive exploration~\cite{ASAP}, which is highly detrimental to subsequent policy fine-tuning. As a result, existing methods fail to achieve a balance between deployment feasibility and exploration effectiveness. Second, in model training, learning whole-body dynamics relies heavily on privileged information obtained from external MoCap systems~\cite{ASAP}, which is costly and hard to deploy. Lastly, during policy training, achieving efficient and stable fine-tuning under limited online data remains a problem. HumanoidNDM resolves these bottlenecks step by step.

In the \textbf{Data Collection} stage, we propose \textbf{F}requency-domain \textbf{E}xploration via \textbf{N}eighborhood \textbf{C}overage \textbf{E}stimation (\textbf{FENCE}). To strike a balance between deployment feasibility and exploration effectiveness, FENCE injects low-frequency noise into the reference motions, leveraging inherent policy robustness for smoother and more stable exploration. We further introduce an adaptive mechanism that dynamically scales the perturbation intensity based on policy neighborhood coverage to enhance exploration efficiency.

In the \textbf{Model Training} stage, while optimizing models in Euclidean space has become a consensus in existing works~\cite{SuperTrack, ControlVAE}, doing so for humanoid robots typically relies on privileged MoCap data. To bypass this, we extend traditional single-frame models with historical observations and differentiable forward kinematics (FK), enabling direct optimization in Euclidean space via purely proprioceptive data, which significantly improves the model's reliability and robustness.

Finally, in the \textbf{Policy Fine-tuning} stage, we adopt a supervised learning paradigm that updates the policy via model gradients~\cite{Dreamer-v1, Dreamer-v2, Dreamer-v3, SHAC}. Unlike prior methods constrained by privileged information and short rollouts that cause greedy updates, our proprioceptive model supports training over longer rollout horizons, thereby achieving efficient optimization across the entire trajectory.

Simulation experiments show HumanoidNDM achieves significant improvement under various training-deployment gaps, surpassing baselines across various metrics. Real-world experiments demonstrate that HumanoidNDM achieves efficient adaptation against challenging disturbances.

The main contributions of this paper are summarized as follows:
\begin{itemize}
    \item We propose FENCE, a deployment-feasible real-world exploration method using low-frequency reference perturbations, along with an adaptive intensity mechanism.
    \item We propose methods for training MoCap-free robust neural dynamics models, and subsequently leverage the model's differentiability to enable efficient policy fine-tuning.
    \item We deploy the first online fine-tuning framework for humanoid tracking policies, validating its ability to adapt to challenging disturbances.
\end{itemize}

\section{Related Work}

\subsection{Learning Robotics Control Policies with Real-World Data}

Existing approaches for bridging the training-deployment gap can be broadly divided into simulator-based adaptation and real-world dynamics learning. Simulator-based adaptation methods align simulation with reality through system identification~\cite{HALO, SPI-Active}, transition compensation~\cite{ASAP, UAN}, or trajectory-level alignment~\cite{DexNDM}. However, these approaches typically carry high OOD risks, and suffer from slow training speeds, which precludes online policy updates.

In contrast, real-world training directly leverages deployment data collected from physical robots to update policies. Model-free RL directly applies off-policy RL methods in real world~\cite{Walk-in-Park, Keep-on-Learning, Grow-your-Limits, Safe-Reinforcement-Learning, Robot-Trains-Robot, Towards-Real-Robot-Learning}, but remains unstable under limited real-world data and complex humanoid dynamics~\cite{Simulation-Distillation, RISE}. Model-based methods improve data efficiency by learning neural dynamics from real-world interactions, either for value-based policy updates~\cite{RWM1, RWM2, LIFT, Learning-to-Walk} or for supervised policy optimization through analytical model gradients~\cite{Without-Reinforcement}. HumanoidNDM adopts the latter paradigm for its data efficiency in real-world online adaptation.

\subsection{Real-world Exploration in Legged Robots}

Exploration is essential for policy improvement~\cite{MoGE, Exploration-Survey, Safe-Exploration}, yet simulation-oriented strategies such as uncorrelated Gaussian action noise can induce unsafe high-frequency torques and destructive vibrations on hardware~\cite{gSDE, Energy-Consumption}. To avoid this, some methods rely on passive environmental stochasticity~\cite{LIFT, ASAP}, but this limits state-action coverage and further optimization. Some other methods try to introduce temporally correlated or trajectory-level perturbations~\cite{DDPG, Pink-Noise, CoLA-Flow-Policy, Exploration-in-DRL-Survey} for smooth exploration, but remain tuning-sensitive and weakly validated on real robots. While guided exploration methods utilizing simulation priors~\cite{SGFT}, Fisher information~\cite{SPI-Active}, or adaptive noise envelopes~\cite{APRL} attempt to realize targeted and controllable exploration, they still struggle with actual deployment feasibility on hardware.

To realize deployment-viable and efficient exploration, we propose FENCE, which injects structured low-frequency perturbations with an adaptive intensity into reference trajectories. Table~\ref{tab:exploration_comparison} further compares FENCE with representative exploration strategies.

\begin{table}[htbp]
\centering
\caption{Comparison of exploration strategies. ``$\sim$'' denotes partial or conditional capability.}
\label{tab:exploration_comparison}
\resizebox{\linewidth}{!}{
\begin{tabular}{lccc}
\toprule
\textbf{Method} & \textbf{Deployment Feasibility} & \textbf{Policy-Relevant Coverage} & \textbf{Adaptive Intensity} \\
\midrule
Uncorrelated Action Noise~\cite{Reinforcement-Learning, Simple-Statistical} & $\times$ & $\checkmark$ & $\times$ \\
Observation Perturbation~\cite{Simple-Noisy-Environment, Neighboring-State-based-Exploration} & $\times$ & $\times$ & $\times$ \\
Passive Dynamic Exploration~\cite{LIFT, ASAP} & $\checkmark$ & $\times$ & $\times$ \\
Temporally Correlated Noise~\cite{DDPG, Pink-Noise} & $\sim$ & $\checkmark$ & $\times$ \\
\textbf{Ours (FENCE)} & $\textbf{\checkmark}$ & $\textbf{\checkmark}$ & $\textbf{\checkmark}$ \\
\bottomrule
\end{tabular}
}
\end{table}
\section{Preliminaries}

The motion tracking task requires the robot to replicate a reference trajectory $\mathcal{T} = \{\hat{q}_t, \hat{\dot{q}}_t\}_{t=1}^T$, where $\hat{q}_t$ and $\hat{\dot{q}}_t$ denote the target joint positions (\textit{qpos}) and velocities (\textit{qvel}). At each step $t$, the locomotion policy $\pi(a_t | o_t)$ receives an observation $o_t$ comprising the proprioceptive state ($q_t$, $\dot{q}_t$, gravity vector $\mathbf{g}_t$, and angular velocity $\boldsymbol{\omega}_t$), previous action $a_{t-1} $, and future reference states from $\mathcal{T}$. The policy outputs action $a_t \in [-1, 1]$. While $\pi$ is agnostic to the training algorithm, our base policy is trained using PPO~\cite{PPO} in the highly parallelized MuJoCo-JAX (MJX)~\cite{mjx}.

Policy output $a_t$ is scaled by a factor $c$ and added to the reference position to get position target $q^{des}_t$: $q^{des}_t = c \cdot a_t + \hat{q}_t$.
A proportional-derivative (PD) controller then computes the joint torques $\tau_t$:
$$\tau_t = K_p (q^{des}_t - q_t) - K_d \dot{q}_t$$
where $K_p$ and $K_d$ are the PD gains. Finally, $\tau_t$ is clipped to the hardware limits and executed.
\section{Method}

This section details HumanoidNDM, which abstracts online real-world adaptation into three stages. We then introduce its three core components: the deployment-viable data collection mechanism FENCE, MoCap-free world model learning, and efficient model-gradient-guided policy fine-tuning.

\begin{figure}[htbp]
    \centering
    \includegraphics[width=0.90\linewidth]{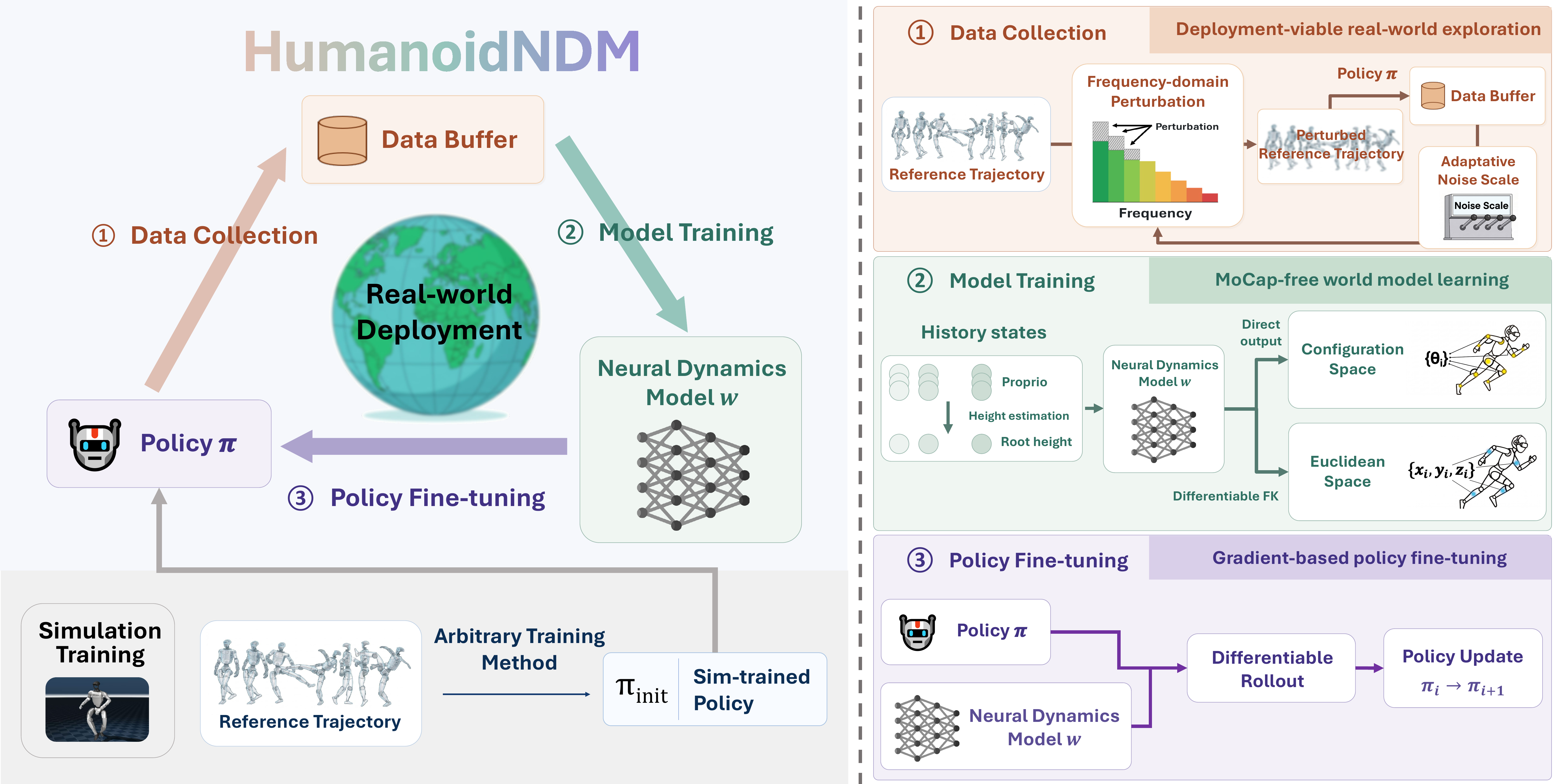}
    \caption{Pipeline of HumanoidNDM. Left: HumanoidNDM uses a simulation-trained policy to collect real-world data for online fine-tuning. Each iteration comprises three stages: \textbf{Data Collection}, \textbf{Model Training}, and \textbf{Policy Fine-tuning}. Right: The key components of each stage.}
    \label{fig:pipeline}
    \vspace{-1em}
\end{figure}

\subsection{Deployment-viable Real-world Exploration with Frequency-Domain Perturbations}

This section introduces FENCE, a data collection mechanism that combines frequency-domain perturbations on reference trajectories with dynamic intensity adjustment.

\paragraph{Frequency-Domain Perturbation on Reference Trajectory}
We propose a low-frequency perturbation method applied to the reference trajectory for real-world exploration inspired by human behavior. When exploring new environments, humans smoothly adjust desired poses (reference motion) instead of injecting high-frequency jitter. By adding low-frequency noise to reference trajectory, our method enables stable exploration that exploits the policy's closed-loop robustness. 

Formally, let the reference joint position sequence be $\mathbf{Q} = [\hat{q}_1, \dots, \hat{q}_T]^\top \in \mathbb{R}^{T \times D}$. We first apply the Discrete Fourier Transform to obtain the frequency-domain spectrum $\tilde{\mathbf{Q}}$:
\begin{equation}
    \tilde{\mathbf{Q}}_k = \sum_{t=1}^T \hat{q}_t e^{-i 2\pi k (t-1) / T},
\end{equation}
where $k \in \{0, 1, \dots, T-1\}$ is the frequency index. To introduce smooth deviations, we construct a frequency-domain noise matrix $\mathbf{Z} \in \mathbb{C}^{T \times D}$, where $\mathbf{Z}_0$ is set to $\mathbf{0}$ to prevent posture collapse. For low-frequency components $0 < k \leq k_c$, we sample complex Gaussian noise $\mathbf{Z}_k \sim \mathcal{CN}(0, \sigma^2 \mathbf{I})$, where $\sigma$ dictates the perturbation intensity, and set \(Z_{T-k}=Z_k^*\) to preserve conjugate symmetry. All remaining frequency components are set to zero. The perturbed reference trajectory $\mathbf{Q}'$ is then recovered via the Inverse DFT:
\begin{equation}
\mathbf{Q}' = \mathcal{F}^{-1}(\tilde{\mathbf{Q}} + \mathbf{Z}).
\end{equation}
Other reference states (e.g., joint velocities) are subsequently re-derived from $\mathbf{Q}'$ to maintain kinematic consistency. During rollouts, the policy $\pi$ tracks $\mathbf{Q}'$ instead of $\mathbf{Q}$.

\paragraph{Adaptive Perturbation Intensity via Policy Neighborhood Coverage}

In model-based training, distribution shifts frequently lead to objective mismatch and model exploitation~\cite{distribution-shift1, distribution-shift2}, making it critical for data collection to cover the state-action subspace visited by the policy~\cite{data-coverage1, data-coverage2}. To address this, we introduce an adaptive perturbation intensity mechanism. 
To provide a quantitative criterion for adjustment, we extend traditional cell-based state space discretization~\cite{pnc} to our continuous control setting, defining the Policy Neighborhood Coverage $C$.
Specifically, let $x_t = [s_t^\top, a_t^\top]^\top$ denote the concatenated state and action at step $t$, obtained by executing the latest policy without perturbation. Around each point $x_t$, we define a local neighborhood of radius $R$, uniformly partitioned into $N_r$ radial bins and $N_d$ directional bins. By aggregating the binary occupancy of the buffer $\mathcal{B}$ across all spatial bins along the entire trajectory, it is formulated as:
\begin{equation}
    C = \frac{1}{T N_r N_d} \sum_{t=1}^{T-1} \sum_{j=1}^{N_r} \sum_{k=1}^{N_d} \mathbb{I}\left( \mathcal{B} \cap B_t^{(j,k)} \neq \emptyset \right),
\end{equation}
where $B_t^{(j,k)}$ denotes the spatial bin at the $j$-th radial and $k$-th directional index. 
Intuitively, $C$ measures the uniformity and density of exploration around the current policy. To implement online intensity adjustment, we discretize perturbation intensity into $K$ candidate levels $\Sigma = \{\sigma_1, \dots, \sigma_K\}$, and utilize Multi-Armed Bandit (MAB) to adaptively select the optimal perturbation level.

\subsection{MoCap-Free Neural Dynamics Model Training with Dual-Space Objective}

Although neural networks are commonly used to model environment dynamics~\cite{Dreamer-v2, Neural-Robot-Dynamics, Contact-Aware-Neural-Dynamics}, learning an accurate world model for humanoid motion tracking is challenging in a MoCap-free setting. During deployment, the robot can only access onboard proprioceptive signals, while external full-body pose measurements are unavailable. A straightforward solution is to optimize the dynamics model in configuration space. However, for humanoid robots, small configuration-space errors may cause large deviations in link positions, end-effector poses, and contact-related body configurations, which can severely degrade downstream policy fine-tuning~\cite{ControlVAE, SuperTrack}.

Following physics-based animation works~\cite{ControlVAE, SuperTrack, ASAP}, we introduce dual-space objectives into neural dynamics model training. The key is to preserve a proprioceptive-only interface while using differentiable Forward Kinematics (FK) for Euclidean-space supervision.

We define the proprioceptive state as $s_t = \{q_t, \dot{q}_t, \mathbf{g}_t, \boldsymbol{\omega}_t, h_t\}$, where the root height $h_t$ is estimated from $q_t$ and $\mathbf{g}_t$ assuming continuous ground contact. The world model $w$ predicts the next state $\hat{s}_{t+1}$ by conditioning on a historical sequence of states and actions of length $H$: 
\begin{equation} 
    \hat{s}_{t+1} = w(\mathbf{s}_{t-H+1:t}, \mathbf{a}_{t-H+1:t}) .
\end{equation}
This temporal context allows the model to implicitly infer recent dynamics. Furthermore, an ensemble framework with multiple model heads is trained via bootstrap sampling to enhance robustness.

To enable end-to-end dual-space optimization, we integrate a differentiable FK model that maps predicted configuration-space states into Euclidean space, yielding 6-DoF link poses in a local gravity-aligned frame. To ensure long-term predictive accuracy, we adopt an autoregressive rollout strategy~\cite{ControlVAE, SuperTrack, RWM1, RWM2}, recursively predicting future states $\hat{s}{t+k}$ over a window of length $L_w$ by appending predictions to the input history, with the rollout loss $\mathcal{L}_{w}$ defined as
\begin{equation} 
\mathcal{L}_{w} = \frac{1}{L_{w}} \sum_{k=1}^{L_{w}} \left( \mathcal{L}_{\text{config\_err}}^{(t+k)} + \mathcal{L}_{\text{euclid\_err}}^{(t+k)} \right) .
\end{equation}
The configuration-space error $\mathcal{L}_{\text{config\_err}}$ comprises the $L_2$ distance for proprioceptive states ($q, \dot{q}, \mathbf{g}, \boldsymbol{\omega}$, and $h$), while the Euclidean-space error $\mathcal{L}_{\text{euclid\_err}}$ penalizes the positional and rotational discrepancies of all links in the local gravity-aligned frame via the differentiable FK model.

\subsection{Efficient and Stable Policy Fine-Tuning via Supervised Gradients}

With the Euclidean-space neural dynamics model $w$ established, we opt for direct optimization by backpropagating analytic gradients through the differentiable model~\cite{SuperTrack, ControlVAE, Without-Reinforcement}. This scheme is chosen for its superior data efficiency and faster convergence compared to model-based reinforcement learning~\cite{RWM1, RWM2}, which is crucial for limited real-world data.

Policy fine-tuning is analogous to the model training process. First, we adopt an autoregressive rollout strategy over a length $L_{\pi}$ to avoid greedy updates. Furthermore, we use dual-space loss as the tracking objective, which significantly enhances the expressiveness of the tracked motions~\cite{DeepMimic, BeyondMimic, GMT}. Incorporating standard regularization terms $\mathcal{L}_{\text{reg}}$, the policy fine-tuning loss $\mathcal{L}_{\pi}$ is formulated as:
\vspace{-0.5em}
\begin{equation}
    \mathcal{L}_{\pi} = \frac{1}{L_{\pi}} \sum_{k=1}^{L_{\pi}} \left( \mathcal{L}_{\text{config\_track}}^{(t+k)} + \mathcal{L}_{\text{euclid\_track}}^{(t+k)} + \mathcal{L}_{\text{reg}}^{(t+k)} \right).
\end{equation}

\section{Experiments}
In this section, we first present baseline comparisons and detailed ablation studies in simulation to validate our core algorithmic designs.
Then we deploy HumanoidNDM on Unitree G1 to demonstrate its capability in closing the training-deployment gap under diverse physical disturbances.

\subsection{Evaluation in Simulation}

\paragraph{Experiments Setup}

The experiments are conducted in MuJoCo using a Unitree G1 model. We designed five distinct motions and trained a deployable initial policy. The five motions include planar and vertical motions, as well as actions emphasizing high dynamics and extreme balance. 
To systematically simulate training-deployment gap, we introduce three distinct types of disturbance. (1) \textbf{inertial variations}: adding mass and shifting the CoM of specific links. (2) \textbf{contact variations}: altering ground friction coefficients. (3) \textbf{actuation variations}: scaling the $K_p$ of specific joints.

\paragraph{Baselines}
We compare HumanoidNDM with: (a) \textbf{PPO w/ Disturb.}, trained directly using PPO with the exact deployment disturbances; 
(b) \textbf{Direct Transfer}, where policies trained without disturbance are directly deployed to perturbed environments; 
(c) \textbf{ASAP}~\cite{ASAP}, which aligns sim-to-real dynamics using a delta action model trained on real-world transitions; 
and (d) \textbf{RWM}~\cite{RWM2}, another model-based method that differs from HumanoidNDM across all three stages. 
HumanoidNDM, ASAP, and RWM are trained using 200 trajectories collected from the deployment environment.

\paragraph{Metrics} 
We evaluate tracking performance using 4 key metrics, all computed relative to the reference motion: (a) $E_{\text{link\_pos}}$, the mean position error of all links in a local gravity-aligned frame; (b) $E_{\text{link\_rot}}$, the mean rotation error of all links in a local gravity-aligned frame; (c) $E_{\text{qpos}}$, the tracking error of all joints; and (d) $E_{\text{key\_qpos}}$, the joint error of the most expressive, task-specific joints.

To evaluate the stability of different exploration methods during data collection, we introduce two additional metrics: (a) $S_{\text{qpos}}$, the full-body qpos smoothness, which measures the joint jitter; and (b) $S_{\text{roll-pitch}}$, the roll-pitch smoothness, which evaluates the whole-body stability of the robot.

\subsubsection{Comparison with Baselines}

\begin{table*}[t]
\centering
\caption{Comparison of HumanoidNDM and baselines across 5 random trials (mean$\pm$std).}
\label{tab:exp_summary}
\resizebox{0.98\textwidth}{!}{
\begin{tabular}{ll cccc cccc cccc}
\toprule
\multicolumn{2}{c}{Disturbance} & \multicolumn{4}{c}{\cellcolor{red!10}Extra Weight} & \multicolumn{4}{c}{\cellcolor{green!10}Scaled KP} & \multicolumn{4}{c}{\cellcolor{blue!10}Small Friction} \\
\cmidrule(lr){3-6} \cmidrule(lr){7-10} \cmidrule(lr){11-14}
Trajectory & Method & $E_{\text{link\_pos}}$ & $E_{\text{link\_rot}}$ & $E_{\text{qpos}}$ & $E_{\text{key\_qpos}}$ & $E_{\text{link\_pos}}$ & $E_{\text{link\_rot}}$ & $E_{\text{qpos}}$ & $E_{\text{key\_qpos}}$ & $E_{\text{link\_pos}}$ & $E_{\text{link\_rot}}$ & $E_{\text{qpos}}$ & $E_{\text{key\_qpos}}$ \\
\midrule
\multirow{5}{*}{Walk}
& \cellcolor{gray!15}PPO w/ Disturb. & \cellcolor{gray!15}3.11 & \cellcolor{gray!15}0.28 & \cellcolor{gray!15}2.01 & \cellcolor{gray!15}3.83 & \cellcolor{gray!15}2.74 & \cellcolor{gray!15}0.28 & \cellcolor{gray!15}1.70 & \cellcolor{gray!15}1.94 & \cellcolor{gray!15}2.45 & \cellcolor{gray!15}0.24 & \cellcolor{gray!15}1.46 & \cellcolor{gray!15}1.70 \\
\cmidrule(lr){2-14}
& Direct Transfer & 5.33$\pm$0.00 & 0.55$\pm$0.00 & 3.29$\pm$0.00 & 3.81$\pm$0.00 & 4.92$\pm$0.00 & 0.56$\pm$0.00 & 3.20$\pm$0.00 & 3.80$\pm$0.00 & 4.64$\pm$0.00 & 0.49$\pm$0.00 & 3.06$\pm$0.00 & 3.84$\pm$0.00 \\
& ASAP             & 4.09$\pm$0.20 & 0.41$\pm$0.03 & 3.30$\pm$0.09 & 2.93$\pm$0.20 & \textbf{2.83$\pm$0.12} & 0.47$\pm$0.02 & 2.67$\pm$0.05 & 2.02$\pm$0.15 & \textbf{2.49$\pm$0.08} & 0.31$\pm$0.01 & 2.81$\pm$0.01 & 2.35$\pm$0.09 \\
& RWM              & 4.71$\pm$0.15 & 0.40$\pm$0.03 & 3.15$\pm$0.07 & 2.91$\pm$0.11 & 4.71$\pm$0.28 & 0.32$\pm$0.03 & 2.55$\pm$0.07 & 2.98$\pm$0.38 & 4.37$\pm$0.64 & 0.36$\pm$0.14 & 2.11$\pm$0.18 & 2.33$\pm$0.19 \\
& \textcolor{red}{HumanoidNDM} & \textbf{3.59$\pm$0.24} & \textbf{0.27$\pm$0.03} & \textbf{2.15$\pm$0.07} & \textbf{2.56$\pm$0.23} & 2.90$\pm$0.13 & \textbf{0.22$\pm$0.02} & \textbf{1.80$\pm$0.04} & \textbf{1.86$\pm$0.13} & \textbf{2.49$\pm$0.09} & \textbf{0.13$\pm$0.01} & \textbf{1.54$\pm$0.01} & \textbf{1.55$\pm$0.08} \\
\midrule
\multirow{5}{*}{Balance}
& \cellcolor{gray!15}PPO w/ Disturb. & \cellcolor{gray!15}2.66 & \cellcolor{gray!15}0.56 & \cellcolor{gray!15}3.61 & \cellcolor{gray!15}1.90 & \cellcolor{gray!15}2.46 & \cellcolor{gray!15}0.39 & \cellcolor{gray!15}3.39 & \cellcolor{gray!15}2.01 & \cellcolor{gray!15}2.52 & \cellcolor{gray!15}0.33 & \cellcolor{gray!15}3.04 & \cellcolor{gray!15}1.78 \\
\cmidrule(lr){2-14}
& Direct Transfer & 4.32$\pm$0.00 & 0.69$\pm$0.00 & 3.55$\pm$0.00 & 4.24$\pm$0.01 & 5.88$\pm$0.02 & 0.76$\pm$0.00 & 3.32$\pm$0.00 & 6.72$\pm$0.03 & 6.04$\pm$0.02 & 0.72$\pm$0.00 & 3.28$\pm$0.00 & 7.71$\pm$0.02 \\
& ASAP             & \textbf{3.40$\pm$0.14} & 0.70$\pm$0.04 & 3.89$\pm$0.22 & 2.62$\pm$0.19 & \textbf{2.61$\pm$0.22} & \textbf{0.33$\pm$0.02} & 3.35$\pm$0.11 & \textbf{3.18$\pm$0.55} & \textbf{2.66$\pm$0.21} & 0.40$\pm$0.02 & 3.36$\pm$0.08 & \textbf{1.90$\pm$0.13} \\
& RWM              & 4.00$\pm$0.10 & 0.52$\pm$0.03 & \textbf{3.32$\pm$0.18} & 3.62$\pm$0.29 & 5.03$\pm$0.43 & 0.42$\pm$0.02 & 3.24$\pm$0.08 & 3.77$\pm$0.29 & 4.77$\pm$0.41 & 0.36$\pm$0.04 & 3.07$\pm$0.09 & 3.25$\pm$0.34 \\
& \textcolor{red}{HumanoidNDM} & 3.51$\pm$0.13 & \textbf{0.46$\pm$0.05} & 3.45$\pm$0.16 & \textbf{2.32$\pm$0.16} & 3.82$\pm$0.27 & 0.49$\pm$0.02 & \textbf{3.21$\pm$0.12} & 4.07$\pm$0.49 & 3.24$\pm$0.15 & \textbf{0.35$\pm$0.02} & \textbf{2.94$\pm$0.06} & 2.77$\pm$0.12 \\
\midrule
\multirow{5}{*}{Squat}
& \cellcolor{gray!15}PPO w/ Disturb. & \cellcolor{gray!15}5.59 & \cellcolor{gray!15}1.04 & \cellcolor{gray!15}2.25 & \cellcolor{gray!15}5.50 & \cellcolor{gray!15}2.46 & \cellcolor{gray!15}0.17 & \cellcolor{gray!15}2.05 & \cellcolor{gray!15}5.22 & \cellcolor{gray!15}1.74 & \cellcolor{gray!15}0.17 & \cellcolor{gray!15}1.94 & \cellcolor{gray!15}4.67 \\
\cmidrule(lr){2-14}
& Direct Transfer & 8.97$\pm$0.00 & 1.50$\pm$0.00 & 3.39$\pm$0.00 & 6.20$\pm$0.00 & 5.97$\pm$0.01 & 0.95$\pm$0.00 & 2.66$\pm$0.00 & 6.99$\pm$0.02 & 2.13$\pm$0.00 & 0.10$\pm$0.00 & 1.70$\pm$0.00 & 4.00$\pm$0.00 \\
& ASAP             & 7.55$\pm$0.26 & 1.20$\pm$0.10 & 3.23$\pm$0.11 & 5.88$\pm$0.68 & \textbf{3.86$\pm$0.23} & 0.38$\pm$0.04 & 2.56$\pm$0.09 & 5.82$\pm$0.22 & \textbf{1.86$\pm$0.27} & \textbf{0.05$\pm$0.02} & 1.63$\pm$0.11 & 4.02$\pm$0.51 \\
& RWM              & 7.20$\pm$0.20 & 0.97$\pm$0.11 & 2.91$\pm$0.15 & 5.38$\pm$0.57 & 4.75$\pm$0.56 & 0.54$\pm$0.16 & 2.52$\pm$0.23 & 5.98$\pm$0.51 & 2.56$\pm$0.00 & 0.11$\pm$0.00 & 1.94$\pm$0.19 & 4.51$\pm$0.58 \\
& \textcolor{red}{HumanoidNDM} & \textbf{5.98$\pm$0.30} & \textbf{0.78$\pm$0.07} & \textbf{2.44$\pm$0.08} & \textbf{4.65$\pm$0.46} & 4.44$\pm$0.21 & \textbf{0.34$\pm$0.04} & \textbf{2.39$\pm$0.08} & \textbf{4.76$\pm$0.17} & 2.09$\pm$0.24 & 0.08$\pm$0.02 & \textbf{1.52$\pm$0.08} & \textbf{3.93$\pm$0.44} \\
\midrule
\multirow{5}{*}{Swing}
& \cellcolor{gray!15}PPO w/ Disturb. & \cellcolor{gray!15}2.06 & \cellcolor{gray!15}0.21 & \cellcolor{gray!15}1.50 & \cellcolor{gray!15}1.48 & \cellcolor{gray!15}1.96 & \cellcolor{gray!15}0.28 & \cellcolor{gray!15}2.28 & \cellcolor{gray!15}2.90 & \cellcolor{gray!15}1.07 & \cellcolor{gray!15}0.10 & \cellcolor{gray!15}0.87 & \cellcolor{gray!15}0.83 \\
\cmidrule(lr){2-14}
& Direct Transfer & 4.52$\pm$0.00 & 0.70$\pm$0.00 & 3.01$\pm$0.00 & 2.81$\pm$0.00 & 4.35$\pm$0.00 & 0.86$\pm$0.00 & 2.85$\pm$0.00 & 6.26$\pm$0.00 & 2.16$\pm$0.00 & 0.14$\pm$0.00 & 2.39$\pm$0.00 & 0.90$\pm$0.00 \\
& ASAP             & 4.15$\pm$0.51 & 0.68$\pm$0.06 & 2.71$\pm$0.31 & 2.54$\pm$0.21 & 3.96$\pm$0.30 & 0.79$\pm$0.01 & 3.31$\pm$0.09 & 6.30$\pm$0.18 & 2.02$\pm$0.03 & 0.12$\pm$0.00 & 2.22$\pm$0.05 & 0.82$\pm$0.07 \\
& RWM              & 3.79$\pm$0.24 & 0.40$\pm$0.08 & 2.01$\pm$0.04 & 1.85$\pm$0.22 & 3.33$\pm$0.16 & 0.51$\pm$0.05 & 2.14$\pm$0.07 & 4.79$\pm$0.23 & 1.28$\pm$0.03 & \textbf{0.03$\pm$0.00} & 1.19$\pm$0.05 & \textbf{0.50$\pm$0.03} \\
& \textcolor{red}{HumanoidNDM} & \textbf{2.40$\pm$0.37} & \textbf{0.17$\pm$0.06} & \textbf{1.64$\pm$0.28} & \textbf{1.38$\pm$0.25} & \textbf{2.06$\pm$0.28} & \textbf{0.15$\pm$0.01} & \textbf{1.52$\pm$0.09} & \textbf{3.09$\pm$0.19} & \textbf{1.18$\pm$0.03} & \textbf{0.03$\pm$0.00} & \textbf{0.95$\pm$0.04} & 0.63$\pm$0.06 \\
\midrule
\multirow{5}{*}{Stride}
& \cellcolor{gray!15}PPO w/ Disturb. & \cellcolor{gray!15}3.35 & \cellcolor{gray!15}0.50 & \cellcolor{gray!15}2.83 & \cellcolor{gray!15}5.75 & \cellcolor{gray!15}4.06 & \cellcolor{gray!15}0.63 & \cellcolor{gray!15}3.76 & \cellcolor{gray!15}7.12 & \cellcolor{gray!15}3.92 & \cellcolor{gray!15}0.62 & \cellcolor{gray!15}2.89 & \cellcolor{gray!15}6.24 \\
\cmidrule(lr){2-14}
& Direct Transfer & 4.24$\pm$0.01 & 0.66$\pm$0.00 & 3.72$\pm$0.00 & 6.96$\pm$0.02 & 5.64$\pm$0.00 & 1.11$\pm$0.00 & 4.31$\pm$0.00 & 7.27$\pm$0.01 & 4.63$\pm$0.01 & 0.78$\pm$0.00 & 3.94$\pm$0.00 & 6.83$\pm$0.02 \\
& ASAP             & 4.22$\pm$0.07 & 0.64$\pm$0.02 & 3.36$\pm$0.04 & 7.03$\pm$0.29 & 5.36$\pm$0.24 & \textbf{0.72$\pm$0.02} & 3.83$\pm$0.10 & 7.71$\pm$0.36 & 4.16$\pm$0.06 & \textbf{0.44$\pm$0.02} & \textbf{3.19$\pm$0.04} & \textbf{6.02$\pm$0.06} \\
& RWM              & 5.10$\pm$0.01 & 0.53$\pm$0.16 & \textbf{3.06$\pm$0.14} & \textbf{5.40$\pm$0.16} & 6.07$\pm$0.24 & 0.77$\pm$0.11 & \textbf{3.36$\pm$0.07} & \textbf{5.98$\pm$0.37} & 4.58$\pm$0.03 & 0.63$\pm$0.03 & 3.62$\pm$0.08 & 6.76$\pm$0.20 \\
& \textcolor{red}{HumanoidNDM} & \textbf{4.15$\pm$0.06} & \textbf{0.52$\pm$0.02} & 3.42$\pm$0.04 & 6.25$\pm$0.24 & \textbf{5.35$\pm$0.18} & 0.96$\pm$0.03 & 4.05$\pm$0.07 & 6.88$\pm$0.26 & \textbf{4.01$\pm$0.05} & 0.72$\pm$0.02 & 3.73$\pm$0.04 & 6.73$\pm$0.04 \\
\bottomrule
\end{tabular}
}
\vspace{-1.0em}
\end{table*}

As shown in Table~\ref{tab:exp_summary}, HumanoidNDM demonstrates robust superiority, achieving the best performance in 40 out of 60 evaluated metrics. HumanoidNDM shows a clear advantage in the walk, squat, and swing motions, while its performance is comparable to other methods in the balance and stride motions. Since the training-deployment gap for balance and stride is smaller than that of the other three motions, the higher sample efficiency of HumanoidNDM compared to RWM is not fully highlighted, and ASAP's delta action model is also relatively easier to bridge the dynamic gap between the training and deployment environments.

Across all 15 settings, HumanoidNDM yields the most significant improvements in Euclidean space metrics. Normalized against the Direct Transfer baseline, our method reduces the $E_{\text{link\_pos}}$ and $E_{\text{link\_rot}}$ by 30\% and 47\%, respectively. In contrast, ASAP achieves reductions of 23\% and 27\%, while RWM reduces them by 9\% and 35\%. We point out that this improvement benefits from our explicit emphasis on Euclidean space optimization during model training and policy fine-tuning.

Finally, HumanoidNDM performs on par with the PPO w/ Disturb., which is directly trained using PPO in environments with the exact deployment disturbances. Matching this baseline demonstrates that with only 200 deployment trajectories, HumanoidNDM can model the target dynamics with high fidelity, effectively bridging the training-deployment gap without extensive interactions.

\subsubsection{Ablation Study}

\paragraph{Comparison of Exploration Strategies}

\begin{wrapfigure}{r}{0.60\textwidth}
    \vspace{-0.3cm}
    \centering
    \begin{subfigure}[b]{0.48\linewidth}
        \centering
        \includegraphics[width=\textwidth]{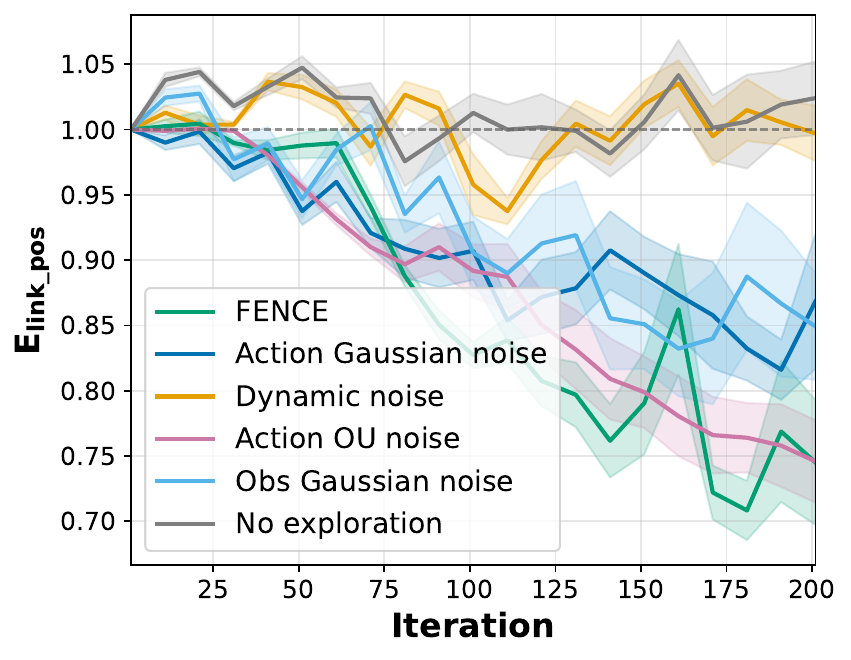}
    \end{subfigure}
    \hspace{2pt}
    \begin{subfigure}[b]{0.48\linewidth}
        \centering
        \includegraphics[width=\textwidth]{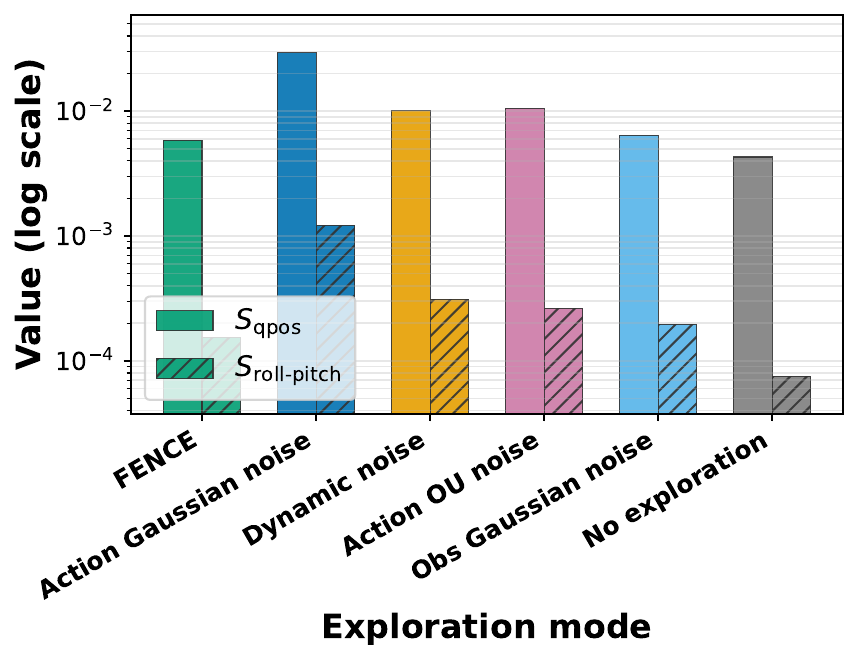}
    \end{subfigure}
    \caption{Comparison between exploration methods.}
    \label{fig:exploration}
    \vspace{-0.4cm}
\end{wrapfigure}

We replace FENCE in the data collection phase of HumanoidNDM with other classic methods.
Figure~\ref{fig:exploration} shows normalized $E_{\text{link\_pos}}$ performance curves during fine-tuning (left) and average collection smoothness (right). Although both FENCE and Action OU noise achieve the best fine-tuning results, FENCE significantly outperforms Action OU in $S_{\text{qpos}}$ and $S_{\text{roll-pitch}}$, achieving stability second only to the no exploration baseline. This proves that FENCE effectively improves data collection stability while ensuring sufficient state-action coverage to support policy fine-tuning.

\begin{wraptable}{r}{0.60\textwidth}
\vspace{-0.5em}
\centering
\caption{Effectiveness of FENCE on baselines (mean$\pm$std).}
\label{tab:asap_rwm_fence}
\vspace{0.2cm}
\resizebox{\linewidth}{!}{
\begin{tabular}{lcccc}
\toprule
\multirow{2}{*}{\textbf{Method}} 
& \multicolumn{2}{c}{\textbf{Policy Evaluation Performance}} 
& \multicolumn{2}{c}{\textbf{Data Collection Performance}} \\
\cmidrule(lr){2-3} \cmidrule(lr){4-5}
& $E_{\text{link\_pos}} \downarrow$ 
& $E_{\text{qpos}} \downarrow$ 
& $S_{\text{qpos}}$ \tiny{$(\times 10^{-3})$} $\downarrow$ 
& $S_{\text{roll-pitch}}$ \tiny{$(\times 10^{-4})$} $\downarrow$ \\
\midrule
ASAP            
& 3.68$\pm$0.20 
& \textbf{3.03$\pm$0.10} 
& \textbf{4.79$\pm$6.47} 
& \textbf{0.95$\pm$3.06} \\
ASAP w/ FENCE   
& \textbf{3.53$\pm$0.18} 
& 3.13$\pm$0.08 
& 5.84$\pm$6.27 
& 1.23$\pm$4.25 \\
\midrule \midrule
RWM             
& 4.42$\pm$0.23 
& \textbf{2.68$\pm$0.11} 
& 27.44$\pm$13.82 
& 11.06$\pm$9.73 \\
RWM w/ FENCE    
& \textbf{4.23$\pm$0.35} 
& 2.72$\pm$0.13 
& \textbf{5.43$\pm$5.12} 
& \textbf{1.48$\pm$4.19} \\
\bottomrule
\end{tabular}
}
\vspace{-0.5em}
\end{wraptable}
To demonstrate the generalizability of FENCE across different training frameworks, we apply it to ASAP and RWM baselines. The original ASAP implementation relies on random resets for data diversity, which is impractical in real world. For the default RWM, we use action Gaussian noise for data collection. As shown in Table~\ref{tab:asap_rwm_fence}, replacing these default strategies with FENCE maintains the fine-tuning performance. Furthermore, FENCE achieves data collection stability comparable to the impractical random reset approach, while significantly outperforming action Gaussian noise.

\paragraph{Dual Space Objective vs. Configuration-space-only Objective}

\begin{figure}[htbp]
    \centering
    \begin{subfigure}[b]{0.22\textwidth}
        \centering
        \includegraphics[width=\textwidth]{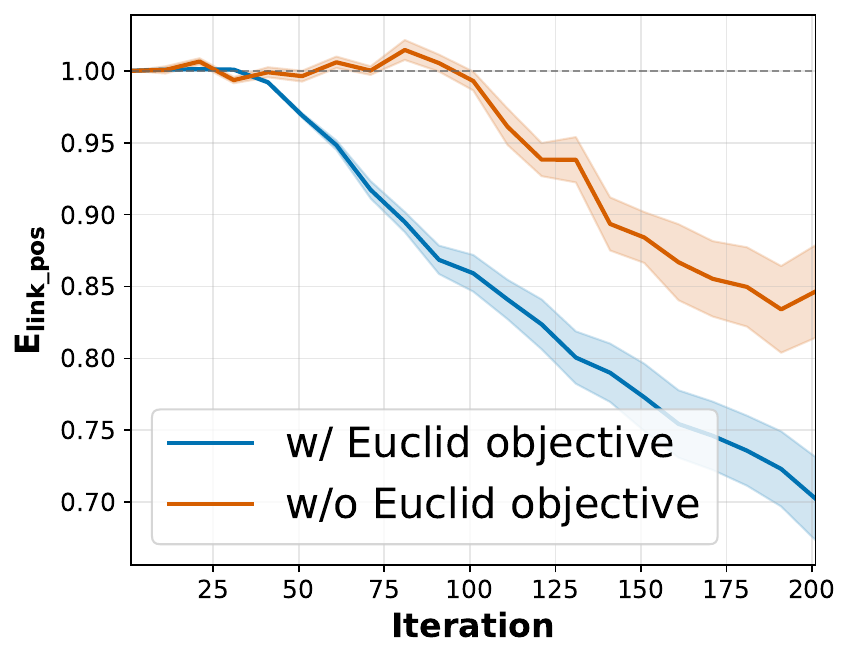}
    \end{subfigure}
    \hfill
    \begin{subfigure}[b]{0.22\textwidth}
        \centering
        \includegraphics[width=\textwidth]{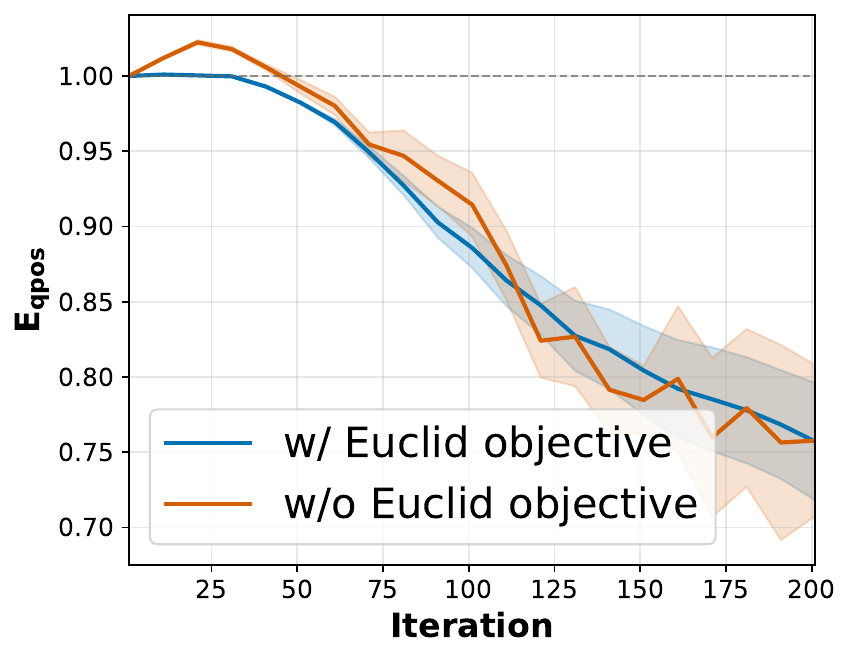}
    \end{subfigure}
    \hspace{20pt}
    \begin{subfigure}[b]{0.22\textwidth}
        \centering
        \includegraphics[width=\textwidth]{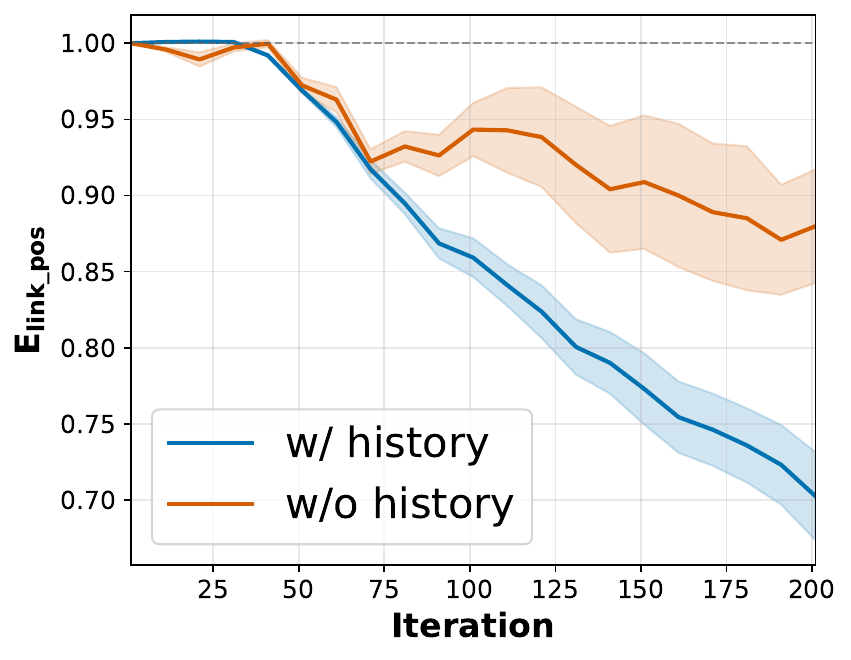}
    \end{subfigure}
    \hfill
    \begin{subfigure}[b]{0.22\textwidth}
        \centering
        \includegraphics[width=\textwidth]{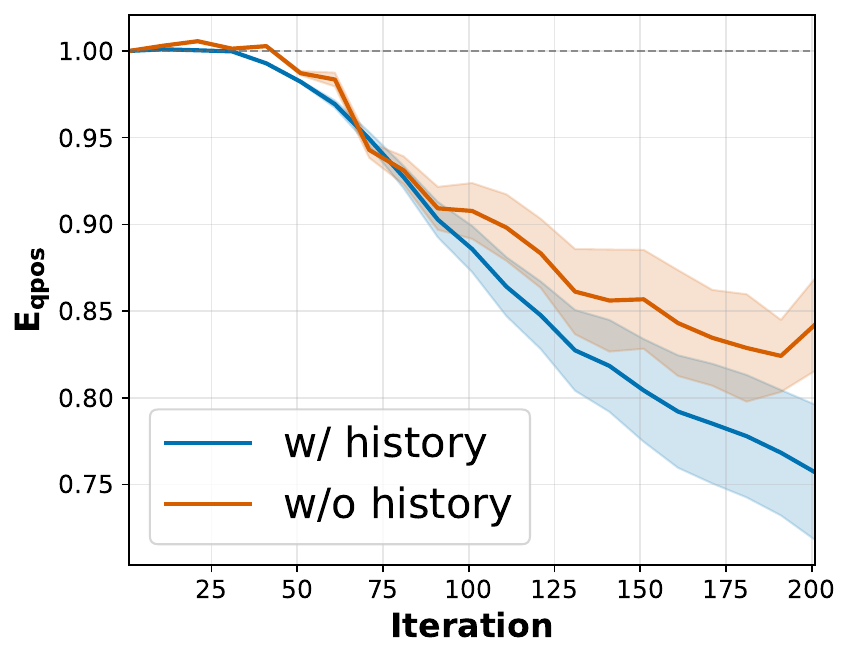}
    \end{subfigure}
    
    \caption{Left: Training w/ and w/o Euclidean-space objective. Right: Model w/ and w/o history.}
    \label{fig:ablation}
\vspace{-0.5em}
\end{figure}

The two leftmost plots in Figure~\ref{fig:ablation} compare optimizing with dual-space and configuration-space-only objectives. As shown, optimizing solely in the configuration space performs significantly worse than the dual-space approach, degrading both the Euclidean ($E_{\text{link\_pos}}$) and configuration ($E_{\text{qpos}}$) metrics.

\paragraph{History-based World Model v.s. Single Frame World Model}

The right plots of Figure~\ref{fig:ablation} compare the history-based neural dynamics model against a single-frame variant. We can see that replacing the traditional single-frame model with history-based architecture effectively enhances both training stability and overall tracking performance.

\subsection{Evaluation in Real-World}

\paragraph{Performance Improvement Under Disturbances}
As shown in Figure~\ref{fig:teaser}, we select 6 motions with corresponding disturbances for real-world experiments. These encompass highly dynamic behaviors (walking, lateral raise, arm swing) and tasks emphasizing precise control (pen capping, swallow balance, bowing). We apply weight disturbances to 3 motions and hard-to-model disturbances to the rest. We use 50 real-world trajectories to fine-tune Pen Capping, and 200 for the others.

As illustrated in Table~\ref{tab:real_performance}, HumanoidNDM achieves clear performance improvements across all 6 tasks. Each experiment is repeated 3 times. The gains are especially significant in our defined task-specific metrics, reflecting a substantial enhancement in motion completion. Notably, in the 3 settings with weight disturbances, our fine-tuned policy demonstrates performance comparable to, or even better than (e.g., Swallow Balance), the policy directly trained in simulation with explicitly modeled disturbances.

\begin{table*}[t]
  \centering
  \caption{Performance under added weight disturbances (top) and hard-to-model disturbances (bottom) (mean$\pm$std). The third column represents the task-specific metric reflecting task completeness.}
  \label{tab:real_performance}
  \resizebox{\textwidth}{!}{
    \begin{tabular}{l *{9}{c}}
      \toprule
      Setting &
      \multicolumn{3}{c}{Swallow Balance (w/ leg weights)} &
      \multicolumn{3}{c}{Arm Swing (w/ hand weights)} &
      \multicolumn{3}{c}{Pen Capping (w/ hand weight)} \\
      \cmidrule(lr){2-4} \cmidrule(lr){5-7} \cmidrule(lr){8-10}
      Method &
      $E_{\text{link\_pos}} \downarrow$ & $E_{\text{qpos}} \downarrow$ & air time (s) $\uparrow$ &
      $E_{\text{link\_pos}} \downarrow$ & $E_{\text{qpos}} \downarrow$ & arm height (m) $\uparrow$ &
      $E_{\text{link\_pos}} \downarrow$ & $E_{\text{qpos}} \downarrow$ & EE distance (cm) $\downarrow$ \\
      \midrule
      \rowcolor{gray!15}
      PPO w/ Disturb. & 4.72 $\pm$ 0.28 & \textbf{2.90 $\pm$ 0.39} & 0.40 $\pm$ 0.11 & \textbf{2.72 $\pm$ 0.01} & 2.34 $\pm$ 0.03 & \textbf{1.43 $\pm$ 0.02} & \textbf{6.58 $\pm$ 0.00} & \textbf{3.27 $\pm$ 0.00} & \textbf{0.00 $\pm$ 0.00} \\
      Direct Transfer & 5.34 $\pm$ 0.06 & 3.40 $\pm$ 0.07 & 0.00 $\pm$ 0.00 & 5.85 $\pm$ 0.02 & 3.33 $\pm$ 0.01 & 0.98 $\pm$ 0.00 & 6.96 $\pm$ 0.00 & 3.50 $\pm$ 0.00 & 3.50 $\pm$ 0.00 \\
      \textcolor{red}{HumanoidNDM} & \textbf{4.44 $\pm$ 0.13} & 2.98 $\pm$ 0.34 & \textbf{1.90 $\pm$ 0.61} & 2.98 $\pm$ 0.00 & \textbf{2.30 $\pm$ 0.01} & 1.40 $\pm$ 0.01 & 6.82 $\pm$ 0.00 & 3.36 $\pm$ 0.00 & \textbf{0.00 $\pm$ 0.00}\\
      \midrule \midrule
      Setting &
      \multicolumn{3}{c}{Lateral Raise (w/ elastic bands)} &
      \multicolumn{3}{c}{Walking (w/ water bucket)} &
      \multicolumn{3}{c}{Bowing (w/ damaged motor)} \\
      \cmidrule(lr){2-4} \cmidrule(lr){5-7} \cmidrule(lr){8-10}
      Method &
      $E_{\text{link\_pos}} \downarrow$ & $E_{\text{qpos}} \downarrow$ & arm height (m) $\uparrow$ &
      $E_{\text{link\_pos}} \downarrow$ & $E_{\text{qpos}} \downarrow$ & deviation angle ($^\circ$) $\downarrow$ &
      $E_{\text{link\_pos}} \downarrow$ & $E_{\text{qpos}} \downarrow$ & noise level (dB) $\downarrow$ \\
      \midrule
      \rowcolor{gray!15}
      PPO w/ Disturb.          & \multicolumn{3}{c}{N/A} & \multicolumn{3}{c}{N/A} & \multicolumn{3}{c}{N/A} \\
      Direct Transfer & 7.54 $\pm$ 0.58 & 5.23 $\pm$ 0.19 & 1.06 $\pm$ 0.07 & 4.12 $\pm$ 0.22 & 3.67 $\pm$ 0.31 & 38 $\pm$ 17 & 4.40 $\pm$ 0.20 & 4.45 $\pm$ 0.14 & 73 $\pm$ 8\\
      \textcolor{red}{HumanoidNDM}            & \textbf{4.65 $\pm$ 0.17} & \textbf{4.40 $\pm$ 0.03} & \textbf{1.35 $\pm$ 0.06} & \textbf{4.01 $\pm$ 0.27} & \textbf{3.26 $\pm$ 0.14} & \textbf{5 $\pm$ 3} & \textbf{4.06 $\pm$ 0.15} & \textbf{4.01 $\pm$ 0.03} & \textbf{55 $\pm$ 5}\\
      \bottomrule
    \end{tabular}
  }
  \vspace{-1em}
\end{table*}

\paragraph{Comparison of Exploration Methods}

\begin{wrapfigure}{r}{0.40\textwidth}
\vspace{-0.3cm}
\centering
\captionof{table}{Real-world comparison of exploration methods (mean$\pm$std).}
\label{tab:smoothness_comparison}
\vspace{0.1cm}
\resizebox{\linewidth}{!}{
\begin{tabular}{lcc}
\toprule
\textbf{Method} & 
$\boldsymbol{S_{\text{qpos}} \downarrow}$ & 
$\boldsymbol{S_{\text{roll-pitch}} \downarrow}$ \\
\midrule
\textbf{FENCE} & 
\textbf{0.099 $\pm$ 0.018} & 
\textbf{0.126 $\pm$ 0.023} \\
Action Gaussian Noise & 
1.275 $\pm$ 0.313 & 
1.678 $\pm$ 0.455 \\
Action OU Noise & 
0.707 $\pm$ 0.180 & 
0.739 $\pm$ 0.210 \\
Obs Gaussian Noise & 
1.151 $\pm$ 0.315 & 
1.033 $\pm$ 0.218 \\
\cellcolor{gray!15}No Noise & 
\cellcolor{gray!15}0.076 $\pm$ 0.005 & 
\cellcolor{gray!15}0.106 $\pm$ 0.008 \\
\bottomrule
\end{tabular}
}
\vspace{-0.3cm}
\end{wrapfigure}

To demonstrate the superiority of FENCE in real-world data collection, we deploy different exploration methods for the lateral raise, where parameters remain the same to simulation. As shown in Table~\ref{tab:smoothness_comparison}, the other 3 exploration methods cause obvious joint jitter and roll-pitch shaking, significantly widening the stability gap with FENCE. Meanwhile, FENCE remains nearly as stable as No Noise.

\section{Conclusion and Limitations}

We present HumanoidNDM, a framework that achieves the first online fine-tuning for humanoid tracking policies through deployment-viable exploration, Euclidean-space model training, and efficient model-gradient-guided policy fine-tuning. 
Experimental results show that HumanoidNDM effectively closes the training-deployment gap, enabling real-world adaptation to diverse disturbances. 
However, our current method is limited to time-invariant disturbances. In addition, future work could incorporate prior knowledge for more directed and efficient exploration.


\clearpage
\acknowledgments{This work was partially supported by New Generation Artificial Intelligence-National Science and Technology Major Project (No. 2025ZD0122905)}


\bibliography{example}  
\clearpage





\appendix

\section{Implementation Details of HumanoidNDM}

\subsection{Deployment-viable Real-world Exploration with Frequency-Domain Perturbations}

\subsubsection{Weighted Policy Neighborhood Coverage Metric}
As introduced in the main text, we formulate the online adjustment of perturbation intensity $\sigma$ as a Multi-Armed Bandit (MAB) problem over $K$ discrete levels $\Sigma = \{\sigma_1, \dots, \sigma_K\}$. 

However, coverage across all timesteps should not be treated equally during an episode. For instance, aggressive exploration is unnecessary during a steady standing phase where tracking errors are minimal. To address this, we compute a weighted Policy Neighborhood Coverage metric $\bar{C}$, using the step-wise global positional tracking error $e_t$ as the weighting factor:
\begin{equation}
    \bar{C} = \frac{\sum_{t=1}^{T} e_t C_t}{\sum_{t=1}^{T} e_t}
\end{equation}
where 
\begin{equation}
    C_t = \frac{1}{N_r N_d} \sum_{j=1}^{N_r} \sum_{k=1}^{N_d} \mathbb{I}\left( \mathcal{B} \cap B_t^{(j,k)} \neq \emptyset \right)
\end{equation}

In our implementation, the number of radial bins $N_r$ and directional bins $N_d$ are maintained as constant values across all trajectories. Conversely, the neighborhood radius $R$ is treated as a motion-specific hyperparameter. The rationale behind this design lies in the varying dependence of policy fine-tuning on the dynamics model's gradients. For motions characterized by relatively simple dynamics or stable contact states (e.g., \textit{Swing}), the learned neural dynamics model typically exhibits robust interpolation and extrapolation capabilities. In such scenarios, the model can provide reliable, well-behaved gradients even if the collected data distribution is relatively sparse and widely dispersed, allowing us to adopt a larger radius $R$. On the other hand, for motions that necessitate fine-grained control and involve highly non-linear, sensitive dynamics (e.g., \textit{Balance}), the model's capacity for accurate interpolation and extrapolation naturally diminishes. Consequently, a smaller $R$ is required to enforce dense data coverage within a tightly localized neighborhood, ensuring the model accurately captures the precise dynamics necessary for effective policy fine-tuning.

For each candidate intensity $\sigma_i$, we maintain a localized recent buffer $\mathcal{B}_i$ to evaluate its local exploration quality. At each iteration, the empirical reward for choosing $\sigma_i$ is evaluated by this weighted Policy Neighborhood Coverage score: $S(\sigma_i) = \bar{C}$.

\subsubsection{Adaptive Intensity Adjustment via Multi-Armed Bandit}
To effectively balance exploration of unknown intensity effects and exploitation of known optimal levels, we employ the Upper Confidence Bound (UCB) algorithm. The priority score $U(\sigma_i)$ for each arm is updated as follows:
\begin{equation}
    U(\sigma_i) = S(\sigma_i) + \beta \sqrt{\frac{\ln N}{n_i}}
\end{equation}
where $n_i$ is the cumulative selection count for level $\sigma_i$, $N$ is the total number of training iterations across all levels, and $\beta$ is a hyperparameter controlling the exploration scale.

To achieve progressive exploration, the bandit is restricted to transition only between the current intensity level and its immediately adjacent levels. The optimal intensity for the subsequent iteration is then selected via:
\begin{equation}
    \sigma^* = \arg\max_{\sigma_i \in \text{adjacent}} U(\sigma_i)
\end{equation}
By iteratively tracking this argmax within adjacent bounds, FENCE dynamically adjusts the exploration intensity to maintain optimal data coverage without triggering mechanical instability.

\subsection{MoCap-Free Neural Dynamics Model Training with Dual-Space Objective}

\subsubsection{Kinematically Constrained Configuration and Gravity Prediction}
In our implementation, the world model $w$ does not predict the next joint configuration or the gravity vector as fully independent, unconstrained outputs. To enforce physical and kinematic consistency during autoregressive rollouts, specific state components undergo tailored integration and coordinate transformation steps.

\paragraph{Joint Configuration Update}
Although the network outputs a residual vector with the same dimensionality as the processed state, the residual component corresponding to the joint position $q_t$ is not directly applied. Instead, the model predicts the next velocity-like term via residual addition:
\begin{equation*}
    \hat{\dot{q}}_{t+1} = \dot{q}_t + \Delta \dot{q}_t
\end{equation*}
where $\Delta \dot{q}_t$ is the predicted joint velocity residual. The next joint position is subsequently obtained through first-order integration:
\begin{equation*}
    \hat{q}_{t+1} = q_t + \hat{\dot{q}}_{t+1}\Delta t
\end{equation*}
This design enforces a strict kinematic relationship between joint positions and velocities, preventing the model from producing arbitrary, discontinuous joint predictions across long rollout horizons.

\paragraph{Gravity Vector Roll-Pitch Transformation}
The onboard proprioceptive gravity vector $\mathbf{g}_t \in \mathbb{R}^3$ possesses geometric redundancy as it resides on a unit sphere, capturing only two underlying degrees of freedom. To eliminate this redundancy and facilitate smoother optimization, we map $\mathbf{g}_t$ into a 2D roll-pitch representation $\mathbf{r}_t = [\phi_t, \theta_t]^\top \in \mathbb{R}^2$ prior to feeding the history sequence into the history encoder:
\begin{equation*}
    \mathbf{r}_t = \mathcal{T}_{\text{3D}\to\text{2D}}(\mathbf{g}_t)
    \label{eq:appendix_g_to_rp}
\end{equation*}
where $\mathcal{T}_{\text{3D}\to\text{2D}}$ denotes the transformation from the 3D gravity vector to Euler angles (roll and pitch). The neural dynamics model then operates on this low-dimensional space and outputs a 2D orientation residual $\Delta \mathbf{r}_t$. The predicted roll-pitch for the next time step is updated via residual addition:
\begin{equation*}
    \hat{\mathbf{r}}_{t+1} = \mathbf{r}_t + \Delta \mathbf{r}_t
\end{equation*}
Finally, the effective 3D gravity vector prediction $\hat{\mathbf{g}}_{t+1}$ is reconstructed by mapping $\hat{\mathbf{r}}_{t+1}$ back into the 3D Cartesian space:
\begin{equation*}
    \hat{\mathbf{g}}_{t+1} = \mathcal{T}_{\text{2D}\to\text{3D}}(\hat{\mathbf{r}}_{t+1})
\end{equation*}
Other unconstrained state components, such as the gyroscope signals $\boldsymbol{\omega}_t$ and the estimated root height $h_t$, are updated directly through standard residual addition.

\subsubsection{Shared-Backbone Multi-Head Ensemble}
To enhance predictive robustness while maintaining computational efficiency, the ensemble framework utilized in our method is implemented as a shared-backbone multi-head ensemble, rather than a collection of fully independent models.

All $M$ ensemble members share the same history encoder and decoder backbone. The historical state-action sequence is encoded by a shared Transformer trunk. Ensemble diversity is injected solely at the final prediction heads, where each head maps the shared feature to a separate next-state prediction.
During training, bootstrap sampling masks are applied independently at the head level, meaning different prediction heads are optimized on different sampled subsets of the same batch. Therefore, the ensemble should be interpreted as a bootstrapped multi-head dynamics model that captures prediction diversity through its output heads, significantly reducing the overhead compared to training $M$ separate encoders and decoders.

\subsubsection{Training Objectives of Neural Dynamics Model}

The neural dynamics model $w$ is trained with the following losses:
\begin{itemize}
    \item $\mathcal{L}^{w}_{\omega}$: the root angular-velocity prediction loss,
    \begin{equation*}
    \mathcal{L}^{w}_{\omega} = \left\| \hat{\boldsymbol{\omega}}_{t+k} - \boldsymbol{\omega}_{t+k} \right\|_2 .
    \end{equation*}
    Here, $\hat{\boldsymbol{\omega}}_{t+k}$ and $\boldsymbol{\omega}_{t+k}$ denote the predicted and target root angular velocities, respectively.

    \item $\mathcal{L}^{w}_{\mathrm{rp}}$: the root roll--pitch prediction loss,
    \begin{equation*}
    \mathcal{L}^{w}_{\mathrm{rp}} = \left\| \hat{\boldsymbol{\rho}}_{t+k} - \boldsymbol{\rho}_{t+k} \right\|_2 .
    \end{equation*}
    Here, $\hat{\boldsymbol{\rho}}_{t+k}$ and $\boldsymbol{\rho}_{t+k}$ denote the predicted and target root roll--pitch angles, respectively.

    \item $\mathcal{L}^{w}_{q}$: the joint-position prediction loss,
    \begin{equation*}
    \mathcal{L}^{w}_{q} = \left\| \hat{q}_{t+k} - q_{t+k} \right\|_2 .
    \end{equation*}
    Here, $\hat{q}_{t+k}$ and $q_{t+k}$ denote the predicted and target joint positions, respectively.

    \item $\mathcal{L}^{w}_{h}$: the root-height prediction loss,
    \begin{equation*}
    \mathcal{L}^{w}_{h} = \left| \hat{h}_{t+k} - h_{t+k} \right| .
    \end{equation*}
    Here, $\hat{h}_{t+k}$ and $h_{t+k}$ denote the predicted and target root heights, respectively.

    \item $\mathcal{L}^{w}_{\mathrm{link\_pos}}$: the link-position loss in a local gravity-aligned frame,
    \begin{equation*}
    \mathcal{L}^{w}_{\mathrm{link\_pos}} = \left\| \hat{\mathbf{P}}_{t+k} - \mathbf{P}_{t+k} \right\|_F .
    \end{equation*}
    Here, $\hat{\mathbf{P}}_{t+k}, \mathbf{P}_{t+k} \in \mathbb{R}^{|\mathcal{B}| \times 3}$ denote the predicted and target local gravity-aligned frame positions of all robot links, respectively.

    \item $\mathcal{L}^{w}_{\mathrm{link\_rot}}$: the link-rotation loss in a local gravity-aligned frame,
    \begin{equation*}
    \mathcal{L}^{w}_{\mathrm{link\_rot}} = \frac{1}{|\mathcal{B}|} \sum_{\ell \in \mathcal{B}} d_{\mathrm{geo}}^{2} \left( \hat{\mathbf{R}}_{\ell,t+k}, \mathbf{R}_{\ell,t+k} \right) ,
    \end{equation*}
    where $\mathcal{B}$ denotes the set of robot links, and $\hat{\mathbf{R}}_{\ell,t+k}, \mathbf{R}_{\ell,t+k} \in SO(3)$ are the predicted and target local gravity-aligned frame rotations of link $\ell$, respectively. The geodesic rotation distance is defined as
    \begin{equation*}
    d_{\mathrm{geo}}(\mathbf{R}_1,\mathbf{R}_2) = \arccos \left( \frac{ \mathrm{tr}(\mathbf{R}_1\mathbf{R}_2^\top)-1 }{2} \right).
    \end{equation*}
\end{itemize}

\subsection{Efficient and Stable Policy Fine-Tuning via Supervised Gradients}

\subsubsection{Initialization of Autoregressive Rollout}

During policy fine-tuning, autoregressive rollouts are not initialized from the beginning of the reference trajectory or from a nominal default pose. Instead, each training rollout is initialized using a short history segment sampled directly from the buffer. 

Concretely, the exploration buffer stores tuples comprising collected real-world states, executed actions, corresponding reference trajectory indices, and previous actions. For each policy update iteration, we sample a sequence consisting of a history prefix of length $H$ and a rollout horizon of length $L_\pi$. The first $H$ collected states $\mathbf{s}_{t-H+1:t}$ and $H-1$ collected actions $\mathbf{a}_{t-H+1:t-1}$ are extracted to initialize the history context for the dynamics model.

This buffer-initialization strategy ensures that policy fine-tuning optimizes the policy starting from states that were actually visited during real-world data collection. Consequently, it anchors the supervised-gradient updates locally to the target-domain state distribution covered by the exploration buffer. Furthermore, it supplies the history-conditioned world model with a realistic and dynamically consistent recent state-action context, which is crucial for accurate forward prediction.

\subsubsection{Training Objectives of Policy}

The policy $\pi$ is fine-tuned with the following losses:
\begin{itemize}
    \item $\mathcal{L}^{\pi}_{\mathrm{rp}}$: the root roll--pitch tracking loss,
    \begin{equation*}
    \mathcal{L}^{\pi}_{\mathrm{rp}} = \left\| \hat{\boldsymbol{\rho}}^{\pi}_{t+k} - \boldsymbol{\rho}_{t+k} \right\|_2^2 .
    \end{equation*}
    Here, $\hat{\boldsymbol{\rho}}^{\pi}_{t+k}$ and $\boldsymbol{\rho}_{t+k}$ denote the rollout and target root roll--pitch angles, respectively.

    \item $\mathcal{L}^{\pi}_{q}$: the joint-position tracking loss,
    \begin{equation*}
    \mathcal{L}^{\pi}_{q} = \left\| \hat{q}^{\pi}_{t+k} - q_{t+k} \right\|_2^2 .
    \end{equation*}
    Here, $\hat{q}^{\pi}_{t+k}$ and $q_{t+k}$ denote the rollout and target joint positions, respectively.

    \item $\mathcal{L}^{\pi}_{\mathrm{sm}}$: the joint-space smoothness regularization loss,
    \begin{equation*}
    \mathcal{L}^{\pi}_{\mathrm{sm}} = \left\| \hat{q}^{\pi}_{t+k} - 2\hat{q}^{\pi}_{t+k-1} + \hat{q}^{\pi}_{t+k-2} \right\|_2^2 .
    \end{equation*}
    Here, $\hat{q}^{\pi}_{t+k}$, $\hat{q}^{\pi}_{t+k-1}$, and $\hat{q}^{\pi}_{t+k-2}$ denote the joint positions at three consecutive rollout steps.

    \item $\mathcal{L}^{\pi}_{\mathrm{link\_pos}}$: the link-position loss in a local gravity-aligned frame,
    \begin{equation*}
    \mathcal{L}^{\pi}_{\mathrm{link\_pos}} = \left\| \hat{\mathbf{P}}^{\pi}_{t+k} - \mathbf{P}_{t+k} \right\|_F .
    \end{equation*}
    Here, $\hat{\mathbf{P}}^{\pi}_{t+k}, \mathbf{P}_{t+k} \in \mathbb{R}^{|\mathcal{B}| \times 3}$ denote the rollout and target link positions of all robot links, respectively.

    \item $\mathcal{L}^{\pi}_{\mathrm{link\_rot}}$: the link-rotation loss in a local gravity-aligned frame,
    \begin{equation*}
    \mathcal{L}^{\pi}_{\mathrm{link\_rot}} = \frac{1}{|\mathcal{B}|} \sum_{\ell \in \mathcal{B}} d_{\mathrm{geo}}^{2} \left( \hat{\mathbf{R}}^{\pi}_{\ell,t+k}, \mathbf{R}_{\ell,t+k} \right) ,
    \end{equation*}
    where $\mathcal{B}$ denotes the set of robot links, and $\hat{\mathbf{R}}^{\pi}_{\ell,t+k}, \mathbf{R}_{\ell,t+k} \in SO(3)$ are the rollout and target rotations of link $\ell$, respectively. 
\end{itemize}

\subsection{Hyperparameters of HumanoidNDM}
Important hyperparameters of HumanoidNDM are listed in Table~\ref{tab:hyperparameters}.

\begin{table}[t]
  \centering
  \small
  \caption{Hyperparameters of HumanoidNDM}
  \label{tab:hyperparameters}
  \resizebox{0.85\textwidth}{!}{
  \begin{tabular}{p{0.68\linewidth}p{0.20\linewidth}}
  \toprule
  Variable & Value \\
  \midrule\midrule
  
  \multicolumn{2}{c}{\textbf{Data Collection}} \\
  \midrule\midrule
  Max perturbation frequency ($k_c$) & 1.0 Hz \\
  Roll-pitch termination threshold & 1.0 \\
  Radial bin number & 2\\
  Directional bin number & 16\\
  MAB exploration scale ($\beta$) & 0.05\\
  \midrule\midrule
  
  \multicolumn{2}{c}{\textbf{Model Training}} \\
  \midrule\midrule
  History length ($H$) & 10 \\
  Autoregressive rollout length ($L_w$) & 8 \\
  Ensemble size ($M$)& 5 \\
  Mini-batch size & 512 \\
  Mini-batches per epoch & 32 \\
  Model learning rate ($\eta_{w}$) & $2\times 10^{-3}$ \\
  Bootstrap probability & 0.8 \\
  Replay-buffer horizon in episodes & 100 \\
  Weight of $\mathcal{L}^{w}_{\omega}$ & 1.0 \\
  Weight of $\mathcal{L}^{w}_{\mathrm{rp}}$ & 1.0 \\
  Weight of $\mathcal{L}^{w}_{q}$ & 1.0 \\
  Weight of $\mathcal{L}^{w}_{h}$ & 1.0 \\
  Weight of $\mathcal{L}^{w}_{\mathrm{link\_pos}}$ & 1.0 \\
  Weight of $\mathcal{L}^{w}_{\mathrm{link\_rot}}$ & 5.0 \\
  \midrule\midrule
  
  \multicolumn{2}{c}{\textbf{Policy Training}} \\
  \midrule\midrule
  Policy learning rate ($\eta_{\pi}$) & $10^{-4}$ \\
  Learning-rate decay factor & 1.0 \\
  Final learning rate ($\eta_{\pi}^{\mathrm{final}}$) & $10^{-7}$ \\
  Autoregressive rollout length ($L_\pi$) & 128 \\
  Mini-batch size ($B_{\pi}$) & 512 \\
  Mini-batches per epoch ($N_{\mathrm{mb}}^{\pi}$) & 2 \\
  Weight of $\mathcal{L}^{\pi}_{q}$ & 10.0 \\
  Weight of $\mathcal{L}^{\pi}_{\mathrm{rp}}$ & 10.0 \\
  Weight of $\mathcal{L}^{\pi}_{\mathrm{sm}}$ & 10.0 \\
  Weight of $\mathcal{L}^{\pi}_{\mathrm{link\_pos}}$ & 1.0 \\
  Weight of $\mathcal{L}^{\pi}_{\mathrm{link\_rot}}$ & 1.0 \\
  \bottomrule
  \end{tabular}}
\end{table}

\subsection{Architecture of the Neural Dynamics Model}
The neural dynamics model employs a history-based Transformer ensemble architecture to predict future states. Let $s_t = [\boldsymbol{\omega}_t, \boldsymbol{\rho}_t, q_t, \dot{q}_t, h_t] \in \mathbb{R}^{64}$ denote the system state and $a_t \in \mathbb{R}^{23}$ denote the action. We utilize a history length of $H=10$ and an ensemble size of $M=5$. The architecture consists of a history encoder $\mathcal{E}_{\mathrm{hist}}$, a shared decoder $\mathcal{D}$, and independent prediction heads.

\paragraph{History Encoding}
To capture temporal dependencies, we process the past $H-1$ state-action pairs. We first obtain the dimension-wise scaled state $\tilde{s}_t$ and construct input tokens $x_i$ via a linear projection $\mathcal{P}: \mathbb{R}^{87}\rightarrow\mathbb{R}^{64}$:
\begin{equation*}
x_i = \mathcal{P}\big([\tilde{s}_{t-H+i}, a_{t-H+i}]\big) + p_i, \qquad i=1,\dots,H-1,
\end{equation*}
where $p_i \in \mathbb{R}^{64}$ is a learnable positional embedding. These tokens are then processed by a Transformer encoder $\mathcal{E}_{\mathrm{hist}}$, which consists of $N_{\mathrm{enc}}=2$ layers, $n_{\mathrm{head}}=4$ attention heads, and a feedforward network width of $d_{\mathrm{ff}}=16$. We extract the history feature $z_t \in \mathbb{R}^{64}$ from the final token:
\begin{equation*}
z_t = \mathcal{E}_{\mathrm{hist}}(x_{1:H-1})_{H-1}.
\end{equation*}

\paragraph{Decoder and Ensemble Prediction}
The history feature $z_t$ is concatenated with the current scaled state $\tilde{s}_t$ and action $a_t$, forming a 151-dimensional input vector. This vector is passed through a shared multilayer perceptron (MLP) decoder $\mathcal{D}$:
\begin{equation*}
u_t = \mathcal{D}([z_t,\tilde{s}_t,a_t]), \qquad \mathcal{D}: 151 \rightarrow 64 \rightarrow 512 \rightarrow 512,
\end{equation*}
where the Exponential Linear Unit (ELU) activation function is applied between the hidden layers.

To model dynamics uncertainty, each member of the ensemble utilizes an independent linear prediction head $g_m:\mathbb{R}^{512}\rightarrow\mathbb{R}^{64}$ to predict the residual of the next state:
\begin{equation*}
\Delta \tilde{s}_{t+1}^{(m)} = g_m(u_t), \qquad m=1,\dots,M.
\end{equation*}

\paragraph{State Integration}
The next state for each ensemble member is derived through residual prediction for most state components, while the joint positions are explicitly integrated from the predicted joint velocities using a time step of $\Delta t = 0.02$\,s:
\begin{equation*}
\begin{aligned}
\hat{\boldsymbol{\omega}}_{t+1}^{(m)} &= \boldsymbol{\omega}_t + \Delta \boldsymbol{\omega}_{t+1}^{(m)}, \\
\hat{\boldsymbol{\rho}}_{t+1}^{(m)} &= \boldsymbol{\rho}_t + \Delta \boldsymbol{\rho}_{t+1}^{(m)}, \\
\hat{\dot{q}}_{t+1}^{(m)} &= \dot{q}_t + \Delta \dot{q}_{t+1}^{(m)}, \\
\hat{h}_{t+1}^{(m)} &= h_t + \Delta h_{t+1}^{(m)}, \\
\hat{q}_{t+1}^{(m)} &= q_t + \hat{\dot{q}}_{t+1}^{(m)}\Delta t.
\end{aligned}
\end{equation*}

\section{Implementation of Baselines}

\paragraph{ASAP}
We ported the official implementation of ASAP to the MuJoCo-Jax simulator. During this migration, we strictly preserved all original Proximal PPO training hyperparameters and reward formulations. Empirically, we observed that the fine-tuning performance is sensitive to the penalty coefficients for the action rate and action norm. Consequently, we conducted a systematic grid search over these two reward parameters for each experimental setting. All reported results for ASAP are averaged over 5 independent trials, utilizing the optimally identified hyperparameter configuration.

\paragraph{RWM}
Similarly, we integrated the official implementation of RWM into the MuJoCo-Jax simulator. While we retained the default PPO training configurations from the original codebase, the loss functions were aligned to be strictly consistent with those employed in HumanoidNDM. Our preliminary evaluations indicated that the rollout length, policy learning rate, and the initial standard deviation of the action distribution critically influence the downstream fine-tuning efficacy. Therefore, a grid search was performed across these three hyperparameters under every setting. All reported results for RWM are averaged over 5 independent trials, utilizing the optimally identified hyperparameter configuration.

\section{Simulation Experiment Settings}
We designed 5 motions paired with 3 kinds of disturbances in simulation experiments.

\noindent
\begin{minipage}[t]{0.75\linewidth}
    \vspace{0pt}
    \paragraph{Walk}
    Starting from the default pose, the robot takes four steps directly forward, stops, and returns to the initial default pose. Natural arm swings are maintained throughout motion.
    \begin{itemize}
        \item \textit{Extra Weight:} The mass of the \texttt{torso} is multiplied by 1.2, and its local center of mass is offset by [-0.2, 0, 0].
        \item \textit{Scaled KP:} The KP of the \texttt{waist\_yaw} is decreased from 300 to 0.
        \item \textit{Small Friction:} The coefficient of friction between the robot's feet and the ground is reduced from 1.0 to 0.3.
    \end{itemize}
\end{minipage}\hfill
\begin{minipage}[t]{0.22\linewidth}
    \vspace{0pt}
    \includegraphics[width=\linewidth]{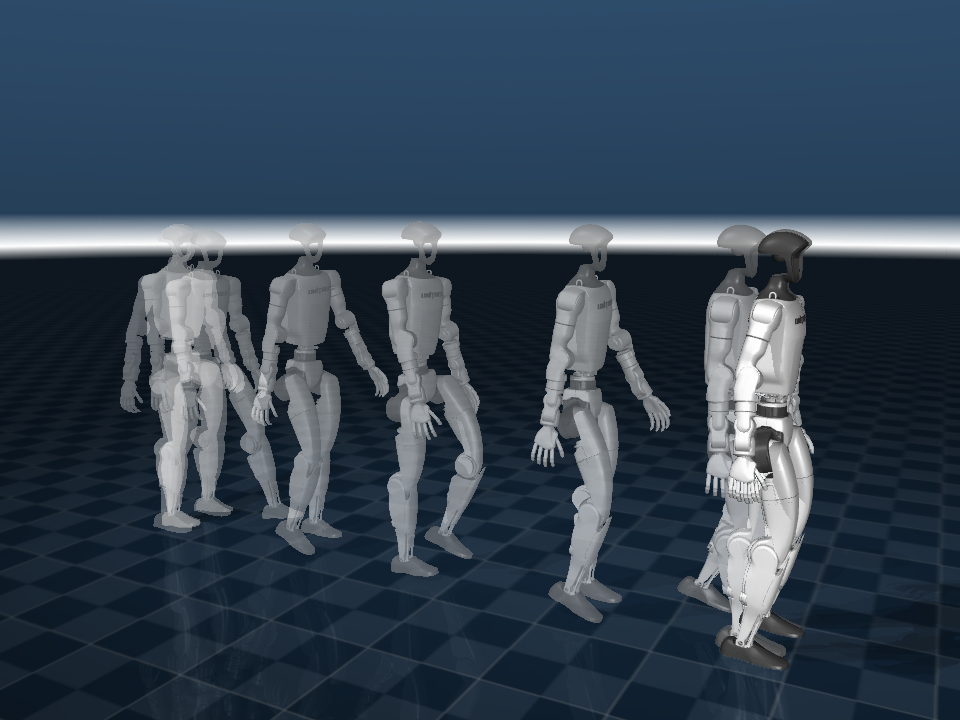}
\end{minipage}
\par\vspace{1em}

\noindent
\begin{minipage}[t]{0.75\linewidth}
    \vspace{0pt}
    \paragraph{Balance}
    Starting from the default pose, the robot lifts its left leg while leaning its torso forward to maintain equilibrium. This posture is held for approximately 2 seconds before the left leg is slowly lowered, returning the robot to the default pose.
    \begin{itemize}
        \item \textit{Extra Weight:} The mass of the \texttt{right\_knee} is multiplied by 1.5, and its local center of mass is offset by [0, -0.2, 0].
        \item \textit{Scaled KP:} The KP of the \texttt{left\_hip\_pitch} is increased from 100 to 300.
        \item \textit{Small Friction:} The coefficient of friction between the robot's feet and the ground is reduced from 1.0 to 0.3.
    \end{itemize}
\end{minipage}\hfill
\begin{minipage}[t]{0.22\linewidth}
    \vspace{0pt}
    \includegraphics[width=\linewidth]{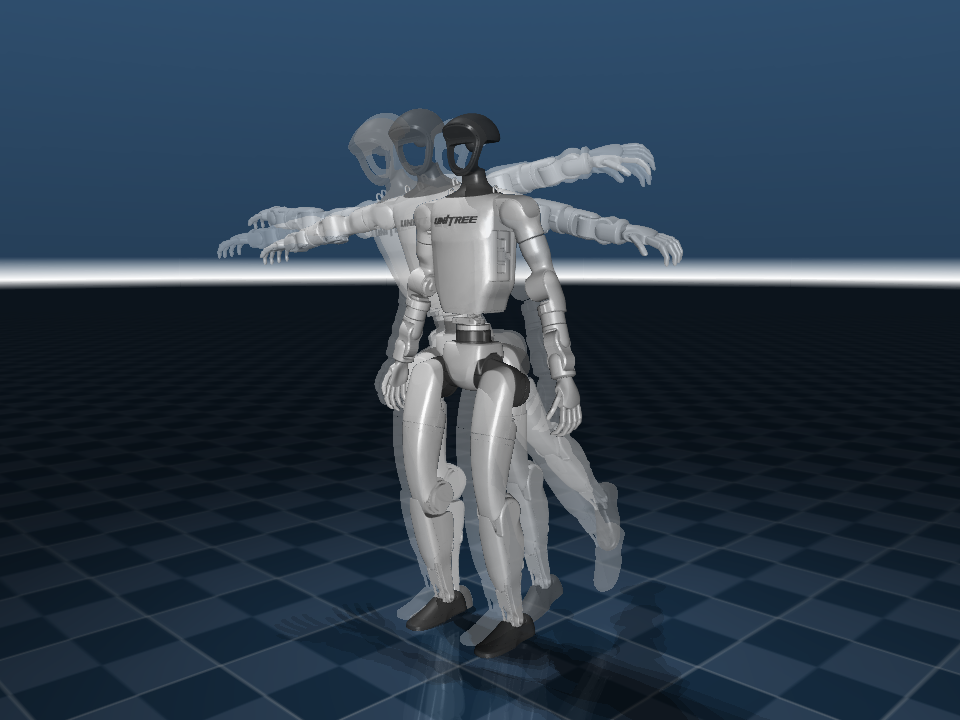}
\end{minipage}
\par\vspace{1em}

\noindent
\begin{minipage}[t]{0.75\linewidth}
    \vspace{0pt}
    \paragraph{Squat}
    Starting from the default pose, the robot raises both arms forward and simultaneously executes a squatting motion. After approximately 2 seconds, the robot stands upright, lowers its arms, and returns to the default pose.
    \begin{itemize}
        \item \textit{Extra Weight:} The masses of the \texttt{left\_shoulder\_roll}, \texttt{left\_shoulder\_yaw}, \texttt{left\_elbow}, \texttt{left\_wrist\_roll} are multiplied by 8, and their respective local centers of mass are offset by [0, -0.3, 0].
        \item \textit{Scaled KP:} The KP of the \texttt{right\_hip\_pitch} is decreased from 100 to 30.
        \item \textit{Small Friction:} The coefficient of friction between the robot's feet and the ground is reduced from 1.0 to 0.3.
    \end{itemize}
\end{minipage}\hfill
\begin{minipage}[t]{0.22\linewidth}
    \vspace{0pt}
    \includegraphics[width=\linewidth]{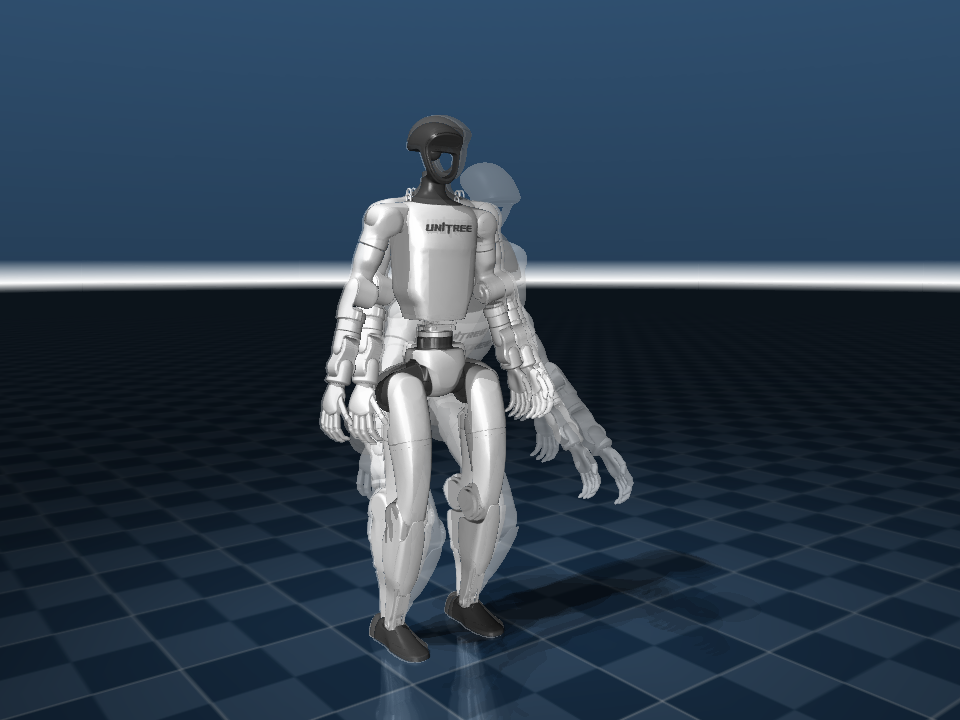}
\end{minipage}
\par\vspace{1em}

\noindent
\begin{minipage}[t]{0.75\linewidth}
    \vspace{0pt}
    \paragraph{Swing}
    Starting from the default pose, the robot raises its right arm vertically while swinging its left arm slightly backward. It then performs two rhythmic, small-amplitude extensions in this configuration, mimicking a dynamic chest expansion exercise. Following this, the robot executes the mirrored motion by swapping the positions of its arms, before finally returning to the default pose.
    \begin{itemize}
        \item \textit{Extra Weight:} The masses of the \texttt{left\_shoulder\_roll}, \texttt{left\_shoulder\_yaw}, \texttt{left\_elbow}, \texttt{left\_wrist\_roll} are multiplied by 4, and their respective local centers of mass are offset by [0, -0.2, 0].
        \item \textit{Scaled KP:} The KP of the \texttt{left\_shoulder\_pitch} is decreased from 90 to 10.
        \item \textit{Small Friction:} The coefficient of friction between the robot's feet and the ground is reduced from 1.0 to 0.3.
    \end{itemize}
\end{minipage}\hfill
\begin{minipage}[t]{0.22\linewidth}
    \vspace{0pt}
    \includegraphics[width=\linewidth]{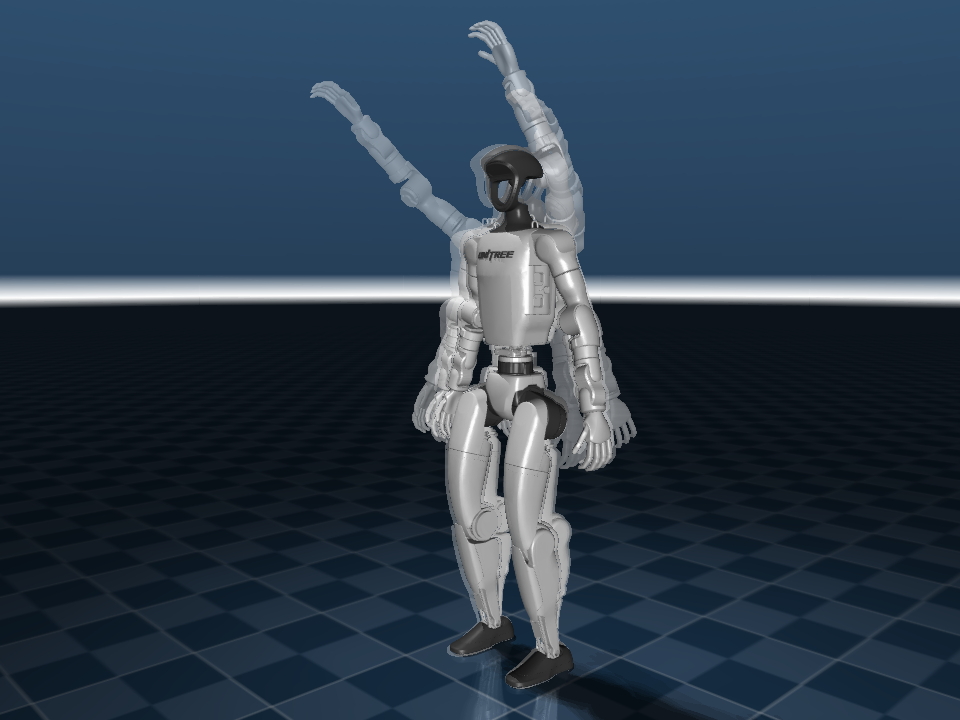}
\end{minipage}
\par\vspace{1em}

\noindent
\begin{minipage}[t]{0.75\linewidth}
    \vspace{0pt}
    \paragraph{Stride}
    Starting from the default pose, the robot brings both arms close to its chest and steps its right foot diagonally into a forward-right lunge. It then retracts the right leg and returns to the default pose.
    \begin{itemize}
        \item \textit{Extra Weight:} The mass of the \texttt{torso} is multiplied by 1.2, and its local center of mass is offset by [-0.2, 0, 0].
        \item \textit{Scaled KP:} The KP of the \texttt{left\_hip\_roll} is decreased from 100 to 30.
        \item \textit{Small Friction:} The coefficient of friction between the robot's feet and the ground is reduced from 1.0 to 0.3.
    \end{itemize}
\end{minipage}\hfill
\begin{minipage}[t]{0.22\linewidth}
    \vspace{0pt}
    \includegraphics[width=\linewidth]{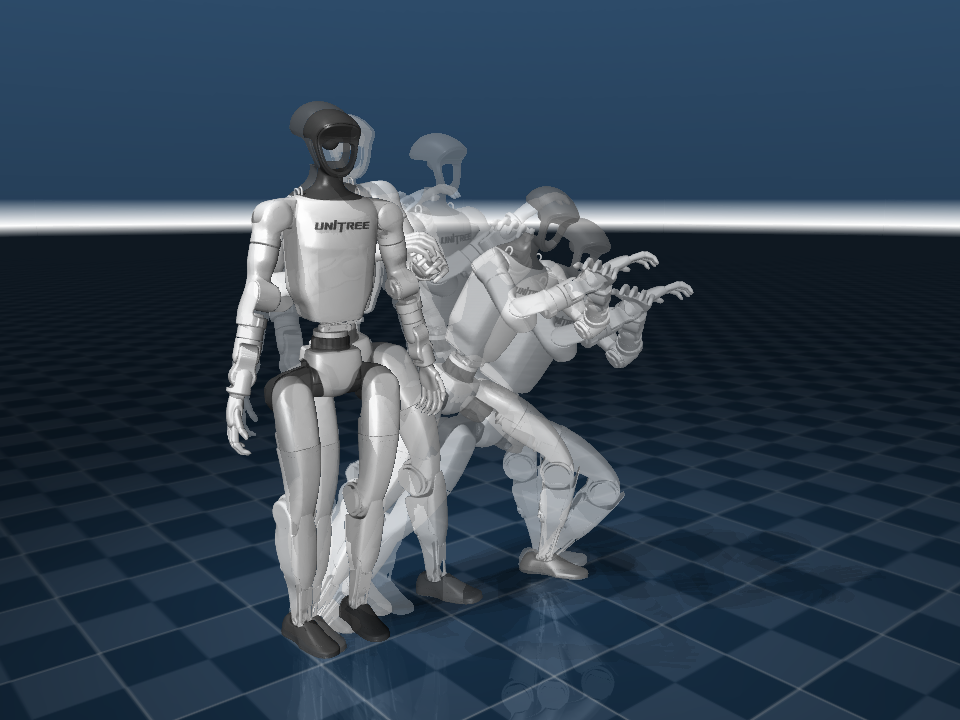}
\end{minipage}

\section{Real-World Experiment Settings}

To evaluate HumanoidNDM in the real world, we designed six real-world tasks. We divided them into two groups based on the type of disturbance: \textbf{Added Weight Disturbances} (which are easier to simulate) and \textbf{Hard-to-Model Disturbances} (which involve complex dynamics like elasticity, fluid sloshing, and damaged motors). The detailed settings are described below.

NDM and policy training are performed on a single RTX 4090 GPU. Each adaptation iteration takes approximately 3 s for NDM and policy optimization and 10 s for data collection. The latter includes returning the robot to its initial position. Thus, 200 iterations can be completed within one hour. All reported real-world results are averaged over 3 independent trials, with no failed runs. Occasional interruptions caused by environmental collisions or motor overheating were handled by restoring the latest checkpoint and resuming training.

\subsection{Added Weight Disturbances}
In this category, we attached weights to specific links of the robot.

\paragraph{Swallow Balance}
\begin{itemize}
    \item \textit{Action Description:} The robot holds its arms out and lifts its left leg back.
    \item \textit{Disturbance:} A 1.5 kg iron weight is strapped to each calf.
    \item \textit{Collected Real-World Data:} 200 trajectories.
    \item \textit{Task-specific Metric (Air Time $\uparrow$):} Measured in seconds (s). This shows how long the left foot stays in the air. A longer time means better balance.
\end{itemize}

\paragraph{Arm Swing}
\begin{itemize}
    \item \textit{Action Description:} The robot swings its right arm up, then its left arm, like marching.
    \item \textit{Disturbance:} A 1.5 kg iron weight is attached to the left hand.
    \item \textit{Collected Real-World Data:} 200 trajectories.
    \item \textit{Task-specific Metric (Arm Height $\uparrow$):} Measured in meters (m). This is the highest point the loaded hand reaches. It shows if the loaded arms can lift the extra weight.
\end{itemize}

\paragraph{Pen Capping}
\begin{itemize}
    \item \textit{Action Description:} The robot holds a pen cap in its left hand and a pen in its right. It brings them together to cap the pen, then moves its arms back down.
    \item \textit{Disturbance:} A 0.75 kg weight is placed only on the left hand.
    \item \textit{Collected Real-World Data:} 50 trajectories.
    \item \textit{Task-specific Metric (EE Distance $\downarrow$):} Measured in centimeters (cm). This is the shortest distance between the pen and cap. A smaller distance means better coordination.
\end{itemize}

\subsection{Hard-to-Model Disturbances}
In this category, we apply disturbances that are hard to model in simulation, including external elastic forces, unpredictable weight shifts from sloshing fluids, and hardware defects such as damaged motors. These experiments further prove the necessity of real-world fine-tuning.

\paragraph{Lateral Raise}
\begin{itemize}
    \item \textit{Action Description:} The robot raises both arms to the side to head height, then lowers them. It does this twice.
    \item \textit{Disturbance:} Two elastic bands connect the right hand to the left and right wrists. This adds elastic forces.
    \item \textit{Collected Real-World Data:} 200 trajectories.
    \item \textit{Task-specific Metric (Arm Height $\uparrow$):} Measured in meters (m). This is the highest point the right arm reaches. It tests if the robot can resist the pulling forces.
\end{itemize}

\paragraph{Walking}
\begin{itemize}
    \item \textit{Action Description:} The robot walks forward four steps and swings its arms naturally.
    \item \textit{Disturbance:} The robot holds a bucket with 2 kg of water in its right hand. The sloshing water shifts the weight unpredictably.
    \item \textit{Collected Real-World Data:} 200 trajectories.
    \item \textit{Task-specific Metric (Deviation Angle $\downarrow$):} Measured in degrees ($^\circ$). This shows how much the robot turns away from a straight path. A smaller angle means it handles the sloshing water better.
\end{itemize}

\paragraph{Bowing}
\begin{itemize}
    \item \textit{Action Description:} The robot steps back with its right foot, draws two circles with its right hand, brings the hand to the front, and bows.
    \item \textit{Disturbance:} We use a robot with damaged motors, which causes obvious stuttering.
    \item \textit{Collected Real-World Data:} 200 trajectories.
    \item \textit{Task-specific Metric (Noise Level $\downarrow$):} Measured in decibels (dB). A lower noise level means the policy generates smoother actions.
\end{itemize}

\section{Simulation Training Curves}
\begin{figure}[H]
    \centering
    \includegraphics[width=0.7\linewidth]{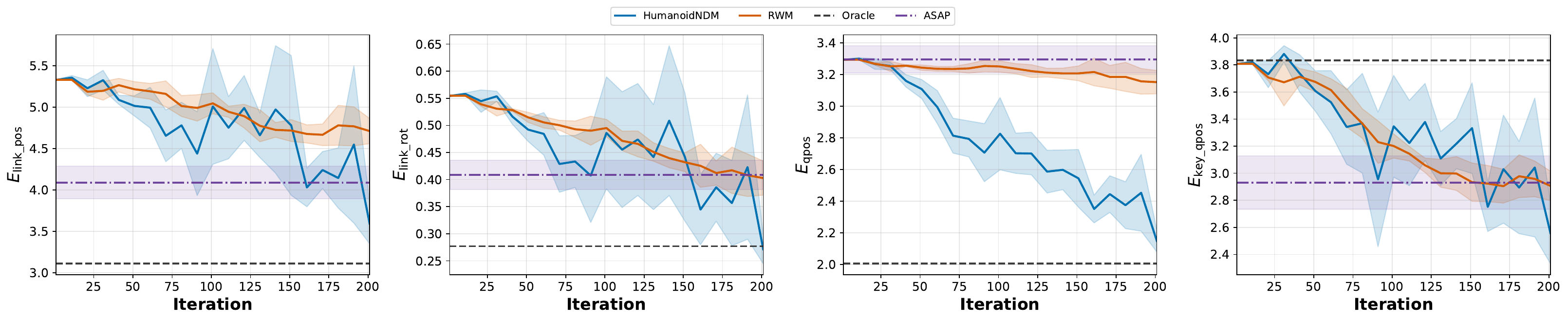} \\
    \includegraphics[width=0.7\linewidth]{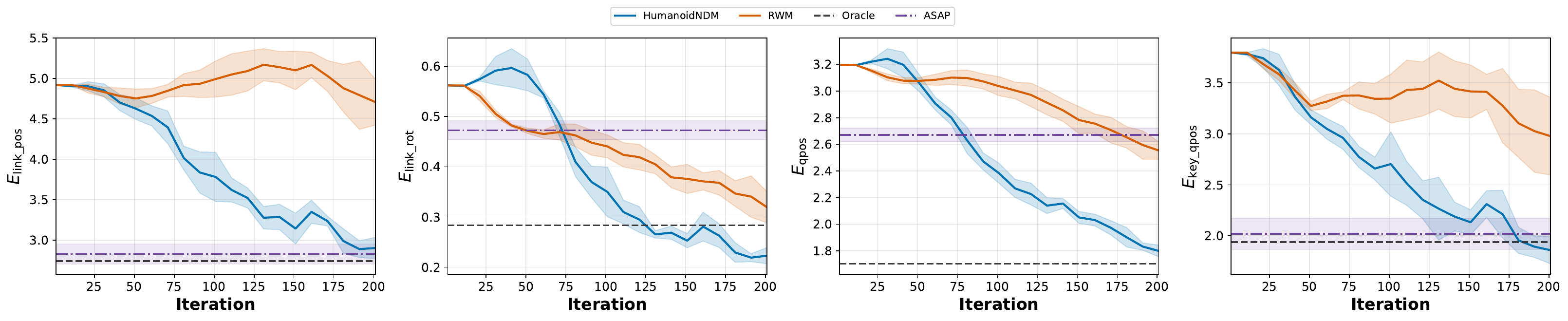} \\
    \includegraphics[width=0.7\linewidth]{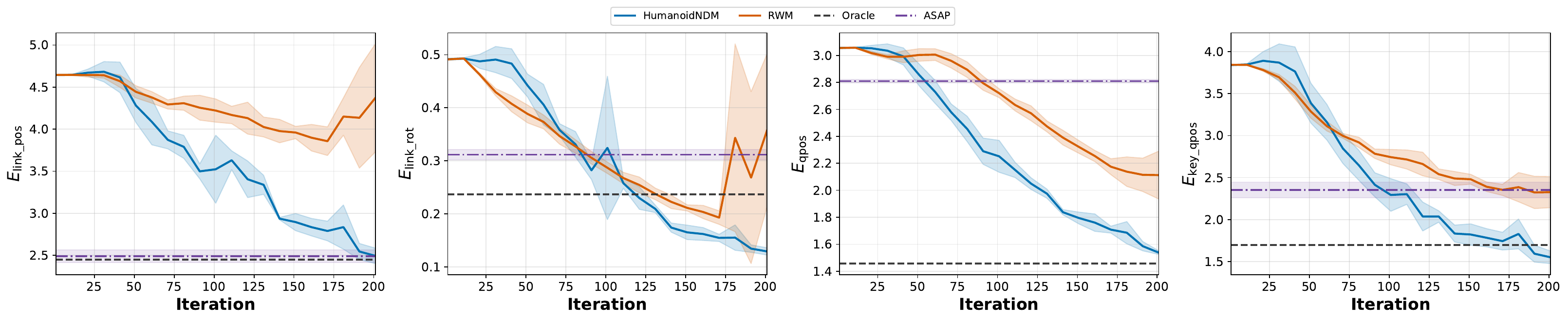}
    \caption{Walk: Extra Weight (top), Scaled KP (middle), Small Friction (bottom).}
    \label{fig:baseline_walk}
\end{figure}

\begin{figure}[H]
    \centering
    \includegraphics[width=0.7\linewidth]{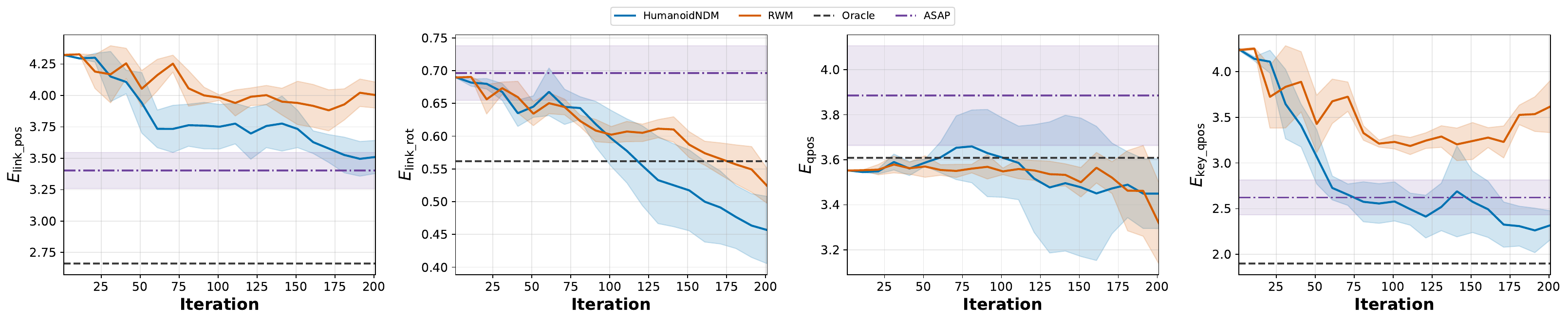} \\
    \includegraphics[width=0.7\linewidth]{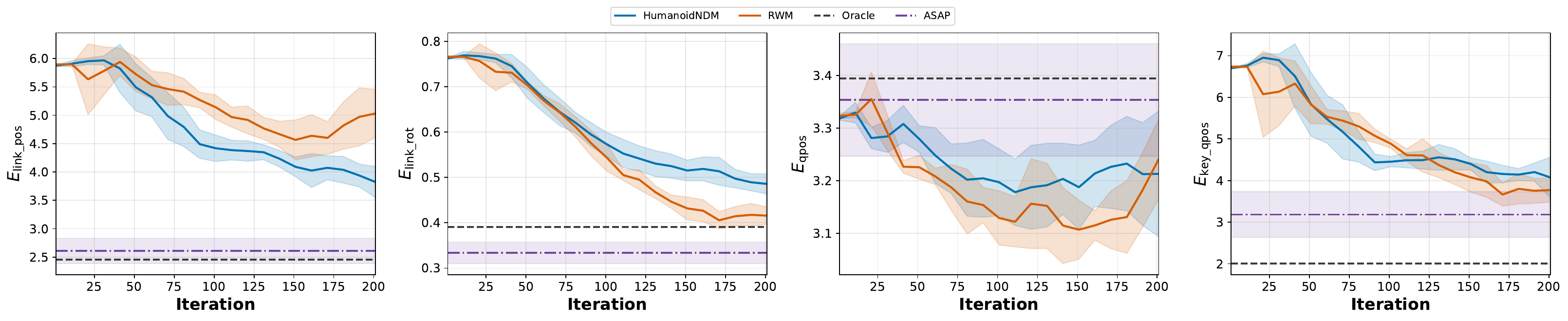} \\
    \includegraphics[width=0.7\linewidth]{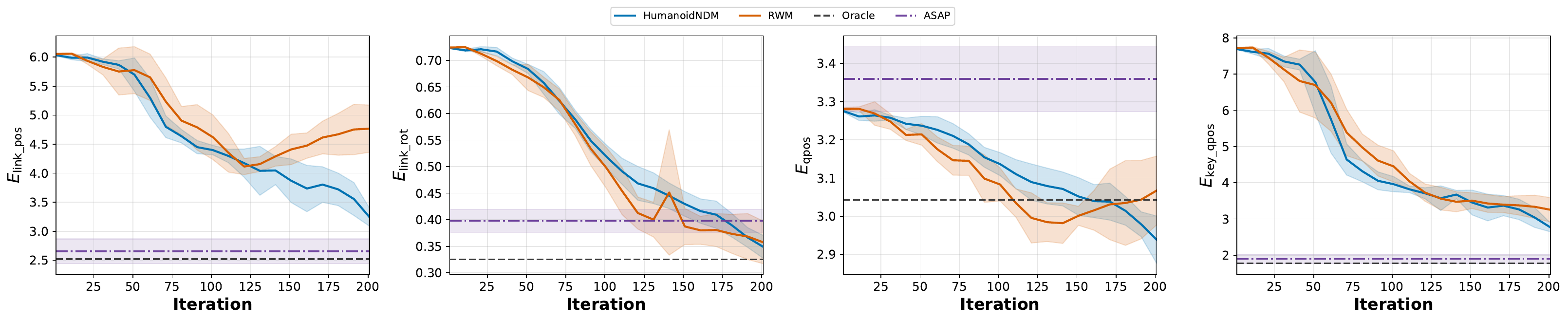}
    \caption{Balance: Extra Weight (top), Scaled KP (middle), Small Friction (bottom).}
    \label{fig:baseline_balance}
\end{figure}

\begin{figure}[H]
    \centering
    \includegraphics[width=0.7\linewidth]{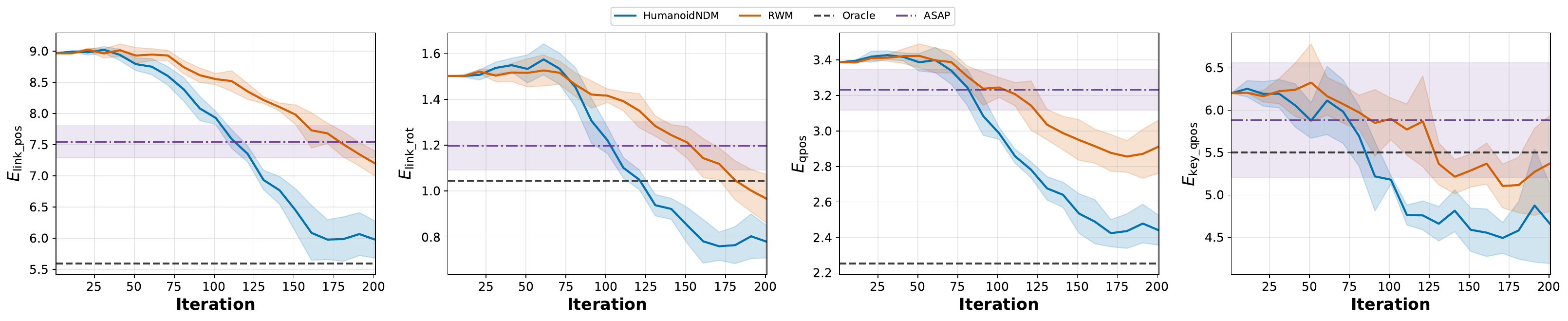} \\
    \includegraphics[width=0.7\linewidth]{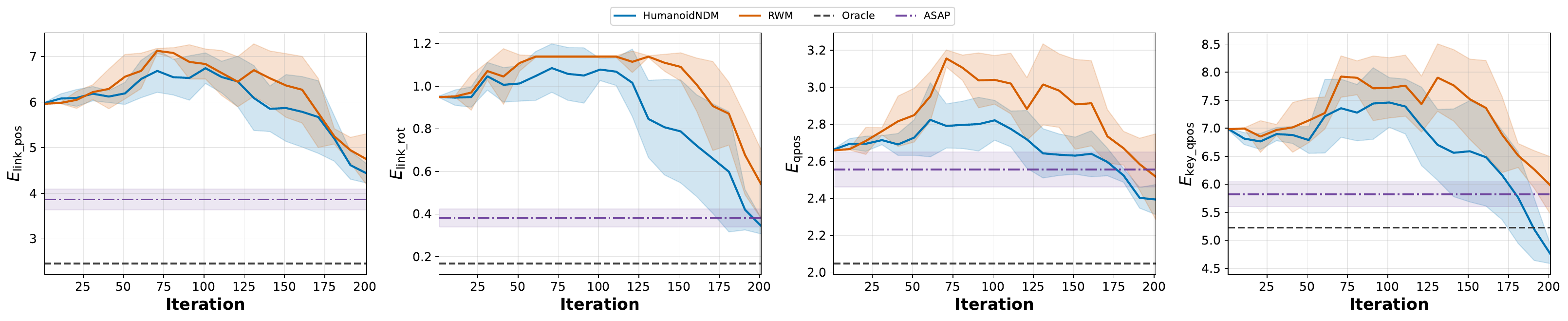} \\
    \includegraphics[width=0.7\linewidth]{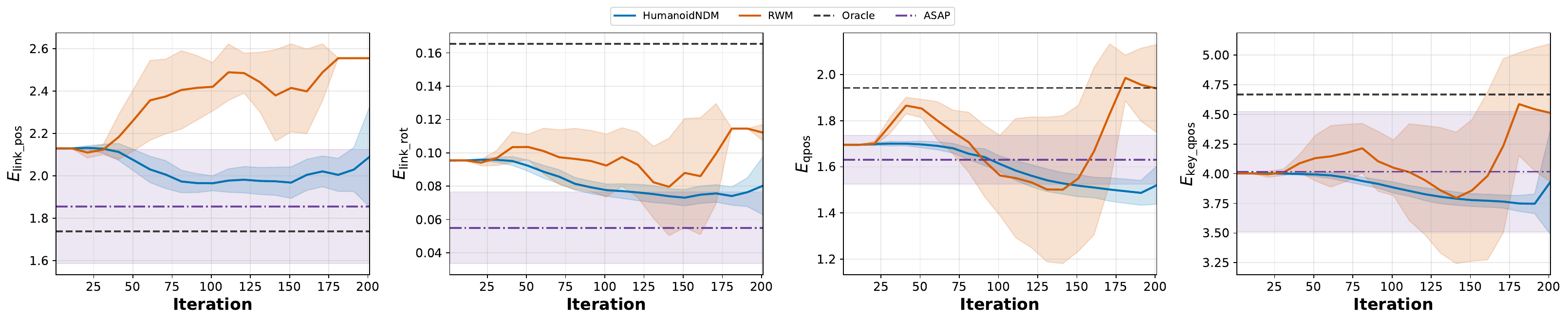}
    \caption{Squat: Extra Weight (top), Scaled KP (middle), Small Friction (bottom).}
    \label{fig:baseline_squat}
\end{figure}

\begin{figure}[H]
    \centering
    \includegraphics[width=0.7\linewidth]{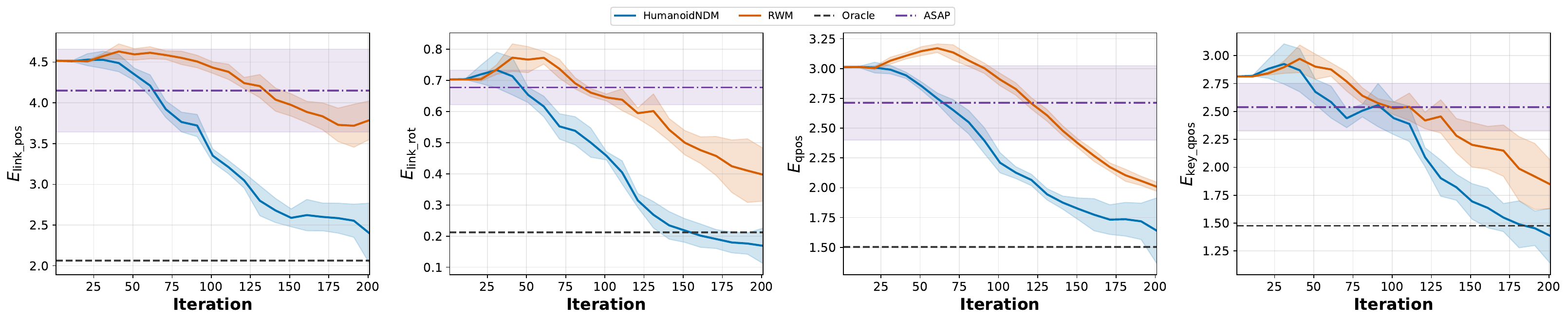} \\
    \includegraphics[width=0.7\linewidth]{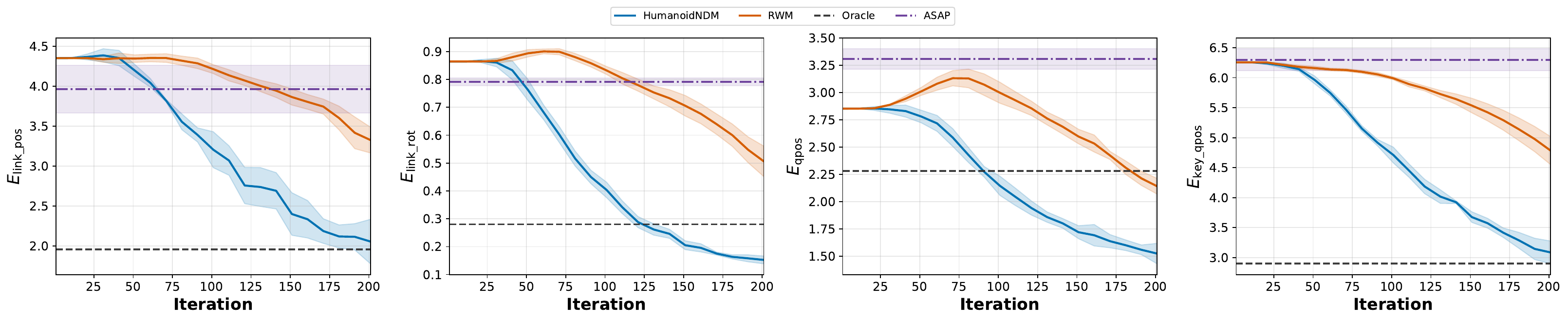} \\
    \includegraphics[width=0.7\linewidth]{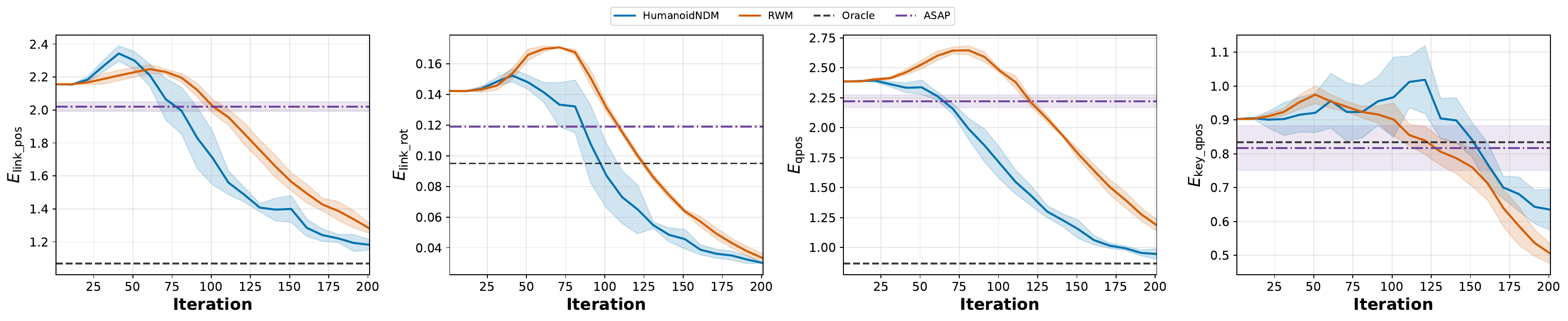}
    \caption{Swing: Extra Weight (top), Scaled KP (middle), Small Friction (bottom).}
    \label{fig:baseline_swing}
\end{figure}

\begin{figure}[H]
    \centering
    \includegraphics[width=0.7\linewidth]{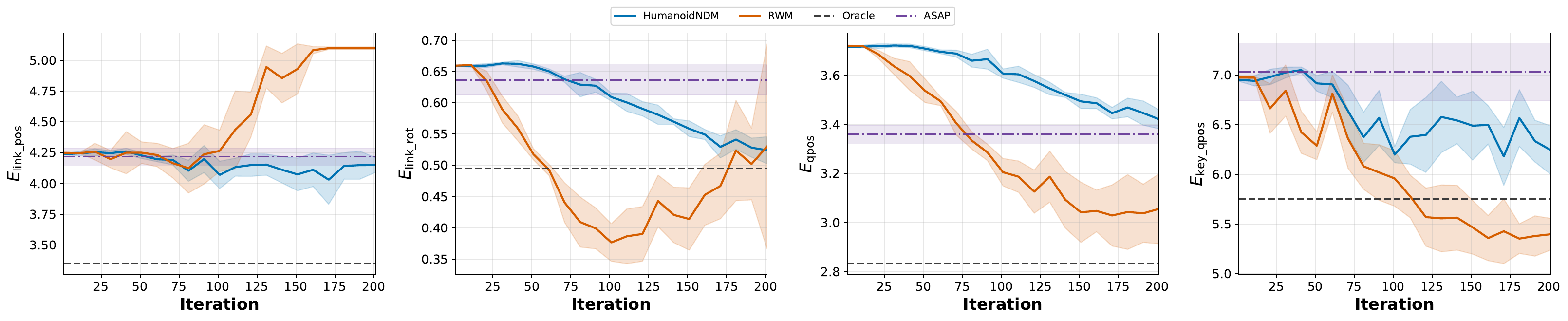} \\
    \includegraphics[width=0.7\linewidth]{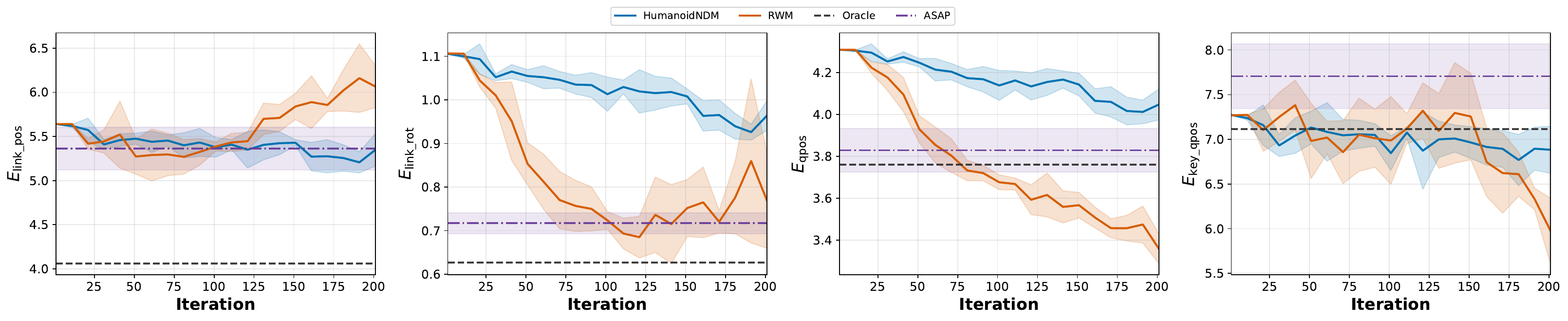} \\
    \includegraphics[width=0.7\linewidth]{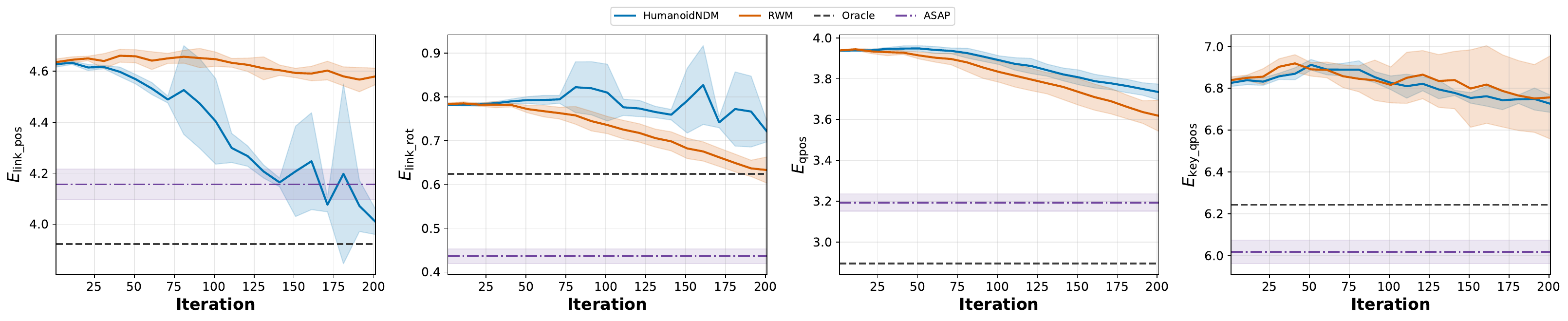}
    \caption{Stride: Extra Weight (top), Scaled KP (middle), Small Friction (bottom).}
    \label{fig:baseline_stride}
\end{figure}

The training curves of HumanoidNDM and baselines are shown in Figures~\ref{fig:baseline_walk}--~\ref{fig:baseline_stride}.
\vspace{-1em}

\vspace{-1em}

\vspace{-1em}

\vspace{-1em}

\section{Analysis of Perturbation Intensity Adaptation}

\begin{figure}[htbp]
    \centering
    \includegraphics[width=\linewidth]{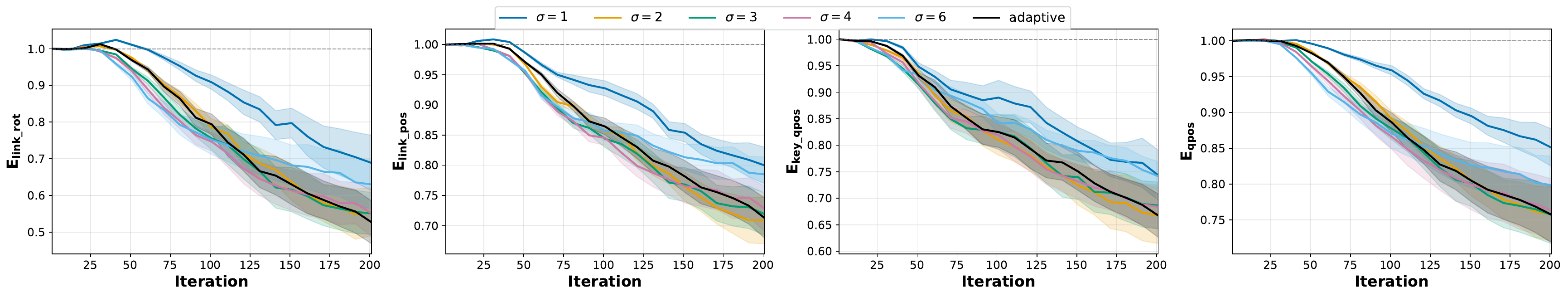}
    \caption{Comparison between the adaptive and fixed $\sigma$ under the default simulation setting.}
    \label{fig:ablation_adaptive1}
\end{figure}

In the simulation setting of our main experiments, we compare adaptive perturbation intensity with fixed scales, as shown in Figure~\ref{fig:ablation_adaptive1}. Fixed $\sigma$ values of 2, 3, and 4 achieve similar fine-tuning performance, significantly outperforming $\sigma=1$ (insufficient exploration) and $\sigma=6$ (overly aggressive exploration). Adaptive intensity achieves nearly identical performance to these well-chosen fixed scales. We attribute this similarity to the initial policies' reward design, which produces nearly indistinguishable policy responses to reference perturbations within a reasonable range of intensities.

\begin{figure}[htbp]
    \centering
    \includegraphics[width=0.3\linewidth]{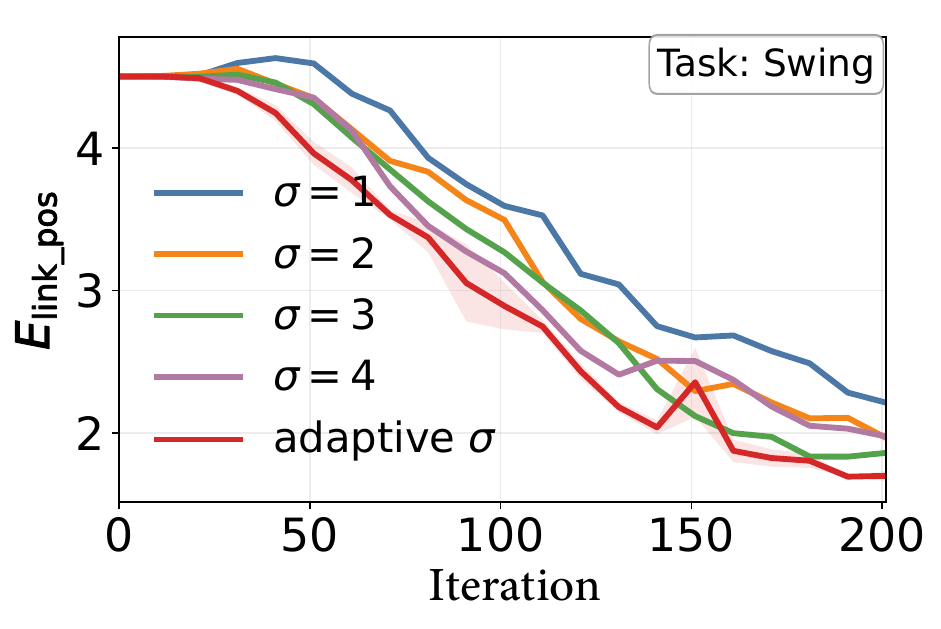}
    \includegraphics[width=0.3\linewidth]{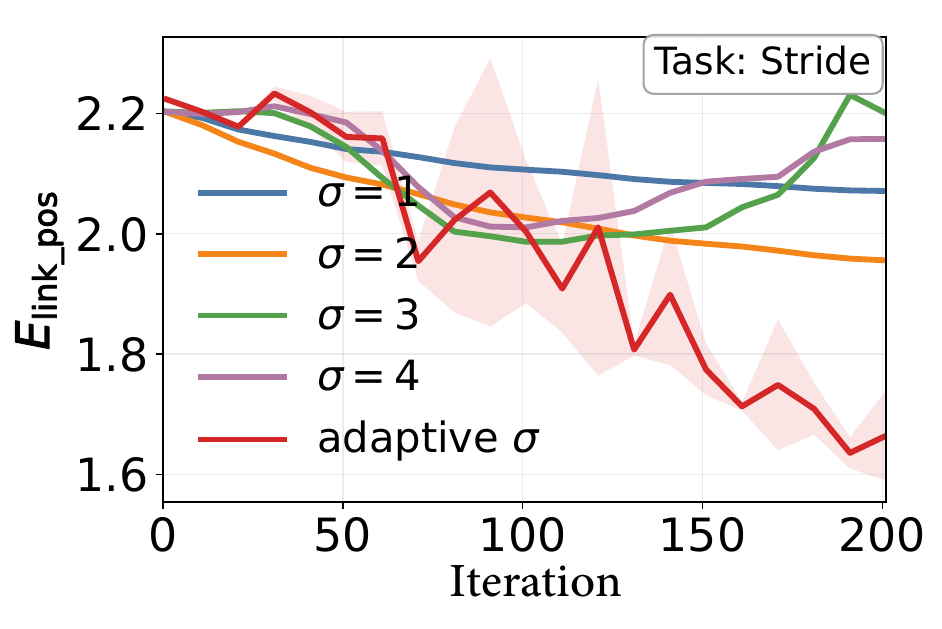}
    \caption{Comparison between adaptive and fixed perturbation intensities in real-world experiments on swing and stride motions.}
    \label{fig:ablation_adaptive2}
\end{figure}

However, adaptive intensity demonstrates clearer advantages in real-world experiments. As shown in Figure~\ref{fig:ablation_adaptive2}, adaptive exploration outperforms all fixed perturbation scales for both stride and swing motions. We hypothesize that this advantage arises from matching perturbation intensity to exploration requirements at different stages. Larger perturbations facilitate broader exploration in early stages, whereas smaller perturbations become sufficient as training approaches convergence, particularly for inherently less stable motions such as stride. Excessive perturbations near convergence may instead introduce unstable data and hinder fine-tuning.

\begin{figure}[htbp]
    \centering
    \begin{subfigure}{\linewidth}
        \includegraphics[width=\linewidth]{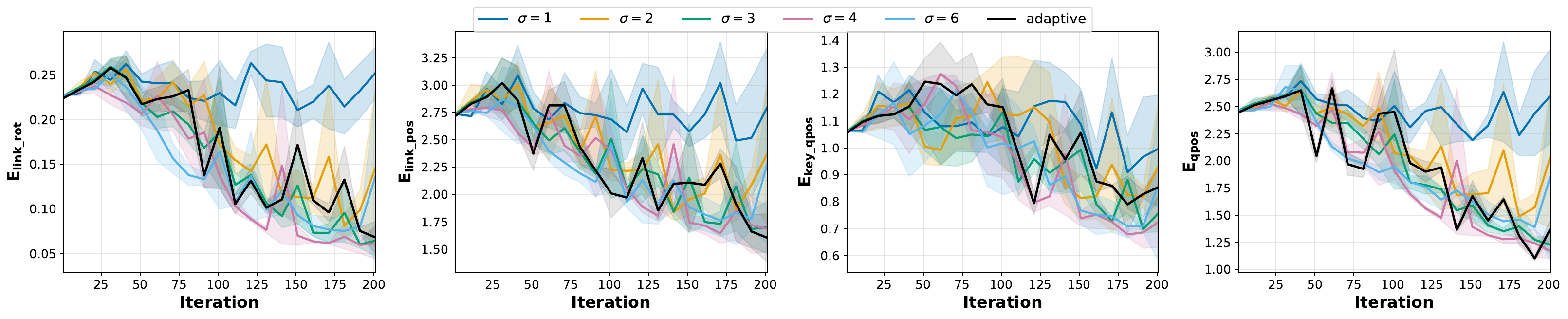}
    \end{subfigure}
    
    \begin{subfigure}{\linewidth}
        \includegraphics[width=\linewidth]{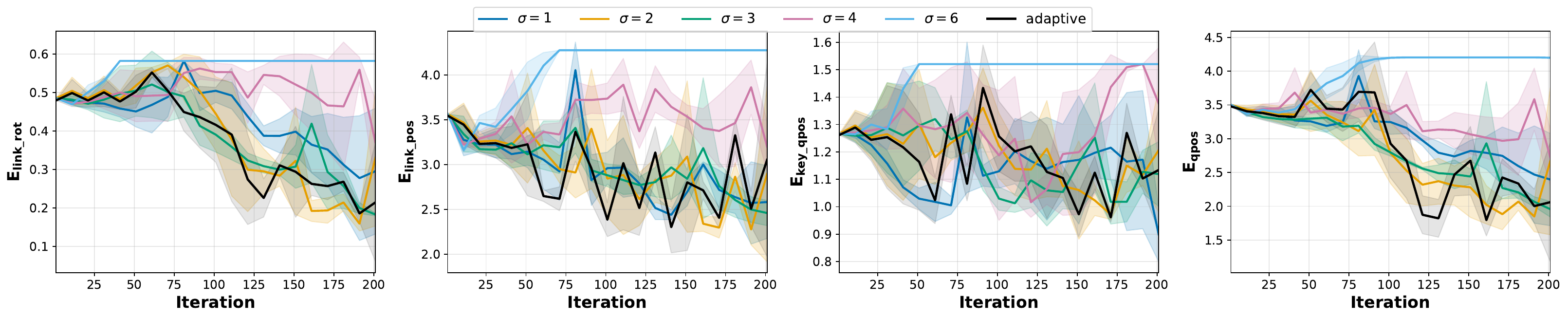}
    \end{subfigure}
    
    \begin{subfigure}{\linewidth}
        \includegraphics[width=\linewidth]{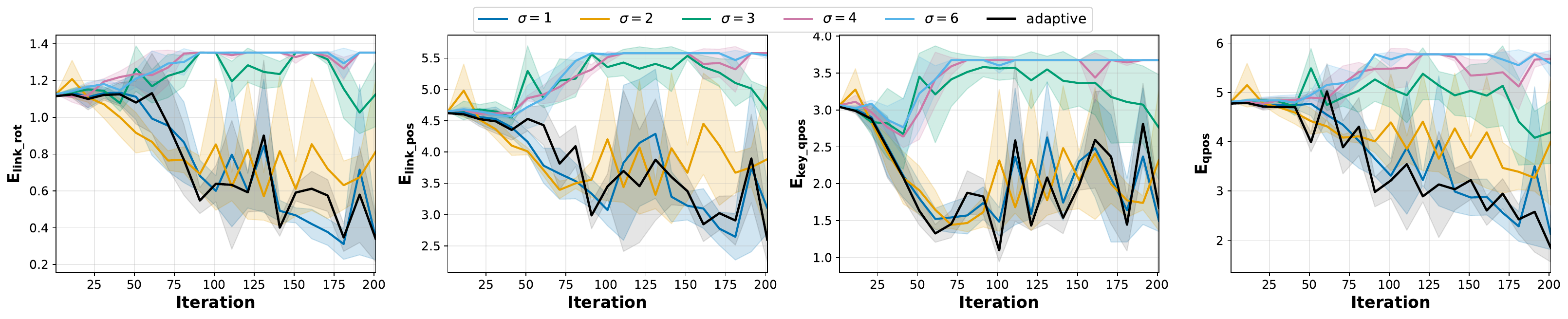}
    \end{subfigure}
    
    \caption{Comparison between the adaptive and fixed $\sigma$ using policies with varying degrees of responsiveness (\textbf{top}: \textbf{weak}; \textbf{middle}: \textbf{medium}; \textbf{bottom}: \textbf{strong}).}
    \label{fig:ablation_adaptative2}
\end{figure}

To further evaluate FENCE's adaptive perturbation intensity mechanism, we trained three initial policies for the \textit{swing} motion with different responsiveness to reference perturbations, primarily controlled by adjusting the tracking-related reward weight during PPO training. The corresponding results are shown in Figure~\ref{fig:ablation_adaptative2}. For weakly responsive policies (top), larger $\sigma$ during data collection improves fine-tuning performance. In contrast, highly responsive policies (middle) require only smaller $\sigma$ for sufficient exploration and optimal fine-tuning performance, whereas excessive perturbations degrade performance. For policies with intermediate responsiveness (bottom), moderate perturbations yield the best fine-tuning results. These findings indicate that the optimal perturbation intensity depends on the initial policy's responsiveness to reference perturbations. FENCE automatically adapts the perturbation intensity to different policies using a single set of hyperparameters, enabling efficient fine-tuning without manually tuning the perturbation scale for each policy.

\section{Applicability Beyond Ground-Contact Motions}

The experiments in the main paper focus primarily on motions with continuous ground contact. 
To further evaluate the applicability of HumanoidNDM to motions involving flight phases, we extend our experiments to a jump motion, during which both feet leave the ground.

For motions with a flight phase, the root-height estimate $h_t$ becomes unreliable due to the absence of valid foot contact. We therefore remove root height from the NDM input, output, and prediction loss, and model dynamics solely from proprioceptive history. For the NDM link-position loss and policy fine-tuning loss, global link positions are computed using the reference root height, avoiding the propagation of root-height estimation errors. All other components remain unchanged.

The jump motion and disturbance settings are described below.

\noindent
\begin{minipage}[t]{0.75\linewidth}
    \vspace{0pt}
    \paragraph{Jump}
    Starting from the default pose, the robot performs a jump motion during which both feet leave the ground. It then lands and returns to the default pose.
    \begin{itemize}
        \item \textit{Extra Weight:} The mass of the \texttt{right\_knee\_link} is multiplied by 1.5, and its local center of mass is offset by [0, -0.2, 0].
        \item \textit{Scaled KP:} The KP of the \texttt{left\_hip\_pitch} is decreased from 100 to 20.
        \item \textit{Small Friction:} The coefficient of friction between the robot's feet and the ground is reduced from 1.0 to 0.3.
    \end{itemize}
\end{minipage}\hfill
\begin{minipage}[t]{0.22\linewidth}
    \vspace{0pt}
    \includegraphics[width=\linewidth]{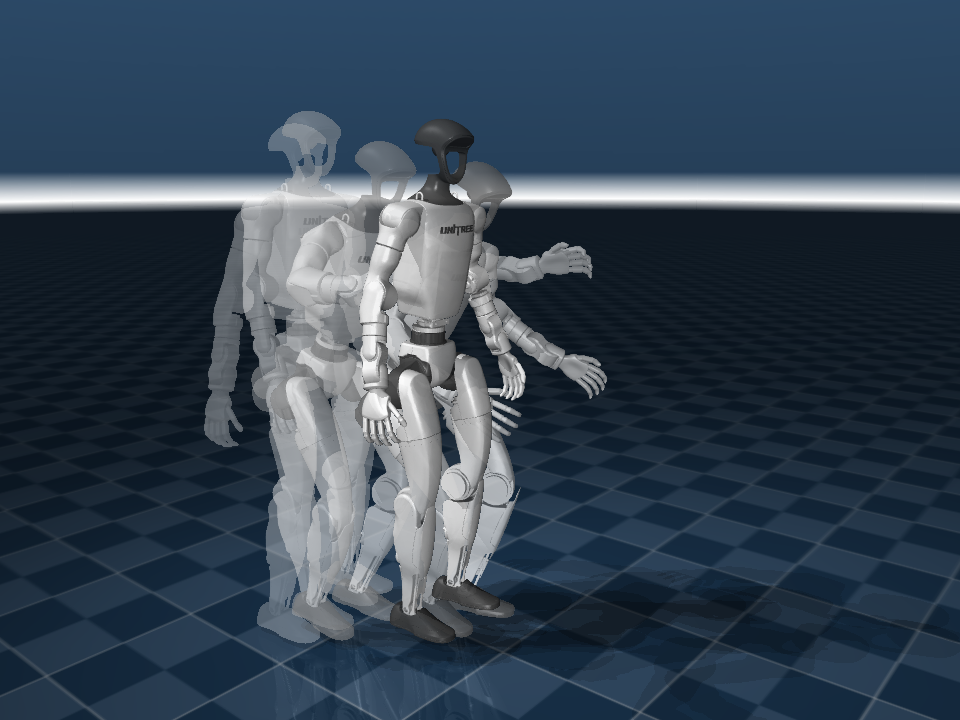}
\end{minipage}
\par\vspace{1em}

We evaluate HumanoidNDM under the same three types of dynamics disturbances used in the main simulation experiments. 
The quantitative results are reported in Table~\ref{tab:exp_jump}.

\begin{table*}[htbp]
\centering
\caption{Performance comparison of HumanoidNDM and baseline methods on the jump motion across 5 random trials.}
\label{tab:exp_jump}
\resizebox{0.93\textwidth}{!}{
\begin{tabular}{ll cccc cccc cccc}
\toprule
\multicolumn{2}{c}{Disturbance} &
\multicolumn{4}{c}{\cellcolor{red!10}Extra Weight} &
\multicolumn{4}{c}{\cellcolor{green!10}Scaled KP} &
\multicolumn{4}{c}{\cellcolor{blue!10}Small Friction} \\
\cmidrule(lr){3-6} \cmidrule(lr){7-10} \cmidrule(lr){11-14}
Trajectory & Method &
$E_{\text{link\_pos}}$ &
$E_{\text{link\_rot}}$ &
$E_{\text{qpos}}$ &
$E_{\text{key\_qpos}}$ &
$E_{\text{link\_pos}}$ &
$E_{\text{link\_rot}}$ &
$E_{\text{qpos}}$ &
$E_{\text{key\_qpos}}$ &
$E_{\text{link\_pos}}$ &
$E_{\text{link\_rot}}$ &
$E_{\text{qpos}}$ &
$E_{\text{key\_qpos}}$ \\
\midrule

\multirow{5}{*}{Jump}
& \cellcolor{gray!15}PPO w/ Disturb.
& \cellcolor{gray!15}3.03
& \cellcolor{gray!15}0.34
& \cellcolor{gray!15}1.90
& \cellcolor{gray!15}1.64
& \cellcolor{gray!15}4.15
& \cellcolor{gray!15}0.50
& \cellcolor{gray!15}2.86
& \cellcolor{gray!15}3.63
& \cellcolor{gray!15}4.42
& \cellcolor{gray!15}0.31
& \cellcolor{gray!15}2.05
& \cellcolor{gray!15}1.77 \\

\cmidrule(lr){2-14}

& Direct Transfer
& 6.74$\pm$0.00
& 1.56$\pm$0.00
& 4.77$\pm$0.00
& 6.50$\pm$0.00
& 7.41$\pm$0.00
& 1.77$\pm$0.00
& 5.01$\pm$0.00
& 8.75$\pm$0.00
& 7.53$\pm$0.00
& 1.80$\pm$0.00
& 4.92$\pm$0.00
& 6.74$\pm$0.00 \\

& ASAP
& \textbf{3.18$\pm$0.11}
& 0.63$\pm$0.07
& 3.30$\pm$0.14
& 3.15$\pm$0.20
& 7.08$\pm$0.09
& 1.98$\pm$0.12
& 4.56$\pm$0.16
& 5.98$\pm$0.41
& 4.59$\pm$0.09
& 0.47$\pm$0.03
& 3.52$\pm$0.02
& 5.05$\pm$0.27 \\

& RWM
& 5.79$\pm$0.99
& 0.92$\pm$0.33
& 3.49$\pm$0.04
& 3.29$\pm$0.33
& 5.66$\pm$0.03
& 0.75$\pm$0.02
& 3.99$\pm$0.01
& 5.81$\pm$0.25
& 4.73$\pm$0.04
& 0.64$\pm$0.01
& 3.73$\pm$0.03
& 3.60$\pm$0.03 \\

& \textcolor{red}{HumanoidNDM}
& 3.31$\pm$0.12
& \textbf{0.37$\pm$0.01}
& \textbf{2.97$\pm$0.03}
& \textbf{2.42$\pm$0.09}
& \textbf{4.68$\pm$0.22}
& \textbf{0.55$\pm$0.03}
& \textbf{3.25$\pm$0.05}
& \textbf{5.78$\pm$0.34}
& \textbf{4.28$\pm$0.98}
& \textbf{0.46$\pm$0.19}
& \textbf{2.63$\pm$0.28}
& \textbf{2.40$\pm$0.68} \\

\bottomrule
\end{tabular}
}
\end{table*}

\begin{figure}[htbp]
    \centering
    \includegraphics[width=0.75\linewidth]{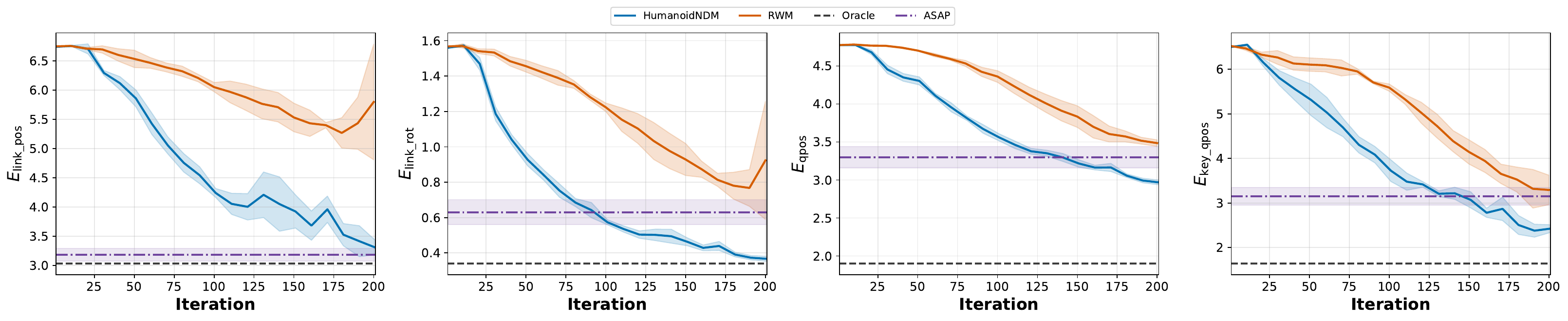} \\
    \includegraphics[width=0.75\linewidth]{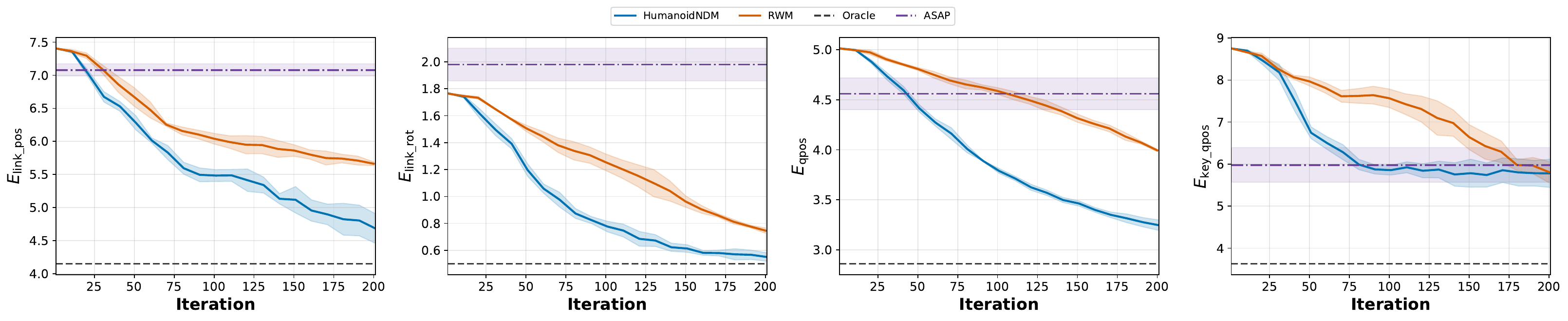} \\
    \includegraphics[width=0.75\linewidth]{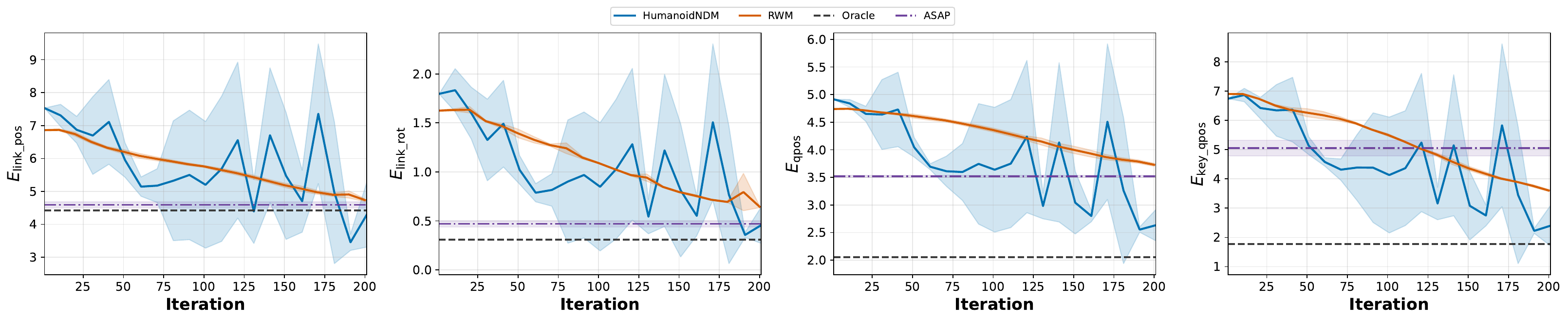}
    \caption{Jump: Extra Weight (top), Scaled KP (middle), Small Friction (bottom).}
    \label{fig:baseline_jump}
\end{figure}

The fine-tuning curves are shown in Figure~\ref{fig:baseline_jump}. HumanoidNDM consistently improves tracking performance under all three disturbances, demonstrating its applicability beyond motions with continuous ground contact.

\section{Effectiveness of NDM Pretraining}

\begin{figure}[htbp]
    \centering
    \vspace{-0.5em}
    \includegraphics[width=0.3\linewidth]{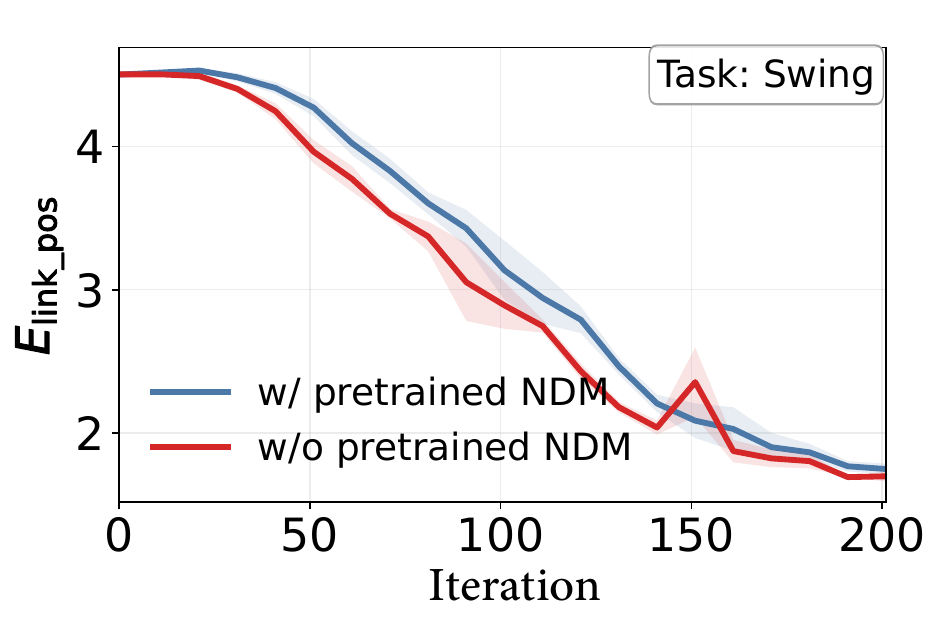}
    \caption{Comparison of HumanoidNDM with and without pretrained NDM.}
    \label{fig:model_pretrained}
\end{figure}

We compare NDM with and without pretraining on domain-randomized simulation data in real-world experiments. Figure~\ref{fig:model_pretrained} shows that pretraining offers no improvement. We attribute this to the structured dynamics shifts brought by real-world disturbance, which are hard to cover by simulator. 

\section{Visualization of Model Ablations}
\begin{figure}[htbp]
    \centering
    \begin{subfigure}[b]{0.24\linewidth}
        \centering
        \includegraphics[width=\linewidth]{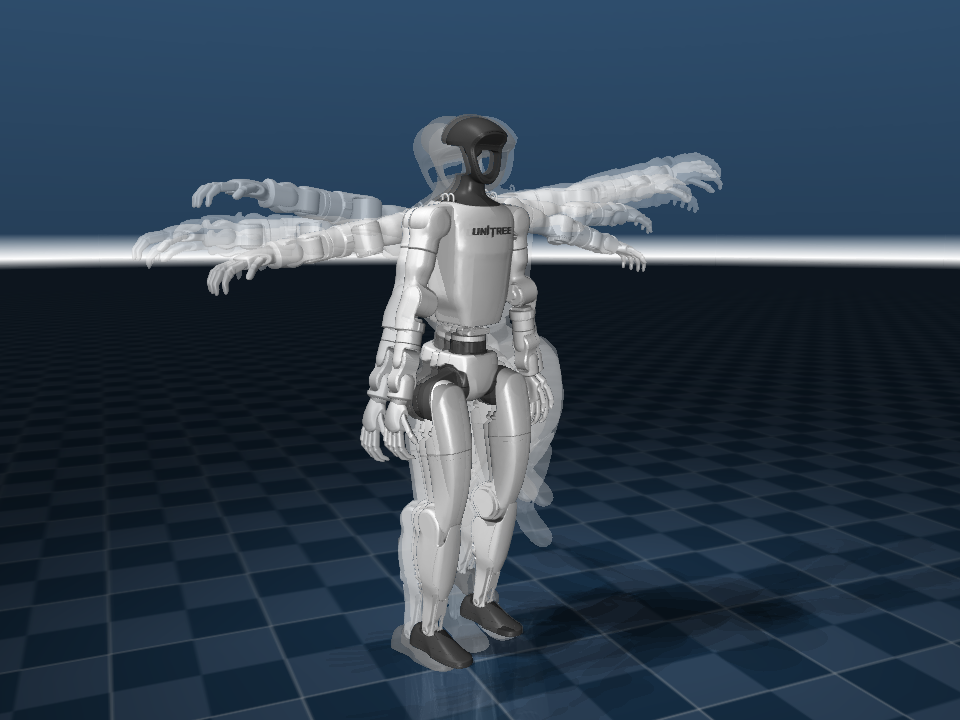}
        \caption{HumanoidNDM}
    \end{subfigure}\hfill
    \begin{subfigure}[b]{0.24\linewidth}
        \centering
        \includegraphics[width=\linewidth]{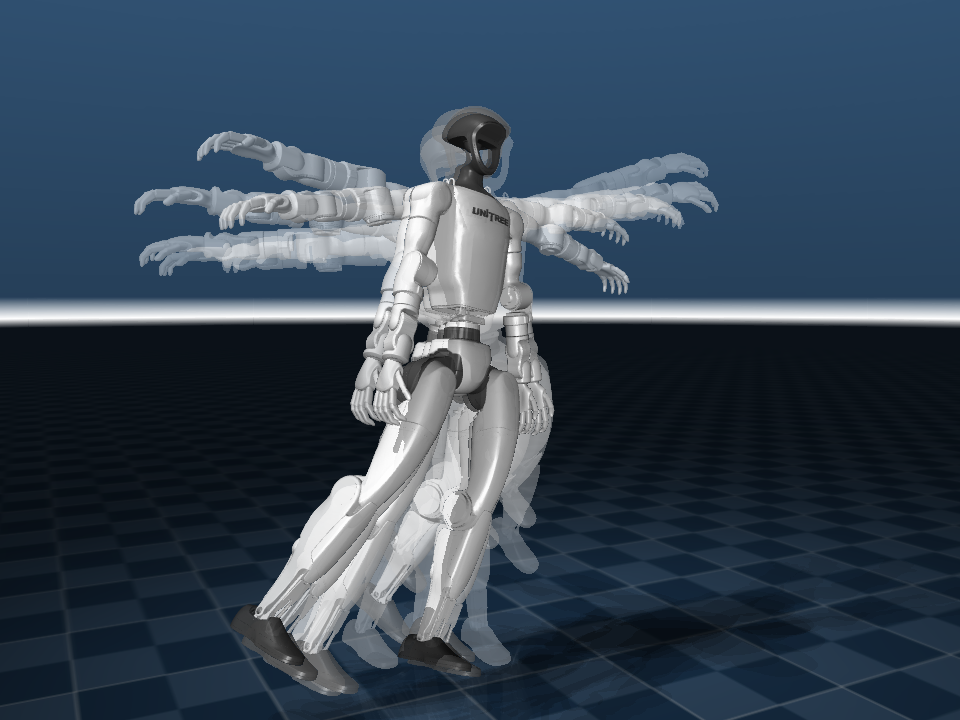}
        \caption{w/o history}
    \end{subfigure}\hfill
    \begin{subfigure}[b]{0.24\linewidth}
        \centering
        \includegraphics[width=\linewidth]{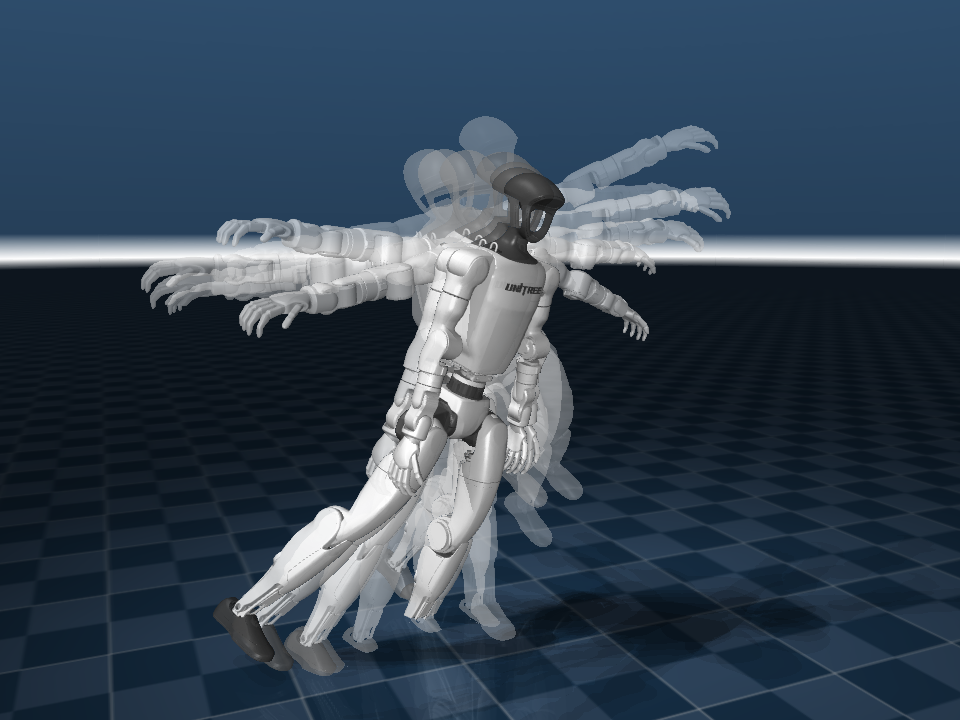}
        \caption{w/o Euclid objective}
    \end{subfigure}\hfill
    \begin{subfigure}[b]{0.24\linewidth}
        \centering
        \includegraphics[width=\linewidth]{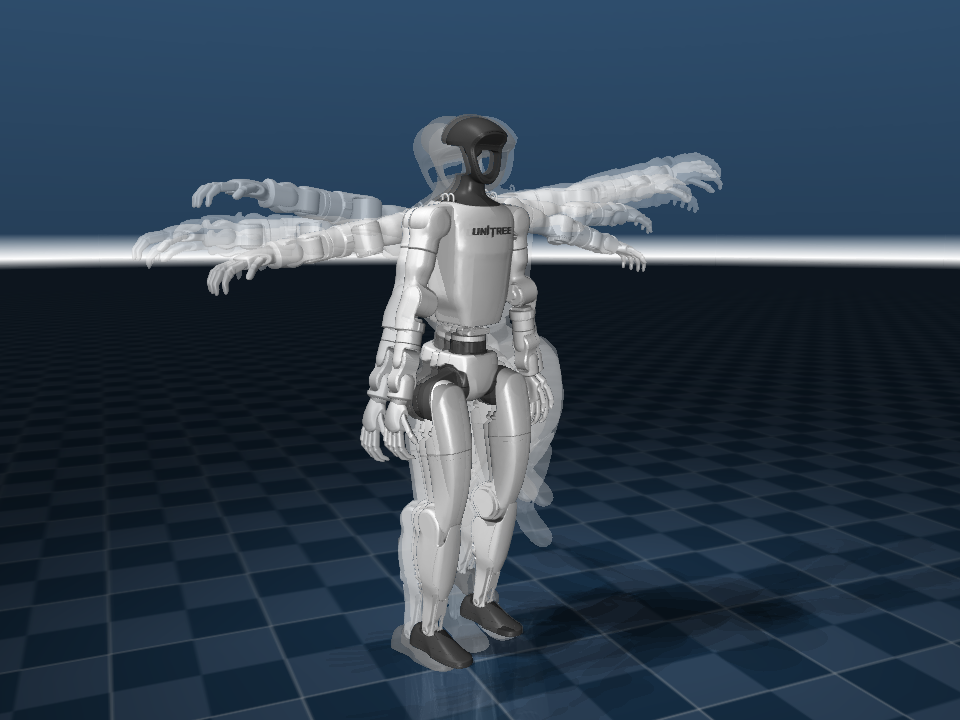}
        \caption{Ground Truth}
    \end{subfigure}
    
    \vspace{0.5em}
    
    \includegraphics[width=0.96\linewidth]{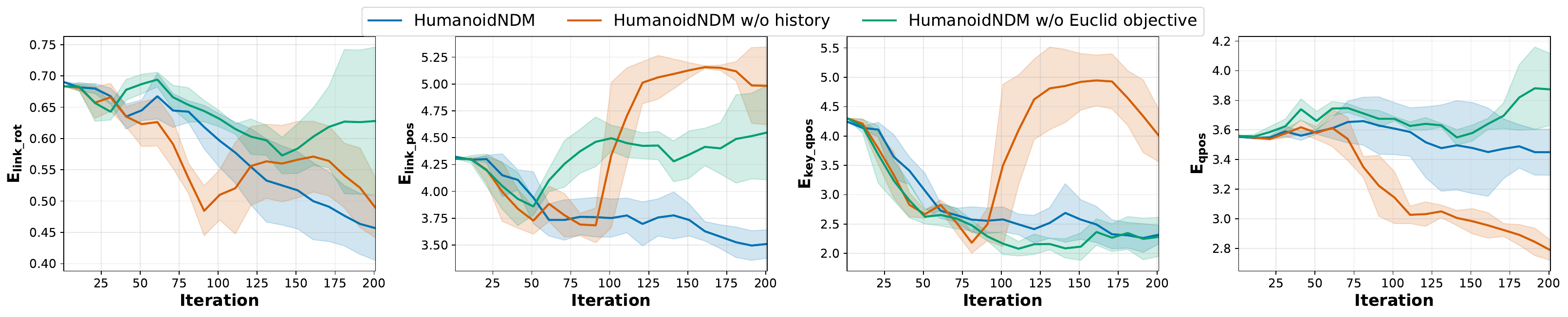}
    
    \caption{Model visualization and performance for balance motion with extra weight.}
    \label{fig:visualization_leg_balance_ankle_weight}
\end{figure}

\begin{figure}[htbp]
    \centering
    \begin{subfigure}[b]{0.24\linewidth}
        \centering
        \includegraphics[width=\linewidth]{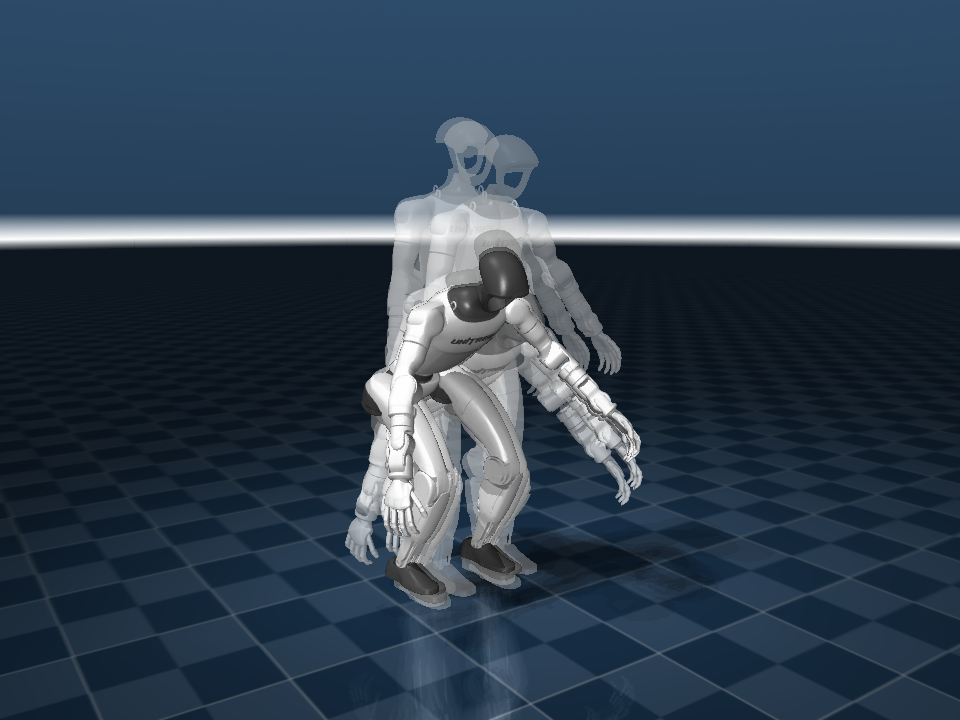}
        \caption{HumanoidNDM}
    \end{subfigure}\hfill
    \begin{subfigure}[b]{0.24\linewidth}
        \centering
        \includegraphics[width=\linewidth]{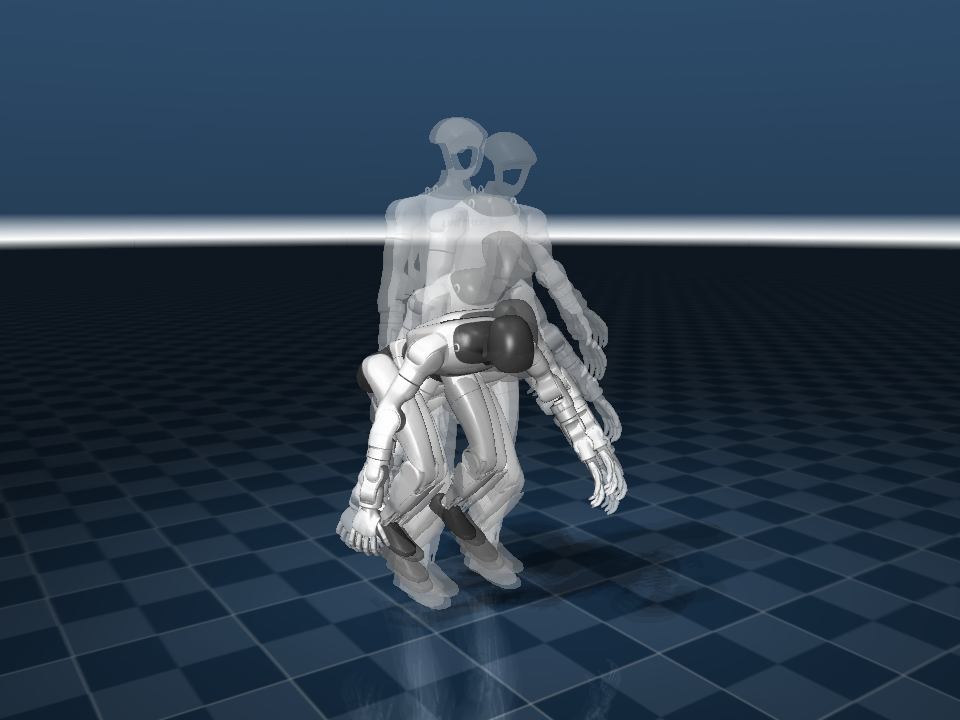}
        \caption{w/o history}
    \end{subfigure}\hfill
    \begin{subfigure}[b]{0.24\linewidth}
        \centering
        \includegraphics[width=\linewidth]{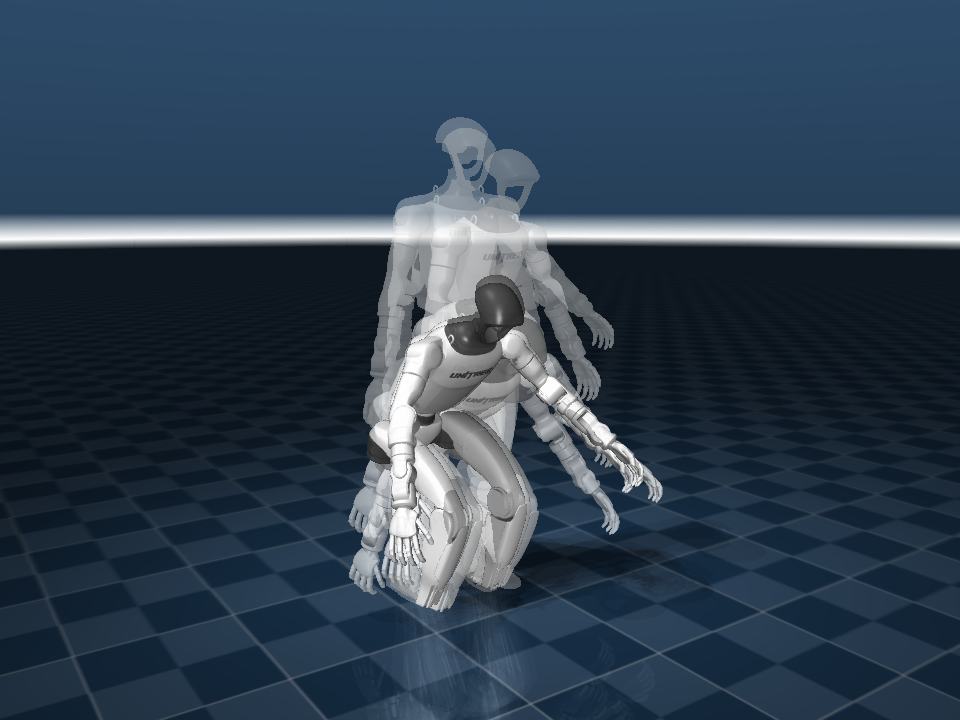}
        \caption{w/o Euclid objective}
    \end{subfigure}\hfill
    \begin{subfigure}[b]{0.24\linewidth}
        \centering
        \includegraphics[width=\linewidth]{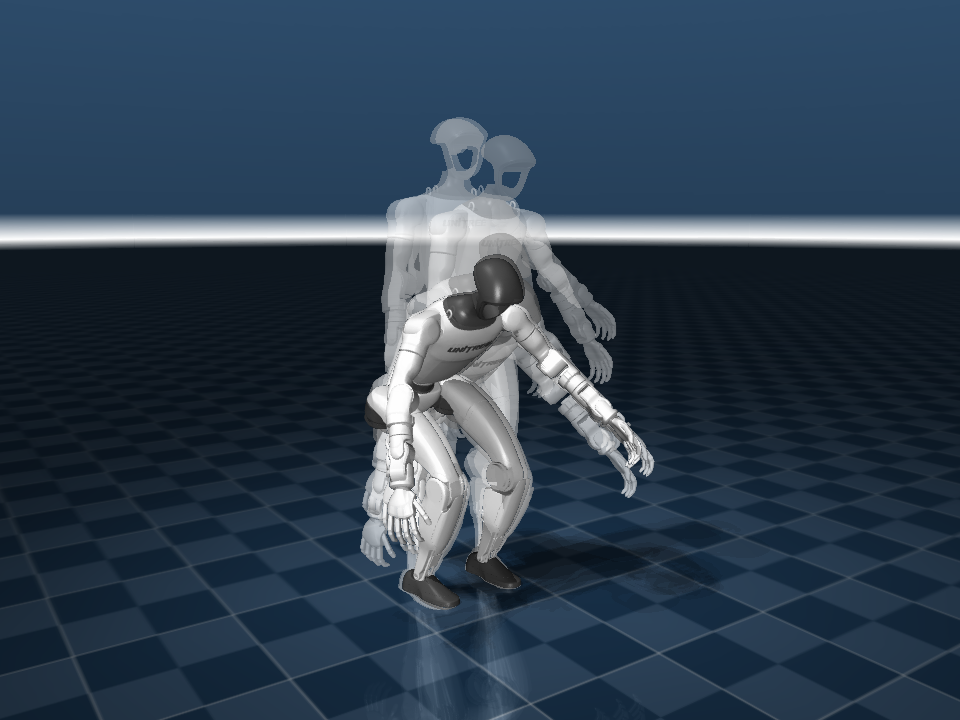}
        \caption{Ground Truth}
    \end{subfigure}
    
    \vspace{0.5em}
    
    \includegraphics[width=0.96\linewidth]{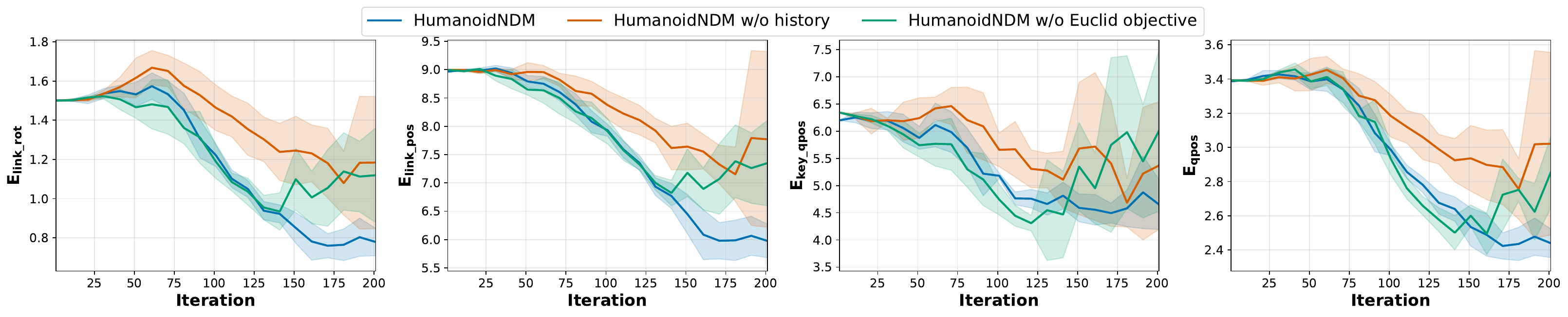}
    
    \caption{Model visualization and performance for squat motion with extra weight.}
    \label{fig:visualization_squat_up_arm_weight}
\end{figure}

\begin{figure}[htbp]
    \centering
    \begin{subfigure}[b]{0.24\linewidth}
        \centering
        \includegraphics[width=\linewidth]{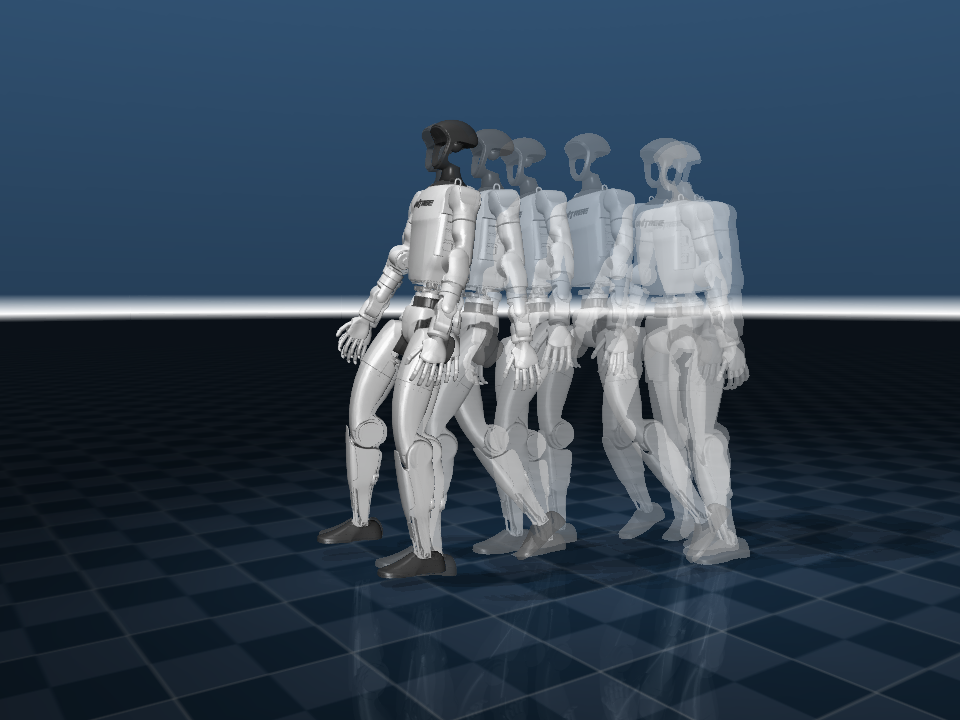}
        \caption{HumanoidNDM}
    \end{subfigure}\hfill
    \begin{subfigure}[b]{0.24\linewidth}
        \centering
        \includegraphics[width=\linewidth]{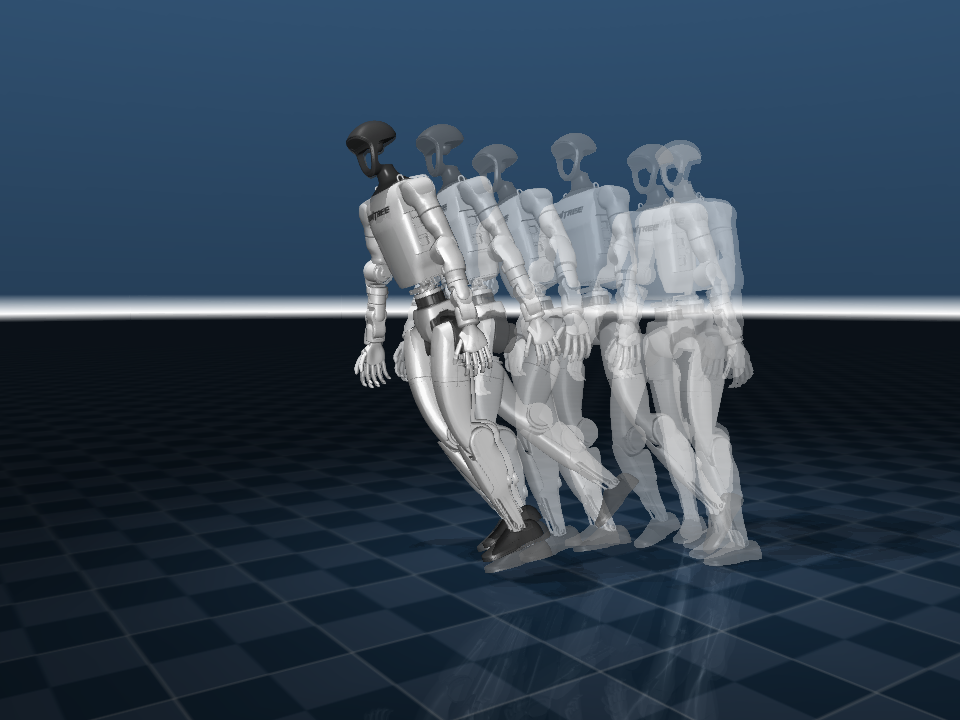}
        \caption{w/o history}
    \end{subfigure}\hfill
    \begin{subfigure}[b]{0.24\linewidth}
        \centering
        \includegraphics[width=\linewidth]{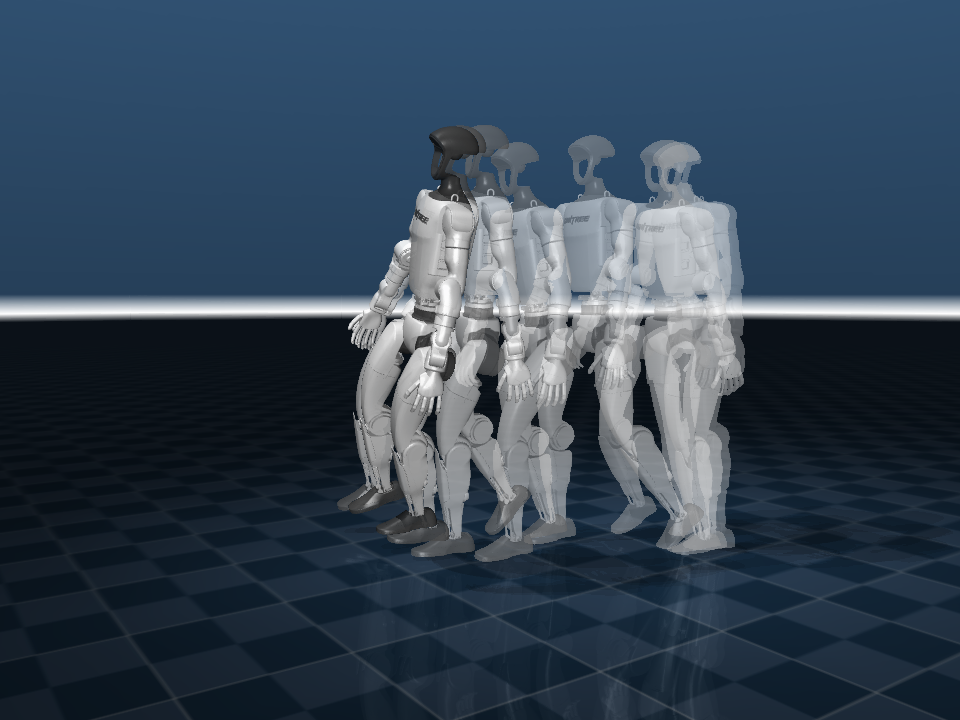}
        \caption{w/o Euclid objective}
    \end{subfigure}\hfill
    \begin{subfigure}[b]{0.24\linewidth}
        \centering
        \includegraphics[width=\linewidth]{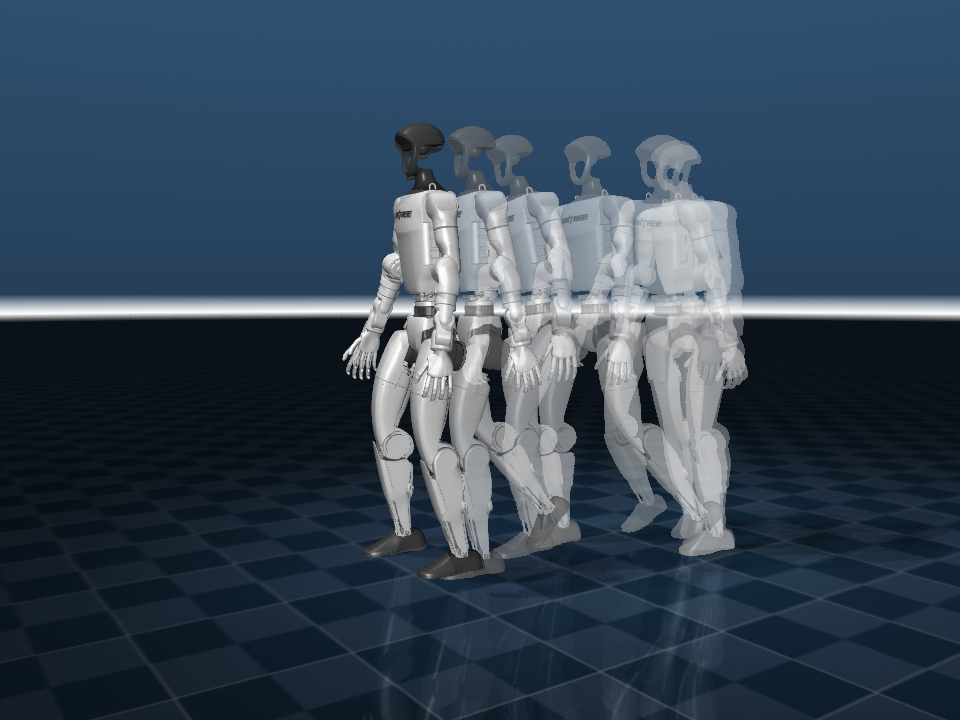}
        \caption{Ground Truth}
    \end{subfigure}
    
    \vspace{0.5em}
    
    \includegraphics[width=0.96\linewidth]{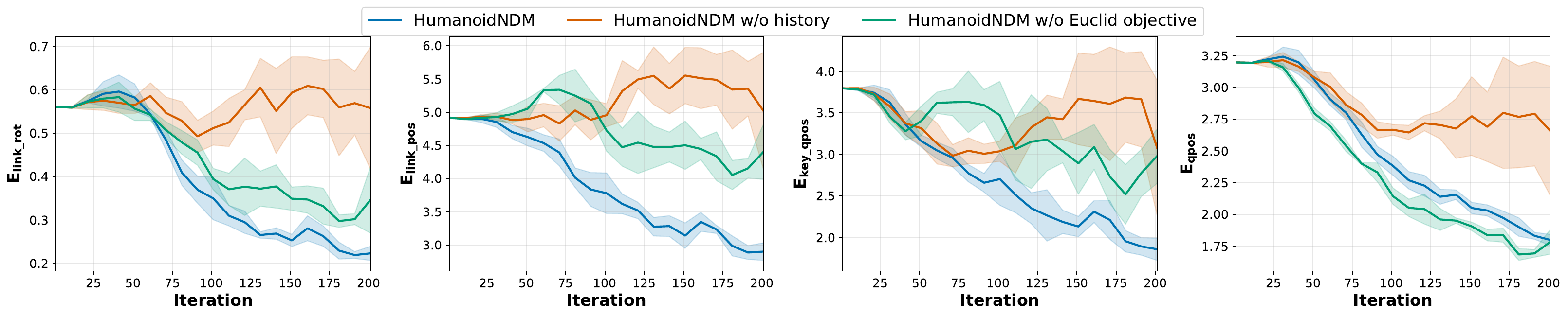}
    
    \caption{Model visualization and performance for walk motion with scaled KP.}
    \label{fig:visualization_walk_turn_torso_kphard}
\end{figure}

\begin{figure}[htbp]
    \centering
    \begin{subfigure}[b]{0.24\linewidth}
        \centering
        \includegraphics[width=\linewidth]{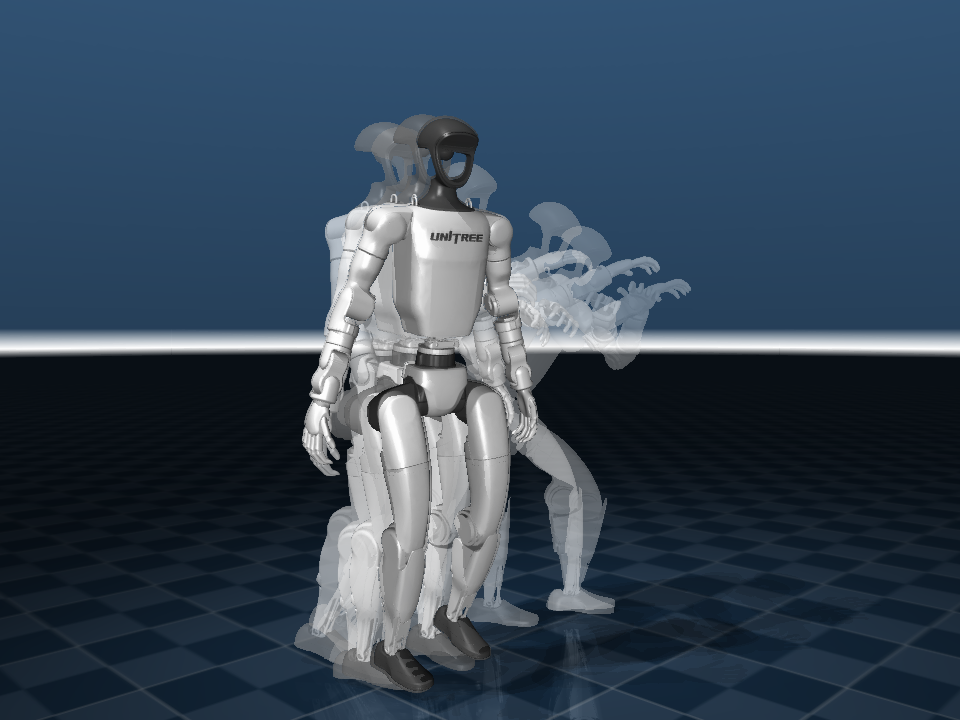}
        \caption{HumanoidNDM}
    \end{subfigure}\hfill
    \begin{subfigure}[b]{0.24\linewidth}
        \centering
        \includegraphics[width=\linewidth]{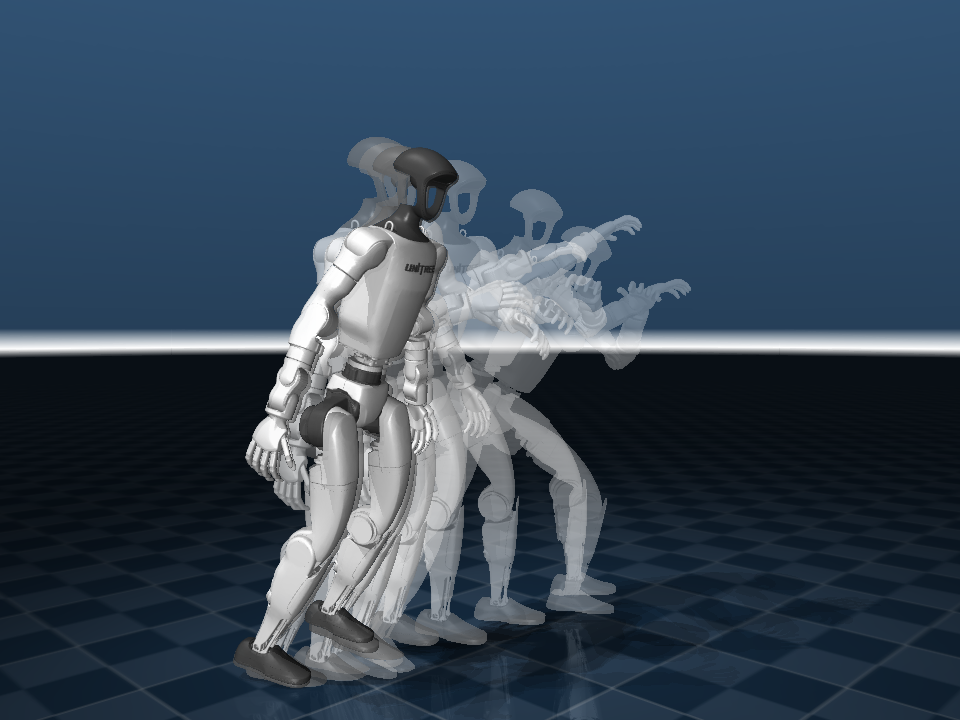}
        \caption{w/o history}
    \end{subfigure}\hfill
    \begin{subfigure}[b]{0.24\linewidth}
        \centering
        \includegraphics[width=\linewidth]{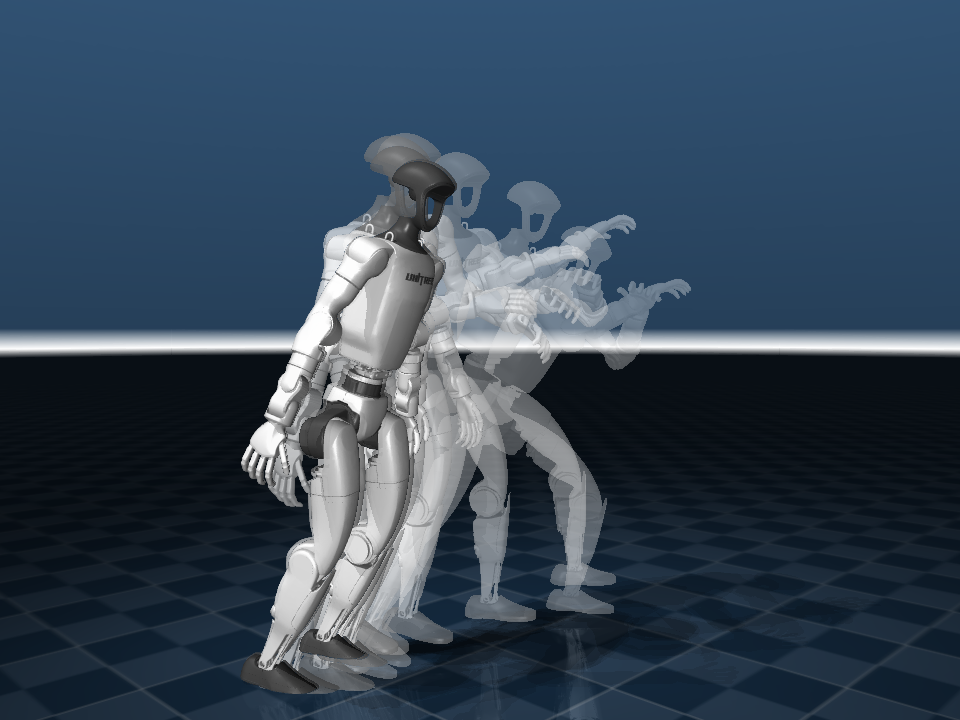}
        \caption{w/o Euclid objective}
    \end{subfigure}\hfill
    \begin{subfigure}[b]{0.24\linewidth}
        \centering
        \includegraphics[width=\linewidth]{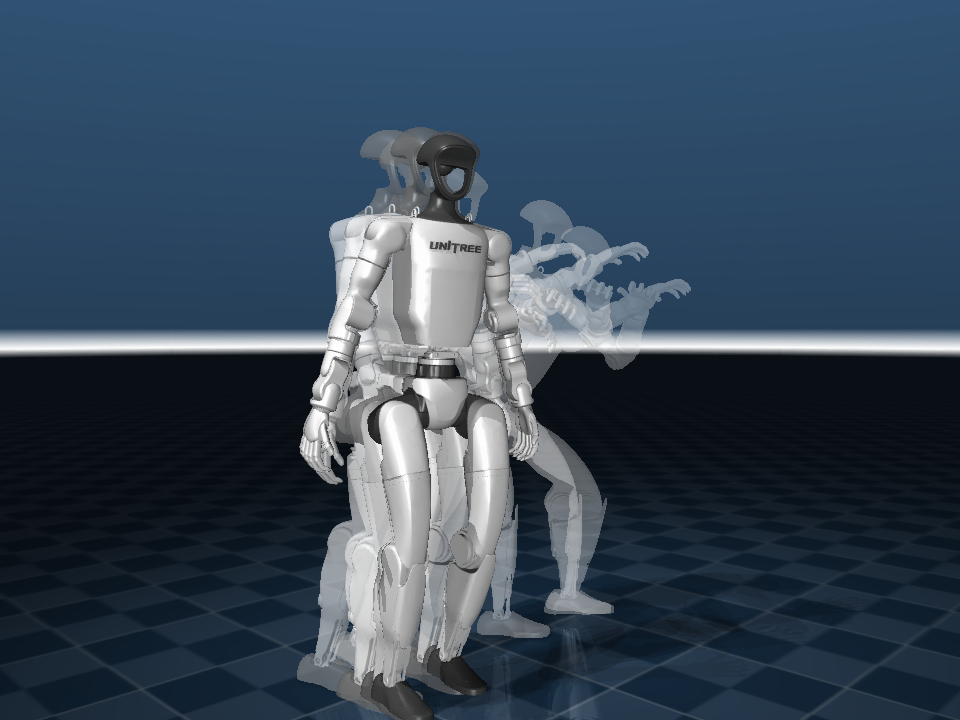}
        \caption{Ground Truth}
    \end{subfigure}
    
    \vspace{0.5em}
    
    \includegraphics[width=0.96\linewidth]{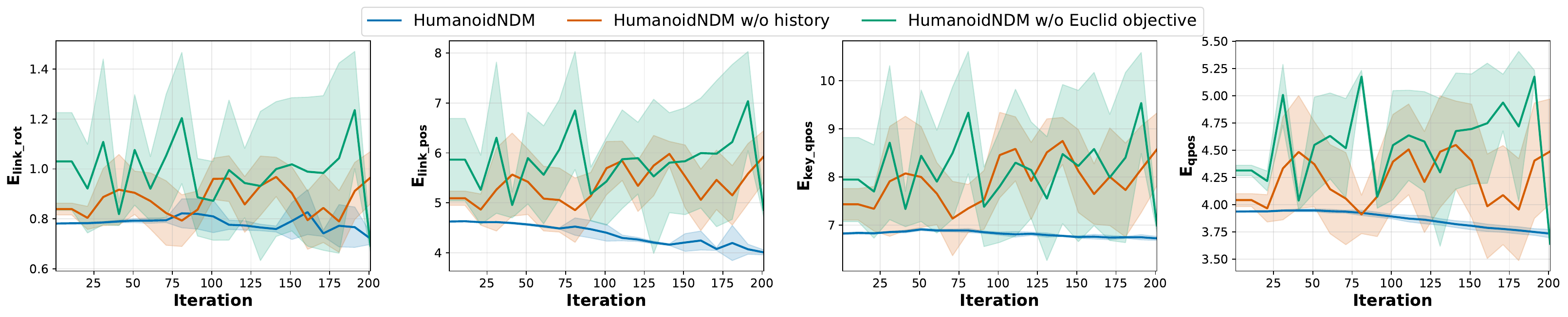}
    
    \caption{Model visualization and performance for stride motion with small friction.}
    \label{fig:visualization_side_leg_friction}
\end{figure}

\begin{figure}[htbp]
    \centering
    \begin{subfigure}[b]{0.24\linewidth}
        \centering
        \includegraphics[width=\linewidth]{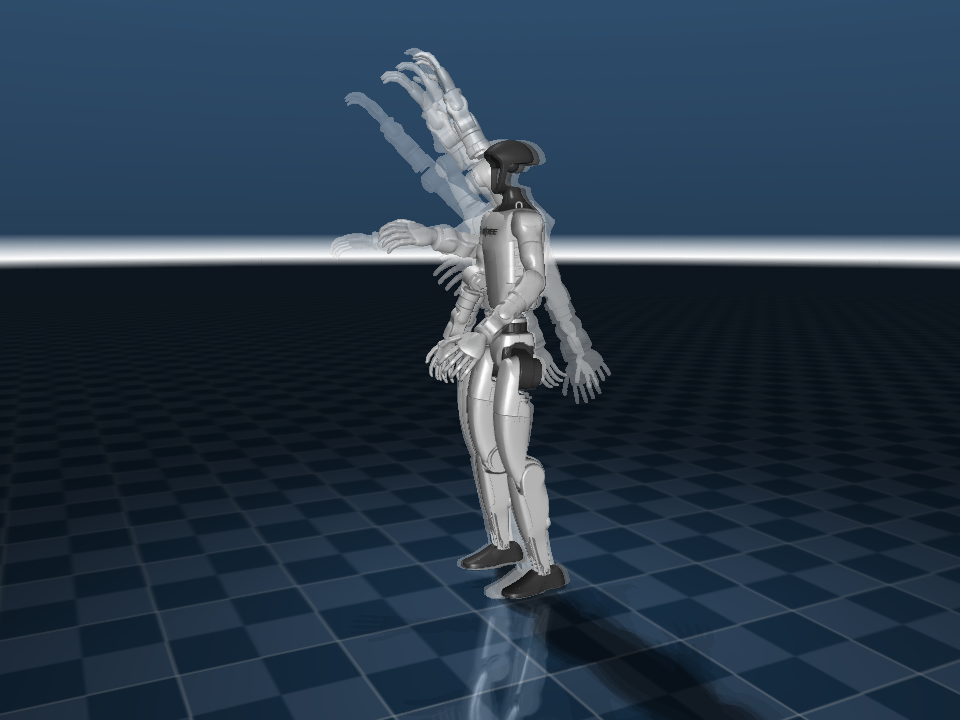}
        \caption{HumanoidNDM}
    \end{subfigure}\hfill
    \begin{subfigure}[b]{0.24\linewidth}
        \centering
        \includegraphics[width=\linewidth]{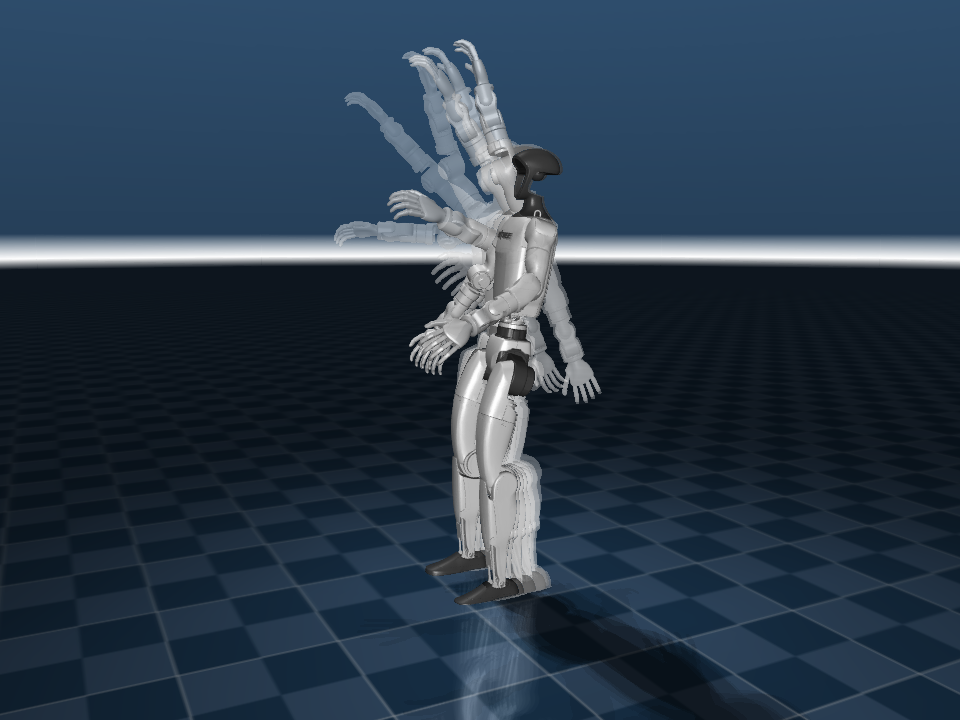}
        \caption{w/o history}
    \end{subfigure}\hfill
    \begin{subfigure}[b]{0.24\linewidth}
        \centering
        \includegraphics[width=\linewidth]{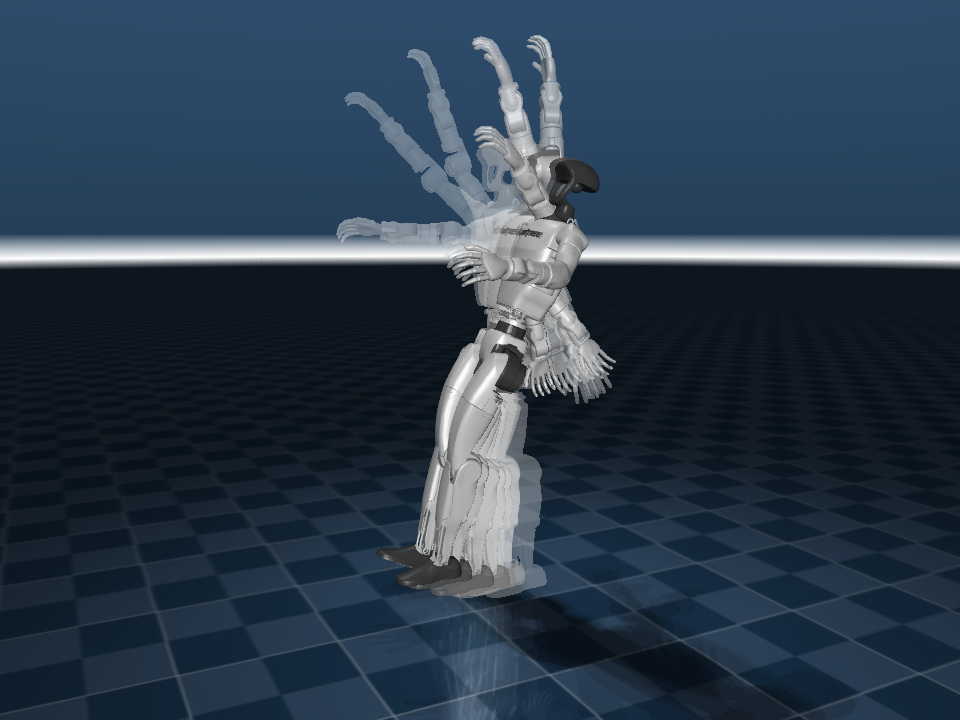}
        \caption{w/o Euclid objective}
    \end{subfigure}\hfill
    \begin{subfigure}[b]{0.24\linewidth}
        \centering
        \includegraphics[width=\linewidth]{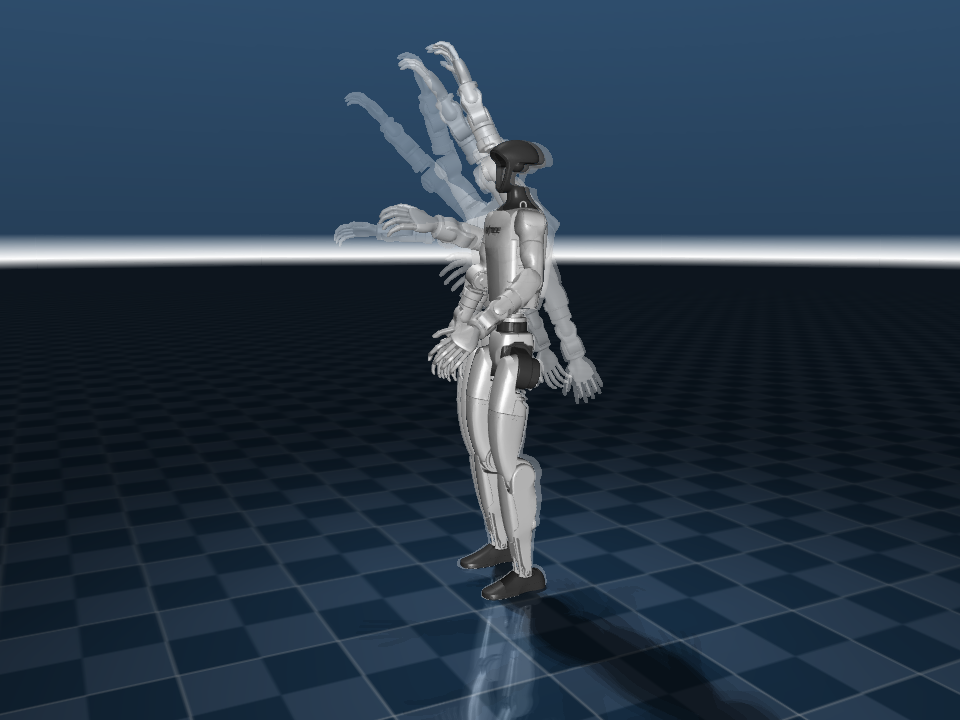}
        \caption{Ground Truth}
    \end{subfigure}
    
    \vspace{1em}
    
    \includegraphics[width=0.96\linewidth]{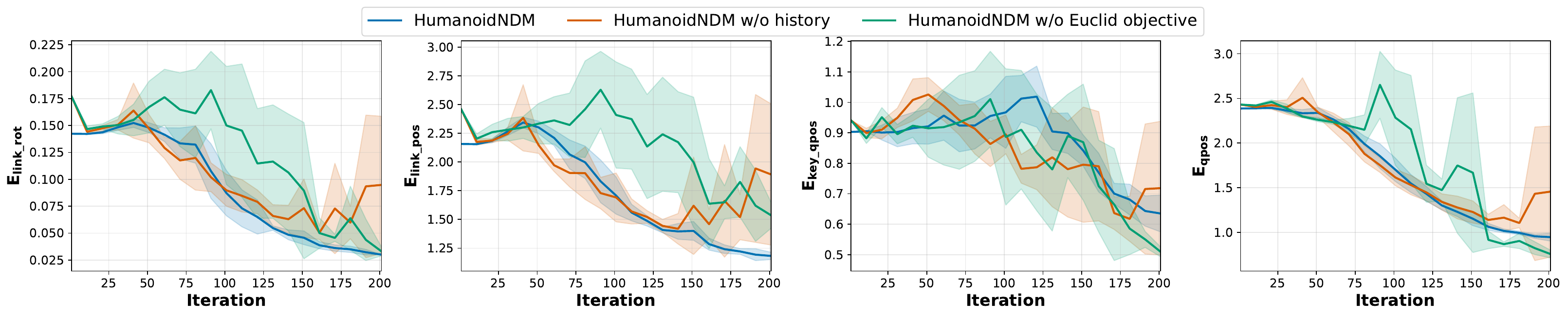}
    
    \caption{Model visualization and performance for swing motion with small friction.}
    \label{fig:visualization_swing_arm_friction}
\end{figure}

We compare three kinds of NDM: original implementation, a variant without history, and one without the Euclidean-space objective. Figures~\ref{fig:visualization_leg_balance_ankle_weight}--\ref{fig:visualization_swing_arm_friction} demonstrate that both history information and the dual-space objective are crucial for rollout accuracy and policy fine-tuning performance.

\vspace{-1em}

\vspace{-1em}

\vspace{-1em}

\vspace{-1em}

\end{document}